\documentclass{article}

\usepackage{arxiv}

\usepackage[utf8]{inputenc} 
\usepackage[T1]{fontenc}    
\usepackage{hyperref}       
\usepackage{url}            
\usepackage{booktabs}       
\usepackage{amsfonts}       
\usepackage{nicefrac}       
\usepackage{microtype}      
\usepackage{lipsum}
\usepackage{graphicx}
\usepackage{amsmath, nccmath, bm}
\usepackage{enumitem}
\usepackage{graphicx, multicol}
\usepackage{graphicx,subcaption,ragged2e}
\usepackage[utf8]{inputenc}
\usepackage{listings}
\usepackage{xcolor}
\usepackage{tgtermes}
\usepackage{amssymb}
\usepackage{float}
\usepackage{comment}
\usepackage{caption}
\usepackage{soul}
\usepackage{longtable}
\graphicspath{ {./} }

\title{PINN-FEM: A Hybrid Approach for Enforcing Dirichlet Boundary Conditions in Physics-Informed Neural Networks}

\author{
Nahil~Sobh$^*$ \\
University of Illinois at Urbana-Champaign\\
\texttt{sobh@illinois.edu} \\
   \And
 Rini Jasmine Gladstone$^*$ \\
  University of Illinois at Urbana-Champaign\\
  \texttt{rjg7@illinois.edu} \\
  \And
 Hadi Meidani \\
  University of Illinois at Urbana-Champaign\\
  \texttt{meidani@illinois.edu} \\
}

\begin{document}
\maketitle
\def\thefootnote{*}\footnotetext{These authors contributed equally to this work}\def\thefootnote{\arabic{footnote}}
\begin{abstract}
Physics-Informed Neural Networks (PINNs) solve partial differential equations (PDEs) by embedding governing equations and boundary/initial conditions into the loss function. However, accurately enforcing Dirichlet boundary conditions remains challenging, as soft enforcement often compromises convergence and reliability, particularly in complex domains. We propose a hybrid approach, PINN-FEM, that combines PINNs with finite element methods (FEM) to impose strong Dirichlet boundary conditions through domain decomposition. This method incorporates FEM-based representations near the boundaries, ensuring exact enforcement without sacrificing convergence. Across six experiments of increasing complexity, PINN-FEM outperforms standard PINN models, demonstrating superior accuracy and robustness. While techniques such as distance functions have been proposed for boundary condition enforcement, they often lack generality for real-world applications. PINN-FEM addresses this gap by leveraging FEM near boundaries, making it highly suitable for industrial and scientific problems.
\end{abstract}


\section{Introduction}
\label{introduction}

The developments in the field of machine learning and deep learning have found applications in diverse areas such as computer vision, natural language processing, and more. In recent years, the concept of scientific machine learning (SciML) has gained traction, utilizing the power of machine learning (ML) to solve scientific computing problems \cite{hey2020machine, mjolsness2001machine}. It achieves this by embedding the underlying physics and governing equations (PDEs or ODEs) of the problems at different stages of ML training. Generally, this can be done using experimental or simulation data or through physical loss functions \cite{karniadakis2021physics}. The traditional approach is to use simulation data for training ML models as supervised models, enabling them to understand the underlying physics by learning patterns from the data. Some of the popular deep learning architectures for solving PDE problems, such as MeshGraphNet \cite{pfaff2020learning}, DeepONet \cite{lu2021learning}, and Neural Operators \cite{li2020fourier}, use a supervised learning setting. The second approach, which employs physical loss functions, adds a soft implementation of underlying governing equations and physical constraints to the training algorithm in the form of loss functions. Physics-Informed Neural Networks (PINNs) \cite{karniadakis2021physics} represent the most popular architecture in this category. PINNs train neural networks to approximate the physical behavior of problems in a spatial-temporal domain, including the satisfaction of initial and boundary conditions. Over recent years, many studies have applied PINNs to solve a wide range of problems in mechanics \cite{moseley2020solving}, fluid dynamics \cite{cai2021physics, jin2021nsfnets}, power systems \cite{misyris2020physics}, and geosciences \cite{almajid2022prediction, li2022deep}.

In recent years, there have been many developments in mesh-free approaches based on PINNs to solve low and high-order PDE problems, such as the collocation method \cite{berg2018unified, raissi2019physics, sirignano2018dgm}, variational principles \cite{ew2018deep, han2018solving}, and Galerkin domain decomposition \cite{kharazmi2019variational, han2018solving}. These approaches leverage the automatic differentiation \cite{baydin2018automatic} capability of neural networks to efficiently compute the derivatives of nonlinear composite functions and use gradient descent algorithms \cite{bottou1991stochastic} to obtain accurate solutions for nonlinear, non-convex optimization problems. Collocation \cite{guo2021deep} and least square methods \cite{chen2020comparison, huang2020int} for training PINN rely on a strong-form to minimize the PDE residuals, whereas the variational principle method uses a weak-form loss formulation \cite{luong2024automatically}. The latter is widely used due to its increased efficiency from the simplicity of the loss function \cite{li2021physics}. This weak-form method constructs the loss function using the variational principle, halving the differential order of the PDE. Moreover, the Neumann and Robin boundary conditions (BCs) of the problem are included in the weak-form loss formulation, simplifying the loss function. The only additional loss term required is to satisfy Dirichlet BCs. \cite{yu2018deep} proposed the deep Ritz method for solving several PDE and eigenvalue problems by defining the loss function with the variational principle corresponding to the PDEs. \cite{samaniego2020energy} presented the deep energy method (DEM), a PINN framework that minimized the total potential energy of conservative structural systems and was successfully applied to solid mechanics problems such as elasticity, plates, and phase-field models \cite{nguyen2020deep, nguyen2021parametric, zhuang2021deep}.

Alternatively, the modification of mesh-free methods to couple with a finite element interpolation near the boundary \cite{belytschko1995coupled, huerta2000enrichment, wagner2001hierarchical} allows the direct imposition of prescribed values for the BCs. This is achieved by modifying the mesh-free methods to include finite elements via interpolation, making them easily applicable to various problems. \cite{belytschko1995coupled} coupled finite elements near the Dirichlet boundaries and element-free Galerkin in the interior of the computational domain to simplify the imposition of essential boundary conditions by performing mixed linear interpolation in the transition region. This required the substitution of finite element nodes by particles and the definition of ramp functions, limiting the transition to the size of one finite element. \cite{huerta2000enrichment} presented a general formulation for continuous blending to couple mesh-free methods with finite elements, enabling the exact imposition of essential (Dirichlet) BCs. \cite{wagner2001hierarchical} proposed a bridging scale method to couple a mesh-free approximation with other interpolations, particularly with finite elements near the Dirichlet boundary. Similarly, \cite{hegen1996element} coupled finite element domains and mesh-free regions using Lagrange multipliers. \cite{mitusch2021hybrid} introduced hybrid FEM-NN models for solving PDEs, combining finite element methods with neural networks. Here, Dirichlet BCs and ICs were enforced via spatial discretization of the domain using finite elements and applying them at mesh nodes. \cite{krongauz1996enforcement} demonstrated that Dirichlet boundary conditions can be enforced as in finite elements by placing finite element mesh layers along the Dirichlet boundary and imposing continuity at the interface between EFG and finite element meshes.

In this paper, we propose a novel hybrid method, PINN-FEM, to address the challenge of enforcing Dirichlet boundary conditions in Physics-Informed Neural Networks (PINNs). Inspired by blending techniques in computational mechanics, our approach integrates finite element methods (FEM) near Dirichlet boundaries with PINN approximations in the interior domain. This domain decomposition enables exact enforcement of Dirichlet boundary conditions while leveraging the flexibility and adaptability of PINNs for solving partial differential equations (PDEs). The proposed PINN-FEM method employs FEM-based representations to enforce boundary conditions, eliminating the need for additional penalty terms or soft enforcement techniques, which often compromise accuracy and convergence. By leveraging the principle of minimum potential energy, our method ensures stability and efficiency in training, seamlessly integrating FEM and PINN components. We validate PINN-FEM through six numerical experiments involving linear elasticity problems with increasing complexity in domain geometry and boundary conditions. These include scenarios with discontinuous, point, and complex boundary conditions. Our results demonstrate that PINN-FEM consistently outperforms baseline PINN models with both soft and exact enforcement strategies. Additionally, the method generalizes well across various configurations, making it a promising approach for industrial and scientific applications requiring accurate boundary condition enforcement.

The rest of the paper is organized as follows: Section~\ref{background} provides a background on Physics-Informed Neural Networks and their limitations. Section~\ref{methodology} details the proposed methodology. Section~\ref{results} presents the analysis of numerical results. This is followed by concluding remarks and a discussion of future work in Section~\ref{discussion}.

\section{Background}
\label{background}

\subsection{Physics-informed neural networks}
\label{pinn}

Physics-Informed Neural Networks (PINNs) \cite{nguyen2022efficient, raissi2019physics} are used to solve forward and inverse problems of physical systems governed by partial differential equations (PDEs) using neural networks. The architecture of a PINN is typically a fully connected neural network, where the inputs are spatial coordinates, and the output is the response of the physical system. For solid mechanics problems, they are used to predict displacement and stress fields. Let us consider a generalized boundary value problem (BVP) in $\mathcal{R}^n$ in solid mechanics, as follows:

\begin{equation}
    \begin{aligned}
    \mathcal{L}(\mathbf{x};u(\mathbf{x};\bm{\theta})) = f(\mathbf{x}), \quad \mathbf{x} \in \Omega, \\
    \mathcal{B}_g(\mathbf{x};u(\mathbf{x};\bm{\theta})) = g, \quad \mathbf{x} \in \Gamma_g, \\
    \mathcal{B}_h(\mathbf{x};u(\mathbf{x};\bm{\theta})) = h, \quad \mathbf{x} \in \Gamma_h,
\label{eq_bvp}
    \end{aligned}
\end{equation}
where, $\mathcal{L}(.)$ is the differential operator and $\mathcal{B}_g(.)$ and $\mathcal{B}_h(.)$ are the Dirichlet and Neumann boundary condition operators on boundaries $\Gamma_g$ and $\Gamma_h$ respectively. Thus the boundary, $\Gamma = \Gamma_g \cup  \Gamma_h$. $f(x)$ is a given function. $\bm{\theta}$ is the parameter vector for the solution, $\mathbf{x}$ is the position vector on the spatial domain $\Omega \subseteq \mathcal{R}^{n}$ bounded by the boundary $\Gamma$, where $n=2$ for the experiments conducted in this study. $u$ is the response of the physical system, and $g$ and $h$ are the values prescribed for the Dirichlet and Neumann boundary conditions. 

To train the PINN model to solve problems in the field of computational solid mechanics, there are two approaches to formulate the loss function: the collocation loss function and the energy-based loss function. The collocation loss function \cite{haghighat2021physics} is based on the strong form \cite{liu2013finite} of the problem, which mathematically enforces the residuals calculated at the collocation points to be zero. Thus, the collocation loss function consists of non-negative residuals of PDEs defined over the domain $\Omega$ and the residuals of the boundary conditions defined on the boundary $\Gamma$. The residuals can be calculated as follows: 

\begin{equation}\label{eqn:l2-redidual}
\begin{aligned}
r_\mathcal{L} (  \bm{\theta}  ) &=\int_{\Omega }( \mathcal{L} (\bm{x}; \bm{\theta} ) - f(\bm{x}))^2  d\! \bm{x},   \\
r_\mathcal{B} (  \bm{\theta}  )&=\int_{ {\Gamma_g}} ( \mathcal{B}_g (  \bm{x}; \bm{\theta} ) -g)^2  d\!\bm{x} + \int_{ { \Gamma_h}} ( \mathcal{B}_h (  \bm{x}; \bm{\theta} -h) )^2  d\!\bm{x}.   
\end{aligned}
\end{equation}
The optimal  parameters $\bm{\theta^*}$ of the model can be calculated according to 
\begin{equation} \label{loss}
\bm{\theta^*}=\underset{{ \bm{\theta} }}{\operatorname{argmin}}\, \underbrace{r_\mathcal{N}(  \bm{\theta} ) +\beta r_\mathcal{B}  (  \bm{\theta}   )}_{J(  \bm{\theta} )},
\end{equation}
where, $\beta$ controls the weight of boundary condition loss, which is a hyper-parameter tuned during the training. The loss function is calculated for the sample of collocation points during each iteration of training and it can be computed as 
\begin{equation} \label{loss-approximate}
\begin{split}
& J(\bm{\theta})  \approx  \frac{1}{m}\sum_{j \in M^{(i)}} J(\bm \theta; \bm x_j ) = \frac{1}{m}\sum_{j \in M^{(i)}} 
\bigg[ \left[\mathcal{L}\left( \bm{x}_j; \hat{u}(\bm{x}_j;\bm{\theta} ) -f(\bm x)\right) \right]^2  \\ &+
\beta_1 \left[\mathcal{B}_g \left(\underbar{$\bm{x}$}^g_j ; \hat{u}(\underbar{$\bm{x}$}^g_j ;\bm{\theta} ) -g \right) \right]^2 +
\beta_2 \left[\mathcal{B}_h \left(\underbar{$\bm{x}$}^h_j ; \hat{u}(\underbar{$\bm{x}$}^h_j ;\bm{\theta} ) -h  \right) \right]^2 \bigg],
\end{split}
\end{equation}
where $M^{(i)}$ is the set of indices of  selected collocation points at iteration $i$ with $|M^{(i)}|=m$, $J(\bm \theta; \bm x_j )$ is the per-sample loss evaluated at the $j$th collocation point,  $\{\underbar{$\bm{x}$}^g_j\}$ denotes the Dirichlet boundary collocation points and  $\{\underbar{$\bm{x}$}^h_j\}$ denotes the Dirichlet boundary collocation points. The model parameters are updated according to
\begin{equation} \label{descent step}
\bm{\theta}^{(i+1)} = \bm{\theta}^{(i)} - \eta^{(i)} \nabla_{\bm{\theta}}{J}(\bm{\theta}^{(i)}),
\end{equation}
where $\nabla_{\bm{\theta}} J$ is calculated using backpropagation \cite{baydin2018automatic}. This is the most popular PINN loss function due to its simplicity in implementation. However, collocation loss function has the disadvantage of possibly satisfying only large-scale PDE residuals in the domain, while the small-scale residuals on boundaries being numerically ignored \cite{kendall2018multi}, owing to the different scales or orders of the loss function terms. However,  this can be improved by balancing the loss terms using adaptive learning schemes like adjustable training weights \cite{liu2019multi} and annealing learning rate \cite{wang2020understanding}. 

The second approach to formulate the loss function to train PINN models for mechanics problems is the energy-based loss function \cite{samaniego2020energy}, which is based on the principle of minimum potential energy. This principle states that the overall potential energy should reach its minimum value when systems are at equilibrium states \cite{lanczos2012variational}. The equilibrium equation, under the small deformation assumption, can be written as:


\begin{equation}
\label{eqb_eqn}
\nabla \cdot \boldsymbol{\sigma} + f = \mathbf{0} \quad \forall \mathbf{x} \in \Omega,
\end{equation}

subject to the boundary conditions:

\begin{equation}
\label{3d_dirichlet}
\mathbf{u} = g \quad \text{on } \Gamma_g,
\end{equation}

\begin{equation}
\label{3d_neumann}
\boldsymbol{\sigma} \cdot \mathbf{n} = h \quad \text{on } \Gamma_h,
\end{equation}

where, $\mathbf{u}$ is the displacement vector, $\boldsymbol{\sigma} = \mathbf{C} :\boldsymbol{\varepsilon}$ is the stress tensor, $\boldsymbol{\varepsilon} = \frac{1}{2} \left( \nabla \mathbf{u} + (\nabla \mathbf{u})^\top \right)$ is the strain tensor, $f$ is the body force vector, $h$ is the traction vector on $\Gamma_h$, $g$ is the prescribed displacement on $\Gamma_g$, and $\mathbf{C}$ is the fourth-order stiffness tensor. Here, Eq.~\eqref{eqb_eqn} is the governing PDE, $\mathcal{L}$ and Eq.~\eqref{3d_dirichlet} and Eq.~\eqref{3d_neumann} are the Dirichlet and Neumann boundary conditions, $\mathcal{B}_g$ and $\mathcal{B}_h$, respectively. Thus, the overall potential energy, $\mathcal{E}$, of the solid system can be written as,

\begin{equation}
\label{energy_loss_eqn}
\mathcal{E}(\mathbf{u}(\bm x;\bm \theta)) = \int_\Omega \left( \frac{1}{2} \boldsymbol{\varepsilon}(\mathbf{u}(\bm x;\bm \theta)) : \mathbf{C} : \boldsymbol{\varepsilon}(\mathbf{u}(\bm x;\bm \theta)) - f \cdot \mathbf{u}(\bm x;\bm \theta) \right) \, d\Omega - \int_{\Gamma_h} h \cdot \mathbf{u}(\bm x;\bm \theta) \, d\Gamma.
\end{equation}
Here, the first term represents the internal energy of the system, while the second term represents the total external energy. Thus, the total potential energy is the sum of the internal and external energy of the system. The PINN model is trained to determine the optimal network parameters, $\bm \theta$, that minimize this potential energy. In this study, we use this energy formulation as the loss function for training both the benchmark and proposed PINN models across all experiments.

\section{Method}
\label{methodology}

In this section, we outline the PINN-FEM methodology for combining Physics-Informed Neural Networks (PINNs) with finite element methods (FEM) to enforce strong Dirichlet boundary conditions for mechanics problems. The proposed approach integrates the strengths of both methods by utilizing a local finite element mesh near the Dirichlet boundary while using PINNs to evaluate the approximate solution in the interior domain. This hybrid method leverages the minimum potential energy formulation to provide a stable and efficient framework for boundary condition enforcement. The following subsections detail the high-level structure of the proposed hybrid method, problem formulation, and the loss functions used for training the models for one-, two-, and three-dimensional boundary value problems.

\subsection{Formulation of PINN-FEM Method}

For a given problem, the computational domain, $\Omega$, is divided into two regions: a boundary region where FEM is applied (denoted as $\Omega_\text{FE}$) and an interior region where PINNs are used (denoted as $\Omega_\text{NN}$). The boundary region includes the Dirichlet boundary and a layer of elements adjacent to it, whereas the rest of the region uses PINN for approximate solution. Fig. \ref{fig.1dBVPdirichlet_oneside} and Fig. \ref{fig.1dBVPdirichlet} show the domain decomposition $\Omega_{FE}$ and $\Omega_{NN}$ for a one-dimensional boundary value problem with Dirichlet BC at one and both ends respectively. Fig. \ref{fig.3dBVPdirichlet} shows the domain decomposition for a two-dimensional boundary value problem with Dirichlet and Neumann boundary conditions. 

In the Dirichlet boundary region, $\Omega_{FE}$, we use finite element method to enforce the Dirichlet boundary conditions. This is done by constructing a finite element mesh that conforms to the geometry of the boundary, where Dirichlet BC is applied. Standard FEM techniques are applied to approximate the solution in this region.  This concept is similar to using approximate distance functions for exact imposition of boundary conditions \cite{sukumar2022exact} in PINNs. Here, the finite element mesh acts as an approximation of the distance function, thus ensuring accurate enforcement of the essential boundary conditions. For the rest of the domain (interior region and boundary region where non-essential BCs are applied), $\Omega_{NN}$, PINN model is used to employ the approximate solution of the PDE. The architecture and hyperparameters for PINN are chosen based on the specific PDE and domain characteristics.

For training the PINN, we use the energy-based loss function based on the variational principle. The neural network is trained to minimize the total potential energy which is the sum of total internal energy and total external energy as given in Eq.~\eqref{energy_loss_eqn}. The potential energy in the essential boundary region, where FE mesh is used, is approximated using finite element basis functions. The finite element method provides the solution $\bm{u}_h$ that minimizes the potential energy in the boundary region. In the interior region, the solution is approximated by a neural network $\bm u(\bm \theta)$. The network is trained to minimize the potential energy in the interior domain, $\mathcal{E}(\bm u(\bm \theta)$. The function space of PINN network is extended to seamlessly integrate with the FEM solution at the boundary in this approach. The continuity and compatibility between the FEM solution in the boundary region and the PINN solution in the interior domain is ensured by aligning the finite element basis functions with the neural network's output at the interface. 

The solution formulation for this implementation is detailed in the following Sections \ref{sec_1d} and \ref{sec_3d} for 1D, 2D and 3D boundary value problems. This hybrid method extends the applicability of PINNs to a wider range of industrial applications by effectively combining the strengths of finite element methods and neural networks. The use of the minimum potential energy formulation provides a robust and efficient framework for solving complex PDEs with strong boundary conditions.

\subsection{One-dimensional boundary value problem}
\label{sec_1d}

\subsubsection{Dirichlet BC at one end}
Without loss of generality, let us consider a one-dimensional boundary value problem (BVP) over the domain $\Omega$ having complementary  Neumann  $\Gamma_{h}$ and Dirichlet boundary $\Gamma_{g}$ with respect to boundary $\Gamma$ of the domain $\Omega$. The model problem is formulated as follows.

\begin{equation}
u_{,xx}+f=0 \quad \forall x \in (0,1), \quad
\label{eqn:bvp_form_1}
\end{equation}

\begin{equation}
{-u}_{,x}(0) = h, \quad
\label{eqn:bvp_form_3}
\end{equation}

\begin{equation}
u(1) = g.
\label{eqn:bvp_form_2}
\end{equation}

The differential operator, $\mathcal{L}$, in Eq.~\eqref{eq_bvp}, for this problem, is given by Eq.~\eqref{eqn:bvp_form_1} and the corresponding boundary condition operators, $\mathcal{B}_h$ and $\mathcal{B}_g$ are given by the equations, Eq.~\eqref{eqn:bvp_form_3} and Eq.~\eqref{eqn:bvp_form_2} respectively. Here, $\Omega = (0,1)$, $\Gamma =\{0,1\}$, $\Gamma_g = \{1\}$, and $\Gamma_h =\{0\}$, $ f: \Omega\to R $, $ h: \Omega\to R $. It is to be noted that, the above equations are obtained by considering the one-dimensional case of the equilibrium equation defined in Eq.\eqref{eqb_eqn}. We can weaken the above PDE, by reducing the order of the spatial derivatives, using variational function and trial solution from appropriate function spaces. The corresponding general form of the energy functional of this one-dimensional BVP is given by:

\begin{equation}
\mathcal{E}(u) = \int_0^1 \left( \frac{1}{2} u_x^2 - f u \right) \, dx  + h u(0).
\label{eqn:bvp_form_1d_oneside}
\end{equation}

It is to be noted that, with the proper selection of the solution space,  $ S = \{u| u\in H^1, u(1)=g \}$,  the Dirichlet boundary conditions can be automatically satisfied and hence, they do not appear explicitly in the energy formulation. 

\begin{figure}[!ht]
\vskip 0.2in
     \centering
     \begin{subfigure}[b]{0.8\textwidth}
         \centering
         \includegraphics[width=\textwidth]{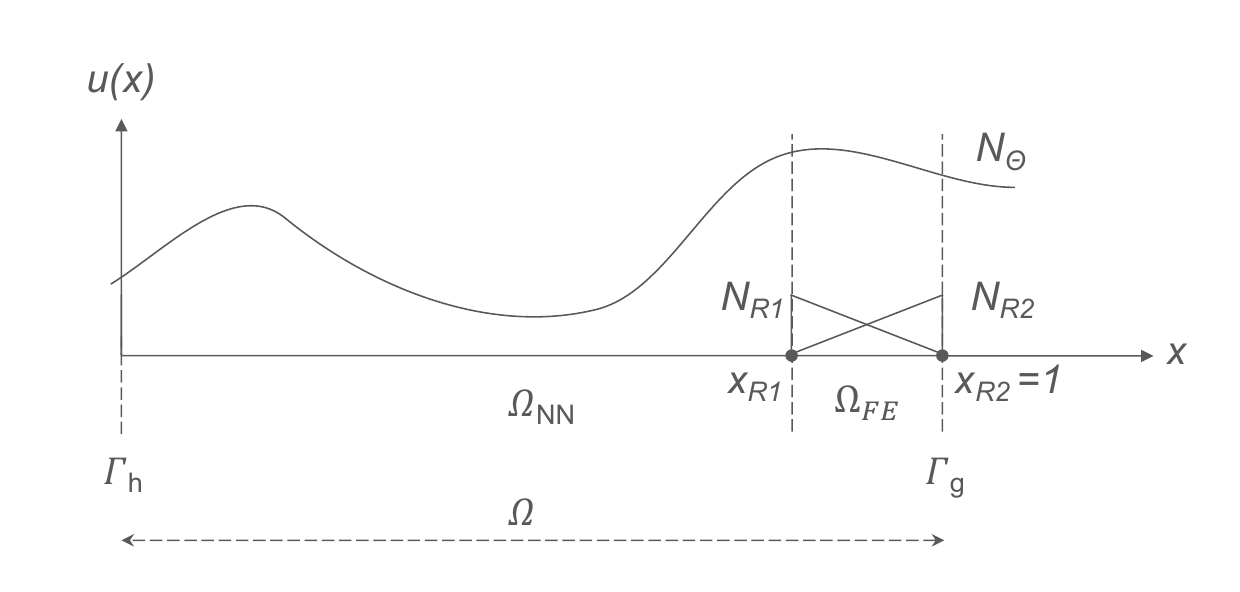}
         \caption{Domain Decomposition into neural network domain, $\Omega_\text{NN}$ and finite element domain, $\Omega_\text{FE}$ with common interface at $x_{R_1}$. The shape functions, $N_{R_1}$ and  $N_{R_2}$ for the finite element are also shown. $\Gamma_h$ is the Neumann boundary and $\Gamma_g$ is the Dirichlet boundary. }
         \label{fig:1d_bvp_domain_oneside}
     \end{subfigure}
     \hfill
     \begin{subfigure}[b]{0.8\textwidth}
         \centering
         \includegraphics[width=\textwidth]{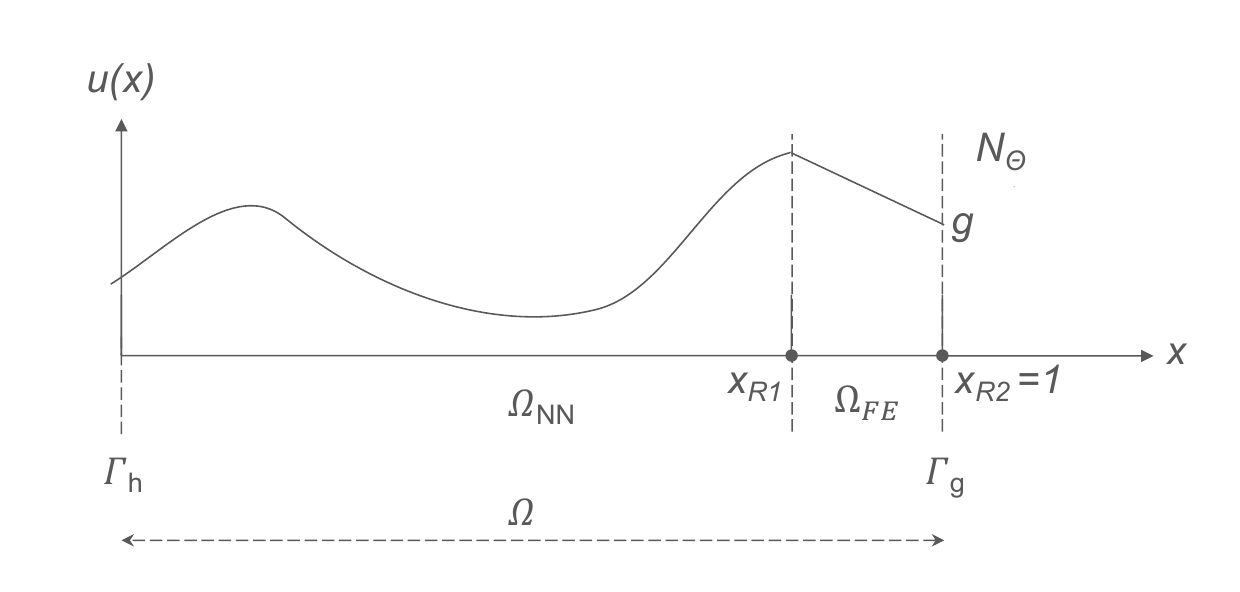}
         \caption{Exact imposition of prescribed displacement, $g$, at the Dirichlet boundary using the PINN-FEM method.}
         \label{fig:1d_bvp_mesh_oneside}
     \end{subfigure}
        \caption{Domain Decomposition with complementary Neumann and Dirichlet boundary conditions at the two ends for a one-dimensional boundary value problem and the corresponding exact BC imposition using the proposed PINN-FEM method.}
        \label{fig.1dBVPdirichlet_oneside}
\vskip -0.2in
\end{figure}

We decompose the domain $\Omega$, $(0,1)$ into two non-overlapping domains - a finite element domain ($\Omega_{\text{FE}}$), which is the boundary region for Dirichlet BC and neural network domain ($\Omega_{\text{NN}}$), which is the interior region as shown in Fig. ~\ref{fig.1dBVPdirichlet_oneside}. The interface between the two domains is denoted by ($\Gamma_{\text{I}}$). The finite element method is applied to the boundary region to enforce the Dirichlet conditions, while the neural network (PINN network) approximates the solution in the interior region. The finite element domain has two boundaries, the Dirichlet boundary and the interface, while the neural network domain contains the remaining boundary conditions (other than the Dirichlet boundary) in addition to the interface. The finite element domain contains only a single layer of elements. For the 1D BVP problem, this single layer consists of a single element. We will further assume that the finite element is a two-node linear element. With these assumptions,  the displacement field,  based on finite element and neural network, is given by:
\begin{equation}
u(x;\bm \theta) = 
    \begin{cases}
        {N}_{\bm \theta}( x)        , & x \in \Omega_{\text{NN}}   \\
                 {N}_{\bm \theta}(x_{R_1}) {N}_{R_1}(x)+   g { N}_{R_2}(x), & x   \in \Omega_{FE} 
    \end{cases}
\label{eqn:disp_form_oneside}
\end{equation}
Here, ${N}_{\bm \theta}$ is the PINN output with neural network parameter. $x_{R_1}$ and $x_{R_2}$ are the nodal points for the finite element and ${N}_{R_1}$ and ${N}_{R_2}$ are the corresponding shape functions for the finite element, as illustrated in Fig. \ref{fig.1dBVPdirichlet_oneside}. The displacement gradients are obtained, in a straightforward manner, by differentiating the finite element and neural network displacement fields:
\begin{equation}
u(x; \bm \theta)_{,x} = 
    \begin{cases}
        {N}_{\bm \theta,x} (x)        , & x  \in \Omega_{\text{NN}}  \\
        N_{\bm \theta,x}|_{x_{R_1}}(x) {N}_{R_1}(x)+  {N}_{\bm \theta}|_{x_{R_1}}(x)   {N}_{R_1,x}(x) + g {N}_{R_2,x}(x)  , & x  \in \Omega_{FE} \\
    \end{cases}
\end{equation}
It is to be noted that the nodal values of the finite element domain are all specified. At the interface, the nodal value is the same as that of the neural network, while at the Dirichlet boundary it is the prescribed displacement. Now, Eq.~\eqref{eqn:bvp_form_1d_oneside} can be formulated as:
\begin{equation}
\mathcal{E}(u) = \int_0^{x_{R_1}}  \frac{1}{2} u_x^2  \, dx  +  \int_{x_{R_1}}^{x_{R_2}}  \frac{1}{2} u_x^2  \, dx + h u(0),
\end{equation}
where, $x_{R_1}$ is the right end for the neural network domain, $\Omega_{\text{NN}}$, where $\Omega_{\text{FE}}$ begins. $x_{R_2}=1$ is the right end of $\Omega_{\text{FE}}$ as shown in Fig \ref{fig.1dBVPdirichlet_oneside}. By substituting the expressions for $u$ ad $u_x$ into this, we get,
\begin{equation}
\begin{aligned}
\mathcal{E}(\bm \theta) &= \int_{0}^{x_{R_1}}  \frac{1}{2} \left({N}_{\bm \theta,x}\right)^2  \, dx \\
&+ \int_{x_{R_1}}^{x_{R_2}}   \frac{1}{2} \left({N}_{\bm \theta,x}|_{x_{R_1}} {N}_{R_1}+  {N}_{\bm \theta}|_{x_{R_1}}   {N}_{R_1,x} + g_{R_2} {N}_{R_2,x}\right)^2 \, dx \\
& + h {N}_{\bm \theta,x}|_{0}
\end{aligned}
\end{equation}
The energy content in the neural network domain, , $\Omega_{\text{NN}}$, is evaluated by discretizing the domain as follows:
\begin{equation}
\mathcal{E}_{NN}(\bm \theta) = \int_{0}^{x_{R_1}}  \frac{1}{2} ({N}_{\bm \theta,x} )^2  \, dx  = \sum_{x_i \in M} \frac{1}{2} ({N}_{\bm \theta,x_i} )^2 \Delta x_i,
\end{equation}
where, $M$ is the set of all the points in the discretized domain. The energy content in the finite element domain is evaluated using 1-point Gauss quadrature as follows:
\begin{equation}
\mathcal{E}_{FE}(\bm \theta) =    \frac{1}{2} ({N}_{\bm \theta,x}|_{x_{R_1}} {N}_{R_1}+  {N}_{\bm \theta}|_{x_{R_1}}   {N}_{R_1,x} + g_{R_2} {N}_{R_2,x})^2  |_{x_c} (x_{R_2}-x_{R_1}),
\end{equation}
where, $x_c$ is the center of the FEM element. Finally, the energy done by the external force is calculated using the traction, $h$, at the Neumann boundary as follows:
\begin{equation}
\mathcal{E}_h(\bm \theta) =  h {N}_{\bm \theta,x}|_{0}
\end{equation}
Thus, the total potential energy loss function for training the proposed PINN model is given as:
\begin{equation}
\mathcal{E}(\bm \theta) = \mathcal{E}_{NN}(\bm \theta) + \mathcal{E}_{FE}(\bm \theta) +\mathcal{E}_h(\bm \theta)
\end{equation}
This energy loss function is, finally, minimized using automatic differentiation, to determine the neural network parameters, $\bm \theta$. 

\begin{figure}[!ht]
\vskip 0.2in
     \centering
     \begin{subfigure}[b]{0.8\textwidth}
         \centering
         \includegraphics[width=\textwidth]{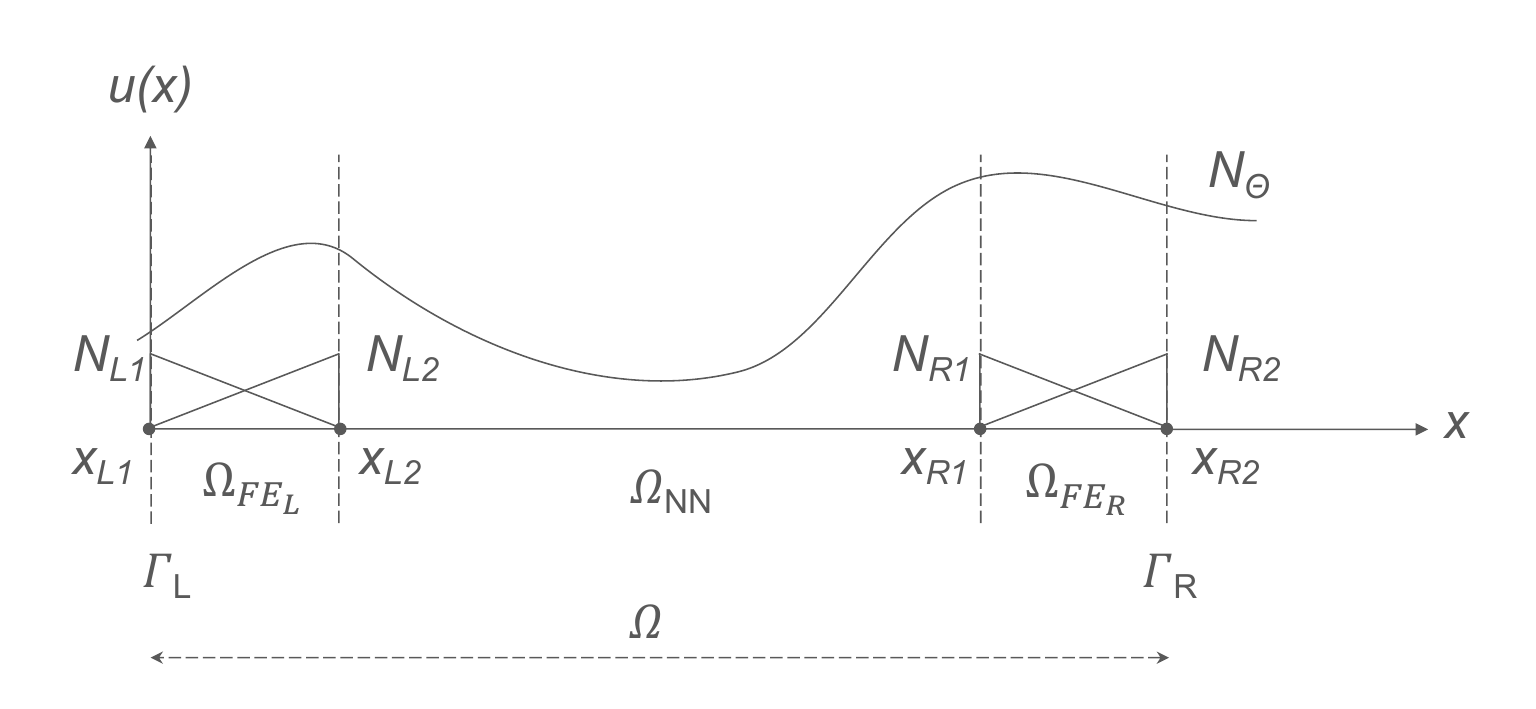}
         \caption{Domain Decomposition into neural network domain, $\Omega_\text{NN}$ and finite element domain, $\Omega_{\text{FE}}$. Here, $\Omega_{\text{FE}}$ consists of two regions, $\Omega_{\text{FE}_L}$ and $\Omega_{\text{FE}_R}$ with common interface with $\Omega_\text{NN}$ at $x_{L_2}$ and $x_{R_1}$ respectively. $N_{L_1}$ and  $N_{L_2}$ are the shape functions for the left finite element domain $\Omega_{\text{FE}_L}$. $N_{R_1}$ and  $N_{R_2}$ are the shape functions for the right finite element domain $\Omega_{\text{FE}_R}$.}
         \label{fig:1d_bvp_domain}
     \end{subfigure}
     \hfill
     \begin{subfigure}[b]{0.8\textwidth}
         \centering
         \includegraphics[width=\textwidth]{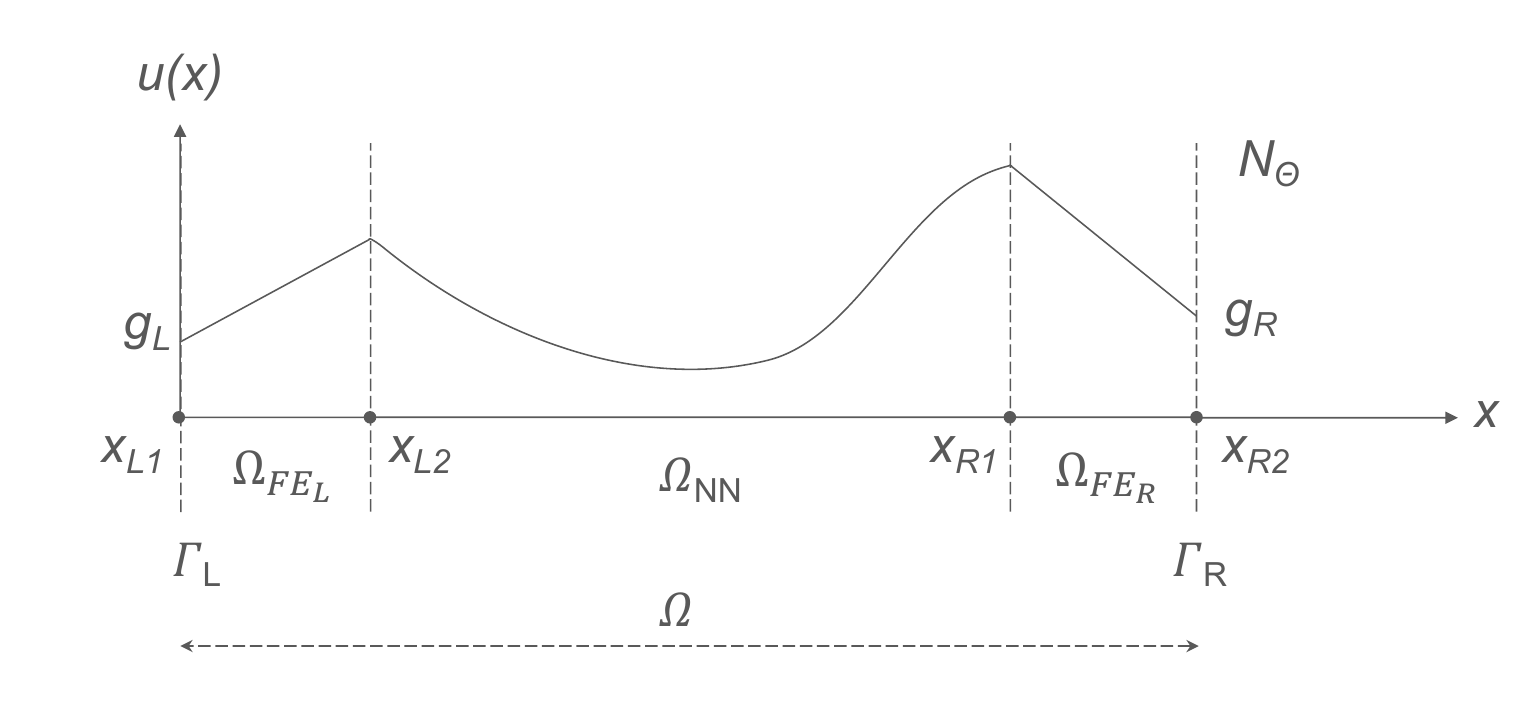}
         \caption{Exact imposition of prescribed displacement, $g_L$, at the left Dirichlet boundary and  $g_R$, at the right Dirichlet boundary using the PINN-FEM method. }
         \label{fig:1d_bvp_mesh}
     \end{subfigure}
        \caption{Domain Decomposition with Dirichlet boundary conditions at both the ends for a one-dimensional boundary value problem and the corresponding exact BC imposition using the proposed PINN-FEM method.}
        \label{fig.1dBVPdirichlet}
\vskip -0.2in
\end{figure}

\subsubsection{Dirichlet BC at both the ends}
Now, let us consider a one-dimensional BVP over the domain $\Omega=(0,1)$ with Dirichlet boundary, $\Gamma_g$ at both the ends, $x=0$ and $x=1$, as shown in Fig. \ref{fig.1dBVPdirichlet}. Thus, the model problem is formulated as:

\begin{equation}
u_{,xx}+f=0 \quad \forall x \in (0,1), \quad
\label{eqn:bvp_form_1_2sides}
\end{equation}

\begin{equation}
u(0) = g_L, \quad u(1) = g_R.
\label{eqn:bvp_form_2_2sides}
\end{equation}
Here, $g_L$ and $g_R$ are the prescribed displacement values at the left and right boundaries respectively. Now, the energy functional for this one-dimensional BVP can be written as:
\begin{equation}
\mathcal{E}(u) = \int_0^1 \left( \frac{1}{2} u_x^2 - f u \right) \, dx,
\label{eqn:bvp_form_1d}
\end{equation}
with Dirichlet BC automatically satisfied with the proper selection of the solution space, $ S = \{u| u\in H^1, u(0)=g_L, u(1)=g_R \}$. The displacement field, based on finite element and neural network, for this problem, is given by:

\begin{equation}
u(x;\bm \theta) = 
    \begin{cases}
         {N}_{\bm \theta}({x_{L_2}}) {N}_{L_2}(x)+   g_L { N}_{L_1}(x), & x   \in \Omega_{FE_L}  \\
        {N}_{\bm \theta}( x)        , & x \in \Omega_{\text{NN}}   \\
        {N}_{\bm \theta}(x_{R_1}) {N}_{R_1}(x)+   g_R { N}_{R_2}(x), & x   \in \Omega_{FE_R} 
    \end{cases}
\label{eqn:disp_form}
\end{equation}
where, $x_{L_1}$ and $x_{L_2}$ are the nodal points for the left finite element and $x_{R_1}$ and $x_{R_2}$ are the nodal points for the right finite element. ${N}_{L_1}$, ${N}_{L_2}$, ${N}_{R_1}$ and ${N}_{R_2}$ are the corresponding shape functions for the finite elements, as illustrated in Fig. \ref{fig.1dBVPdirichlet}. Similarly, the displacement gradients are given by:
\begin{equation}
u(x; \bm \theta)_{,x} = 
    \begin{cases}
        N_{\bm \theta,x}|_{x_{L_2}} {N}_{L_2}(x)+  {N}_{\bm \theta}|_{x_{L_2}}   {N}_{L_2,x} + g_{L} {N}_{L_2,x}  , & x  \in \Omega_{FE_L} \\
        {N}_{\bm \theta,x}         , & x  \in \Omega_{\text{NN}}  \\
        N_{\bm \theta,x}|_{x_{R_1}} {N}_{R_1}(x)+  {N}_{\bm \theta}|_{x_{R_1}}   {N}_{R_1,x} + g_{R} {N}_{R_2,x}  , & x  \in \Omega_{FE_R} \\
    \end{cases}
\end{equation}
Now, Eq.~\eqref{eqn:bvp_form_1d} can be formulated as:
\begin{equation}
\mathcal{E}(u) = \int_{x_{L_1}}^{x_{L_2}}  \frac{1}{2} u_x^2  \, dx  + \int_{x_{L_2}}^{x_{R_1}}  \frac{1}{2} u_x^2  \, dx +  \int_{x_{R_1}}^{x_{R_2}}  \frac{1}{2} u_x^2  \, dx,
\end{equation}
where, $x_{L_2}$ and $x_{R_1}$ are the left and right ends for the neural network domain, $\Omega_{\text{NN}}$ respectively. $x_{L_1}=0$ and $x_{R_2}=1$ for this problem. By substituting the expressions for $u$ ad $u_x$ into this, we get,
\begin{equation}
\begin{aligned}
\mathcal{E}(\bm \theta) &= \int_{x_{L_2}}^{x_{R_1}}  \frac{1}{2} \left({N}_{\bm \theta,x}\right)^2  \, dx \\
&+ \int_{x_{L_1}}^{x_{L_2}}   \frac{1}{2} \left({N}_{\bm \theta,x}|_{x_{L_1}} {N}_{L_1}+  {N}_{\bm \theta}|_{x_{L_1}}   {N}_{L_1,x} + g_{L_2} {N}_{L_2,x}\right)^2 \, dx \\
&+ \int_{x_{R_1}}^{x_{R_2}}   \frac{1}{2} \left({N}_{\bm \theta,x}|_{x_{R_1}} {N}_{R_1}+  {N}_{\bm \theta}|_{x_{R_1}}   {N}_{R_1,x} + g_{R_2} {N}_{R_2,x}\right)^2 \, dx \\
\end{aligned}
\end{equation}
Furthermore, the energy content in the neural network domain, , $\Omega_{\text{NN}}$, is evaluated as:
\begin{equation}
\mathcal{E}_{NN}(\bm \theta) = \int_{x_{L_2}}^{x_{R_1}}  \frac{1}{2} ({N}_{\bm \theta,x} )^2  \, dx  = \sum_{x_i \in M} \frac{1}{2} ({N}_{\bm \theta,x_i} )^2 \Delta x_i,
\end{equation}
and the energy content in the left and right finite element domains are evaluated using 1-point Gauss quadrature as follows:
\begin{equation}
\mathcal{E}_{FE_L}(\bm \theta) =    \frac{1}{2} ({N}_{\bm \theta,x}|_{x_{L_1}} {N}_{L_1}+  {N}_{\bm \theta}|_{x_{L_1}}   {N}_{L_1,x} + g_{L_2} {N}_{L_2,x})^2  |_{x_c} (x_{L_2}-x_{L_1})
\end{equation}

\begin{equation}
\mathcal{E}_{FE_R}(\bm \theta) =    \frac{1}{2} ({N}_{\bm \theta,x}|_{x_{R_1}} {N}_{R_1}+  {N}_{\bm \theta}|_{x_{R_1}}   {N}_{R_1,x} + g_{R_2} {N}_{R_2,x})^2  |_{x_c} (x_{R_2}-x_{R_1})
\end{equation}
Finally, the total potential energy loss function for training the proposed PINN model for this one-dimensional BVP is given as:

\begin{equation}
\mathcal{E}(\bm \theta) = \mathcal{E}_{NN}(\bm \theta) + \mathcal{E}_{FE_L}(\bm \theta) + \mathcal{E}_{FE_R}(\bm \theta)
\end{equation}

\subsection{Two- and three-dimensional boundary value problems}
\label{sec_3d}

\begin{figure}[!ht]
\vskip 0.2in
     \centering
     \begin{subfigure}[b]{0.80\textwidth}
         \centering
         \includegraphics[width=\textwidth]{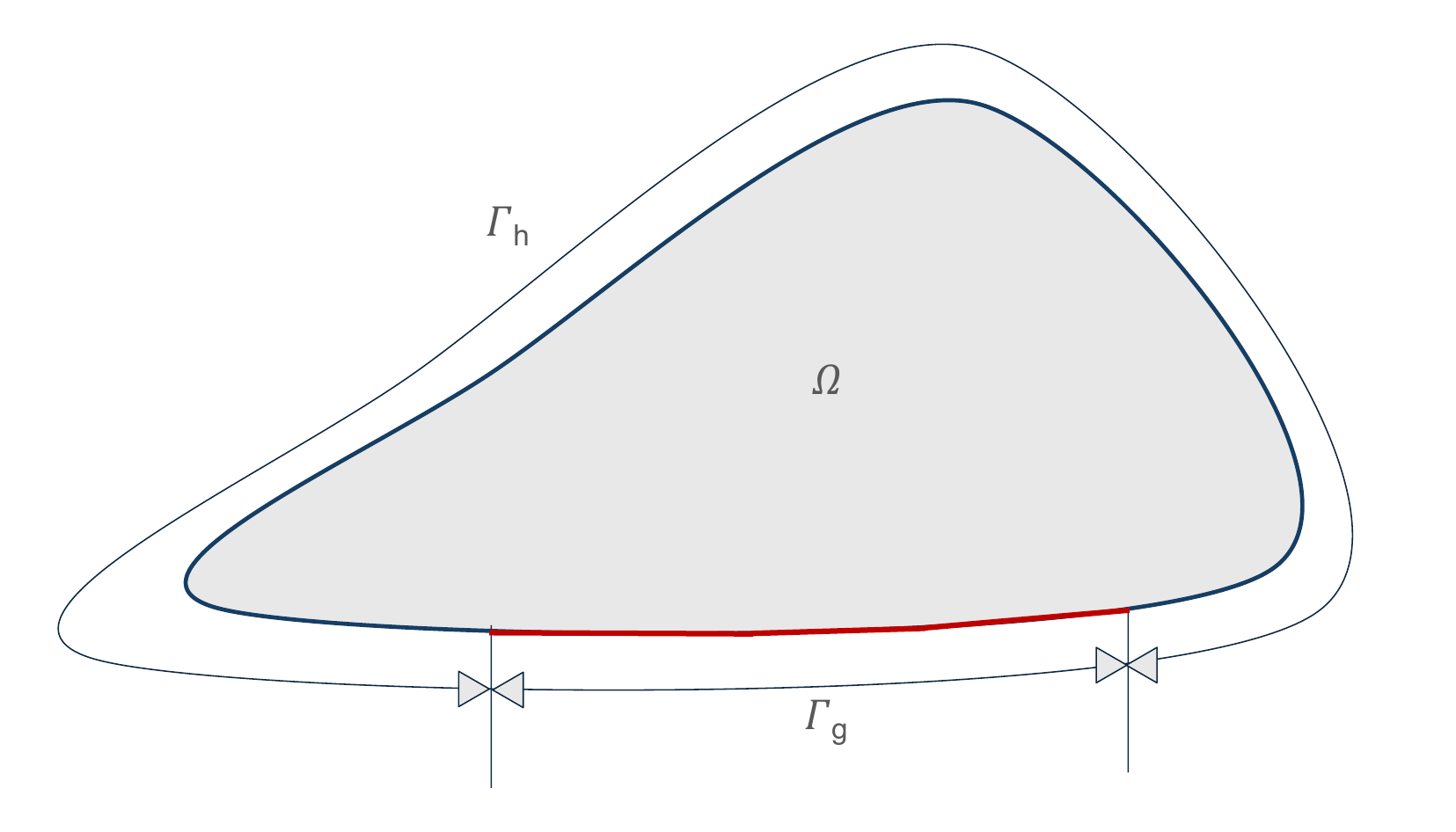}
         \caption{The domain, $\Omega$ and the Dirichlet and Neumann boundary, $\Gamma_g$ and $\Gamma_h$, respectively for a two-dimensional boundary value problem.}
         \label{fig:2d_bvp_domain}
     \end{subfigure}
     \hfill
     \begin{subfigure}[b]{0.80\textwidth}
         \centering
         \includegraphics[width=\textwidth]{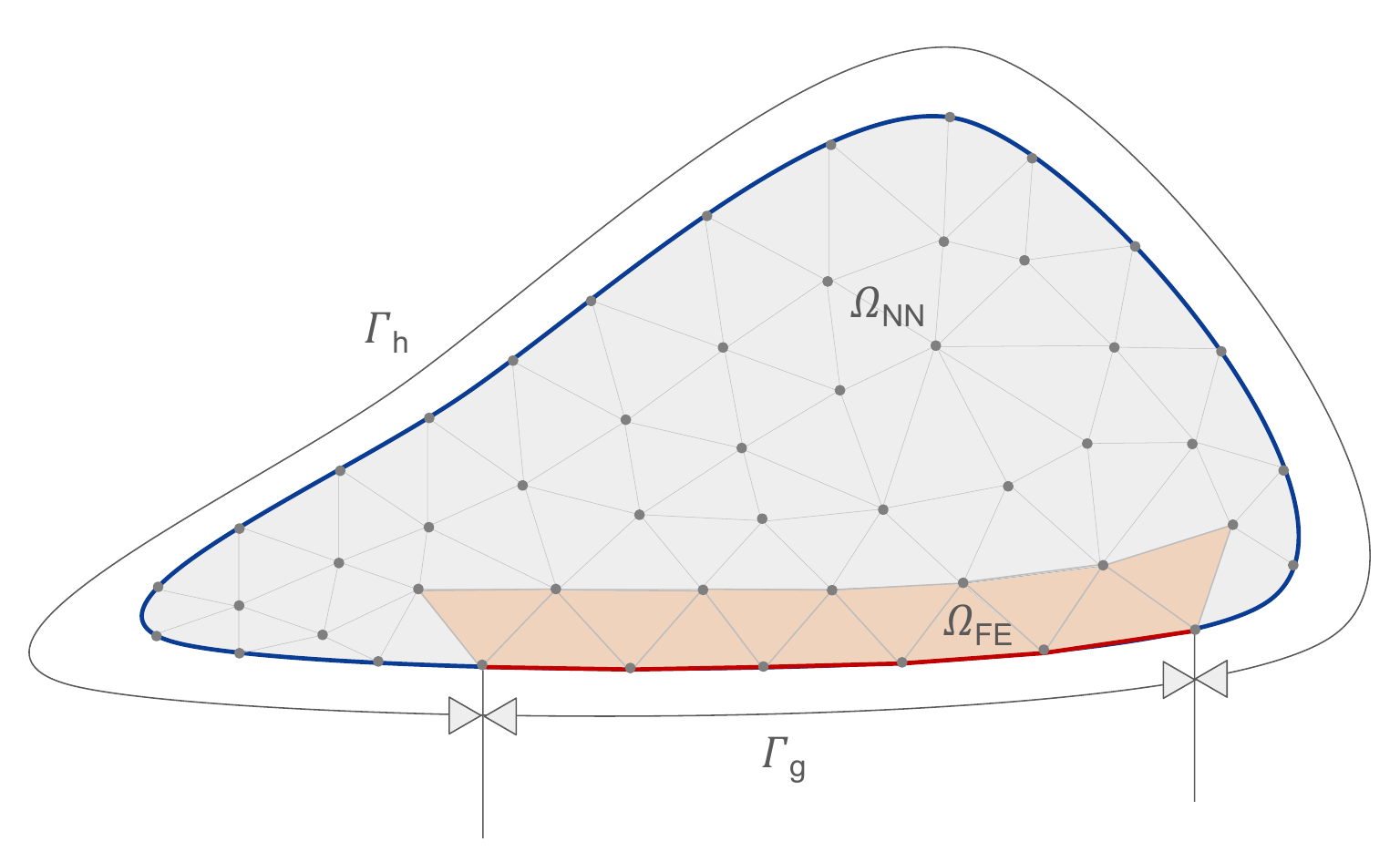}
         \caption{Domain decomposition into finite element domain, $\Omega_\text{FE}$ and neural network domain, $\Omega_\text{NN}$.  $\Omega_\text{FE}$ consists of the finite elements corresponding to the Dirichlet boundary, whereas, $\Omega_\text{NN}$ consists of the elements corresponding to Neumann boundary and interior domain.}
         \label{fig:2d_bvp_mesh}
     \end{subfigure}
        \caption{Domain decomposition for the proposed PINN-FEM method for a two-dimensional boundary value problem.}
        \label{fig.3dBVPdirichlet}
\vskip -0.2in
\end{figure}

Extending the one-dimensional formulation presented earlier, let us now consider a two-dimensional boundary value problem (BVP) over the domain $\Omega$. The boundary $\Gamma$ is partitioned into complementary Dirichlet boundary $\Gamma_g$ and Neumann boundary $\Gamma_h$, as shown in Fig. \ref{fig.3dBVPdirichlet}(a). The governing equations for linear elasticity and the Dirichlet and Neumann boundary conditions for this problem are given in Eq.~\eqref{eqb_eqn}, Eq.~\eqref{3d_dirichlet} and Eq.~\eqref{3d_neumann} respectively. Here, $\mathbf{u} = [u_x, u_y]^\top$ is the displacement vector for the two-dimensional problem. Similar to the one-dimensional case, we decompose the domain $\Omega$ into two non-overlapping regions as shown in Fig. \ref{fig.3dBVPdirichlet}(b):
\begin{itemize}
    \item A finite element domain ($\Omega_{\text{FE}}$), which enforces the Dirichlet boundary conditions on $\Gamma_g$,
    \item A neural network domain ($\Omega_{\text{NN}}$), which approximates the solution in the interior region and on $\Gamma_h$.
\end{itemize}

Now, the displacement field, based on finite element and neural network, $\mathbf{u}(\mathbf{x};\bm{\theta})$, is defined as:
\begin{equation}
\mathbf{u}(\mathbf{x};\bm{\theta}) =
    \begin{cases}
        \mathbf{N}_{\bm \theta}(\mathbf{x}) \mathbf{N}_D(\mathbf{x}) + g, & \mathbf{x} \in \Omega_{\text{FE}}, \\
        \mathbf{N}_{\bm \theta}(\mathbf{x}), & \mathbf{x} \in \Omega_{\text{NN}} \cup \Gamma_h,
    \end{cases}
\label{eqn:disp_field_3d}
\end{equation}
where, $\mathbf{N}_\theta$ is the neural network output parameterized by $\theta$, $\mathbf{N}_D$ is the FEM shape functions for $\Omega_{\text{FE}}$, and $g$ is the prescribed displacement on $\Gamma_g$. The energy functional for the two-dimensional BVP is given by:
\begin{equation}
\label{energy_loss_2d_bvp}
\mathcal{E}(\mathbf{u}(\mathbf{x};\bm{\theta})) = \int_\Omega \left( \frac{1}{2} \boldsymbol{\varepsilon}(\mathbf{u}(\bm x;\bm \theta)) : \mathbf{C} : \boldsymbol{\varepsilon}(\mathbf{u}(\bm x;\bm \theta)) - f \cdot \mathbf{u(\mathbf{x};\bm{\theta})} \right) \, d\Omega - \int_{\Gamma_h} h \cdot \mathbf{u}(\mathbf{x};\bm{\theta}) \, d\Gamma.
\end{equation}
This energy is partitioned as:
\begin{equation}
\mathcal{E}(\bm \theta) = \mathcal{E}_{\text{NN}}(\bm \theta) + \mathcal{E}_{\text{FE}}(\bm \theta) + \mathcal{E}_h(\bm \theta),
\end{equation}
where, $\mathcal{E}_{\text{NN}}$ is the energy in the neural network domain, $\mathcal{E}_{\text{FE}}$ is the energy in the Dirichlet FEM region, and $\mathcal{E}_h$ is the work done by traction forces on $\Gamma_h$. The neural network energy, $\mathcal{E}_{\text{NN}}$ is given by, 
\begin{equation}
\mathcal{E}_{\text{NN}}(\bm \theta) = \int_{\Omega_{\text{NN}}} \frac{1}{2} \boldsymbol{\varepsilon}(\mathbf{u}(\bm x;\bm \theta)) : \mathbf{C} : \boldsymbol{\varepsilon}(\mathbf{u}(\bm x;\bm \theta)) \, d\Omega - \int_{\Omega_{\text{NN}}} f \cdot \mathbf{u}(\bm x; \bm \theta) \, d\Omega.
\end{equation}
The energy in the Dirichlet FEM region, $\mathcal{E}_{\text{FE}}$, is given by, 
\begin{equation}
\mathcal{E}_{\text{FE}}(\bm \theta) = \frac{1}{2} \left( \boldsymbol{\varepsilon}(\mathbf{u}(\bm x;\bm \theta)) : \mathbf{C} : \boldsymbol{\varepsilon}(\mathbf{u}(\bm x;\bm \theta)) \right) \big|_{\mathbf{x}_c} A_{\text{FE}},
\end{equation}
where $A_{\text{FE}}$ is the area of the Dirichlet FEM region and $\mathbf{x}_c$ is the center of the FEM element. The traction work energy on $\Gamma_h$, $\mathcal{E}_h$, is given by,
\begin{equation}
\mathcal{E}_h(\bm \theta) = -\int_{\Gamma_h} h \cdot \mathbf{u}(\bm x; \bm \theta) \, d\Gamma.
\end{equation}
Thus, the total potential energy loss function for 2D BVP is given by,
\begin{equation}
\begin{aligned}
\mathcal{E}(\bm \theta) = & \int_{\Omega_{\text{NN}}} \frac{1}{2} \boldsymbol{\varepsilon}(\mathbf{u}(\bm x;\bm \theta)) : \mathbf{C} : \boldsymbol{\varepsilon}(\mathbf{u}(\bm x;\bm \theta)) \, d\Omega - \int_{\Omega_{\text{NN}}} f \cdot \mathbf{u}(\bm x; \bm \theta) \, d\Omega  \\
& + \frac{1}{2} \left( \boldsymbol{\varepsilon}(\mathbf{u}(\bm x;\bm \theta)) : \mathbf{C} : \boldsymbol{\varepsilon}(\mathbf{u}(\bm x;\bm \theta)) \right) \big|_{\mathbf{x}_c} A_{\text{FE}} \\
& - \int_{\Gamma_h} h \cdot \mathbf{u}(\bm x; \bm \theta) \, d\Gamma.
\label{2d_energy_eqn}
\end{aligned}
\end{equation}
This energy loss function is minimized using automatic differentiation to determine the neural network parameters $\bm \theta$.

The energy functional in Eq.~\eqref{eqn:disp_field_3d} can be extended to three-dimensional case by considering the displacement vector,  $\mathbf{u} = [u_x, u_y, u_z]^\top$. The energy in the Dirichlet FEM region for 3D problem, $\mathcal{E}_{\text{FE}}$, is modified as, 
\begin{equation}
\mathcal{E}_{\text{FE}}(\bm \theta) = \frac{1}{2} \left( \boldsymbol{\varepsilon}(\mathbf{u}(\bm x;\bm \theta)) : \mathbf{C} : \boldsymbol{\varepsilon}(\mathbf{u}(\bm x;\bm \theta)) \right) \big|_{\mathbf{x}_c} V_{\text{FE}},
\end{equation}
where $V_{\text{FE}}$ is the volume of the Dirichlet FEM region. All the other equations for energy formulation for 3D BVP remain the same as that of 2D BVP.

\section{Numerical Results}
\label{results}

We demonstrate the performance  of our approach through six experiments in the following sections. For the first five experiments, we consider a two-dimensional square elastic plate of length, $L$, of 1 mm and thickness, $H$, of 1 mm under plane stress conditions.  The domain is thus defined as $\Omega={(x,y) \quad \forall x,y \in (0,1)}$. The origin of the Cartesian coordinate is placed at the lower right-vertex of the square with the $x$ axis pointing to the right. The plate is subjected to a uniform stretching force on its right edge. The Young’s modulus of the elastic plate, $E$, is 70 MPa, and the Poisson’s ratio, $\nu$, is 0.3. We consider different Dirichlet and Neumann boundary conditions for this domain. For the final experiment, we consider a cantilever beam with a width, $W$, of 1 mm and a depth, $D$ of 0.5 mm with point boundary conditions applied on the right edge and parabolic traction applied in the y-direction on the left edge, under plane strain conditions.  The domain is defined as $\Omega={(x,y) \quad \forall x\in (0,1)}, \quad y\in (-0.25, 0.25)$ and the origin of the Cartesian coordinate is placed at the center of the right edge with the $x$ axis pointing to the right. The details of these experiments are summarized in Table \ref{table:exps_doms_bcs}. 

\subsection{Mesh Generation and Collocation Points}
\label{mesh_gen}

We use Gmsh \cite{geuzaine2020three} to generate the finite element mesh for the domain decomposition approach. The finite element domain, $\Omega_\text{FE}$, near the Dirichlet boundary was meshed with triangular elements. A mesh size of 0.1 mm is selected to balance computational efficiency and accuracy in enforcing boundary conditions. The mesh conforms to the geometry of the Dirichlet boundary, ensuring precise enforcement of essential boundary conditions.

The collocation points for the PINN are chosen as the centroids of the triangles in the finite element mesh. This approach extends the collocation points across the entire computational domain $\Omega$, including both the finite element region ($\Omega_\text{FE}$) near the Dirichlet boundary and the neural network region ($\Omega_\text{NN}$) in the interior. Using centroids ensures an even distribution of collocation points and establishes a robust framework for integrating FEM and PINN components at the interface ($\Gamma_I$).

\begin{table}
  \centering
  \caption{A list of the experiments conducted for the study, with the domains and the corresponding equations of the boundary conditions. Here, $u_x$ and $u_y$ are displacements in $x$ and $y$ directions respectively, $\sigma_{xx}$ is the in-plane normal force and $\sigma_{xy}$ is the in-plane tangential force. The Dirichlet boundary for which FEM is applied, $\Omega_\text{FE}$, is highlighted in red. The Neumann boundary, where traction is applied, is highlighted in blue.}
\begin{tabular}{| p{.30\textwidth} | p{.31\textwidth} |p{.29\textwidth}|} 
\toprule
    Experiment  & Domain & Boundary Conditions \\ \midrule
    Square plate with constant stress
    &
    \begin{minipage}{.3\textwidth}
      \includegraphics[width=\linewidth]{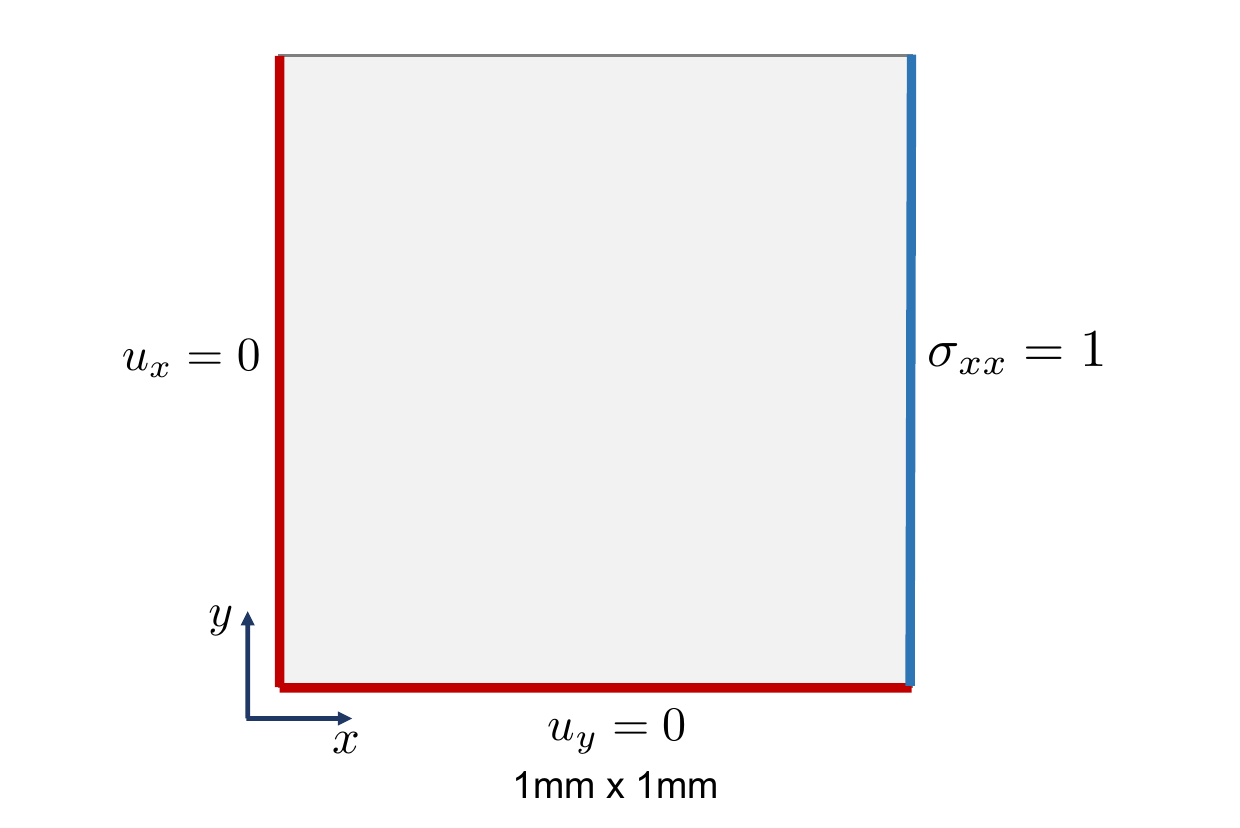}
    \end{minipage}
    &
    \begin{minipage}{.3\textwidth}
   \begin{equation}
    \begin{aligned}
    &u_x|_{x=0}=0, \\
    &u_y|_{y=0}=0, \\
    &\sigma_{xx}|_{x=1}=1,\sigma_{xy}|_{x=1}=0,\\
    &\sigma_{yy}|_{y=1}=0,\sigma_{xy}|_{y=1}=0,\\
\label{bc_example1}
    \end{aligned}
\end{equation} 
 \end{minipage}
    \\ \midrule
Square plate with a circular hole
    &
    \begin{minipage}{.3\textwidth}
      \includegraphics[width=\linewidth]{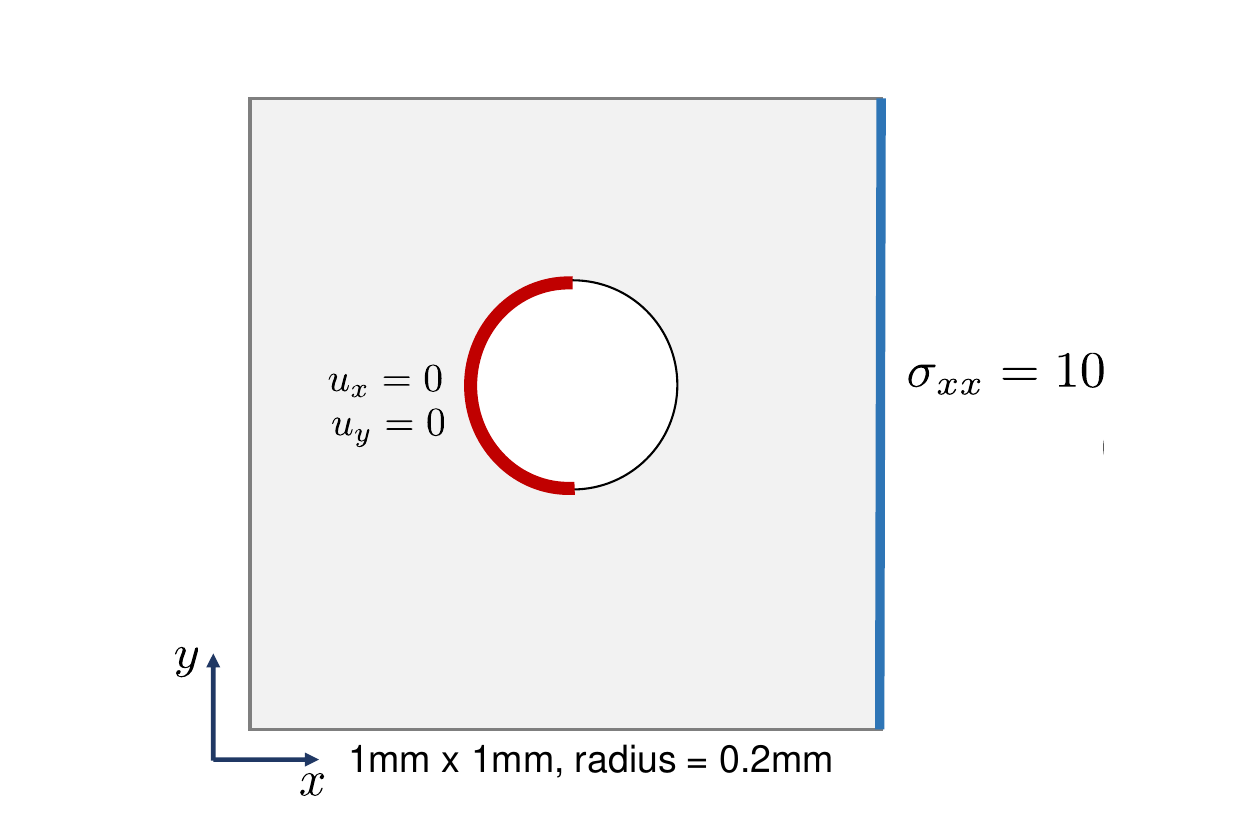}
    \end{minipage}
    &
    \begin{minipage}{.3\textwidth}
   \begin{equation}
    \begin{aligned}
    &u_x=0, u_y=0, \\
    &\forall x<0.5, \\
    &(x-0.5)^2+(y-0.5)^2=0.04, \\
    &\sigma_{xx}|_{x=1}=10,\sigma_{xy}|_{x=1}=0,\\
    &\sigma_{yy}|_{y=1}=0,\sigma_{xy}|_{y=1}=0,\\
\label{bc_example2}
    \end{aligned}
\end{equation} 
 \end{minipage}
    \\ \midrule
Square plate with discontinuous boundaries
    &
    \begin{minipage}{.3\textwidth}
      \includegraphics[width=\linewidth]{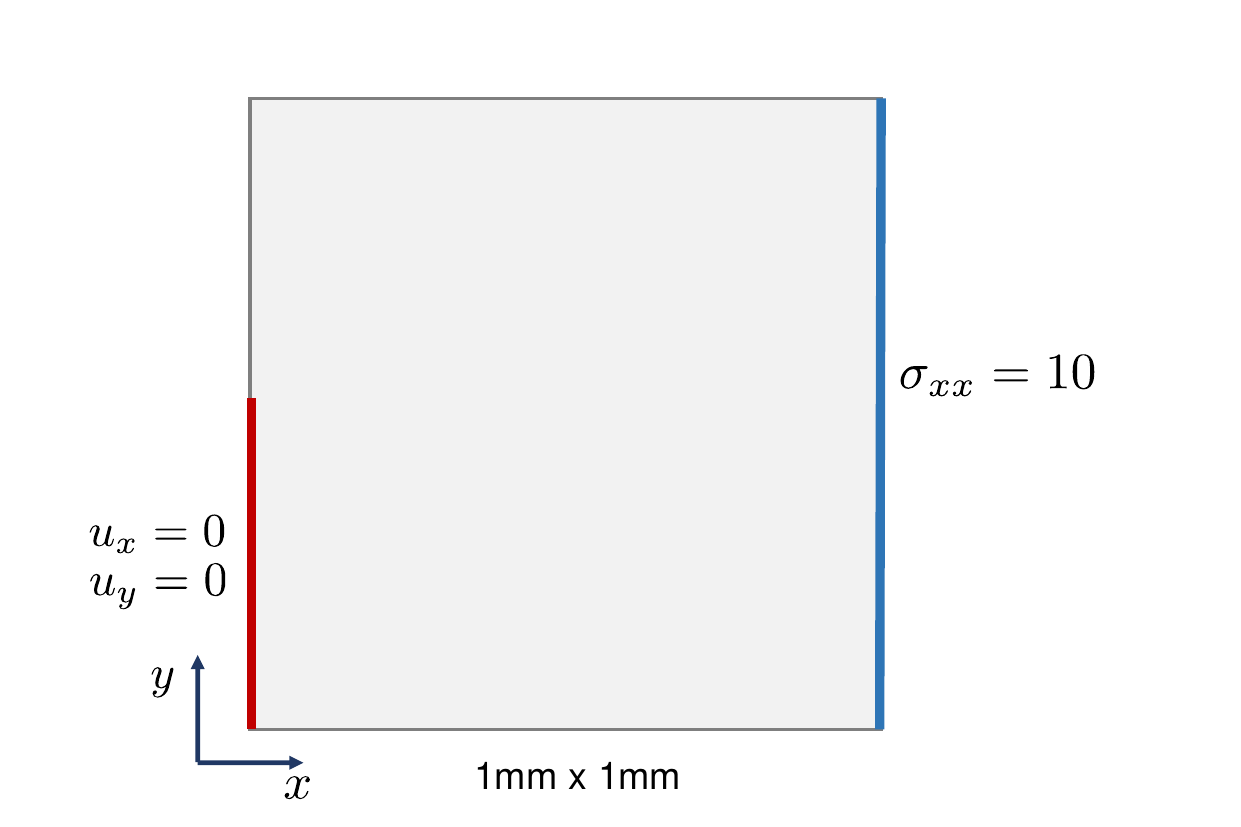}
    \end{minipage}
    &
    \begin{minipage}{.3\textwidth}
   \begin{equation}
    \begin{aligned}
    &u_x|_{x=0, y<0.5}=0, \\
    &u_y|_{x=0, y<0.5}=0, \\
    &\sigma_{xx}|_{x=1}=10,\sigma_{xy}|_{x=1}=0,\\
    &\sigma_{yy}|_{y=1}=0,\sigma_{xy}|_{y=1}=0,\\
\label{bc_example3}
    \end{aligned}
\end{equation} 
 \end{minipage}
    \\ \midrule
Square plate with point boundaries
    &
    \begin{minipage}{.3\textwidth}
      \includegraphics[width=\linewidth]{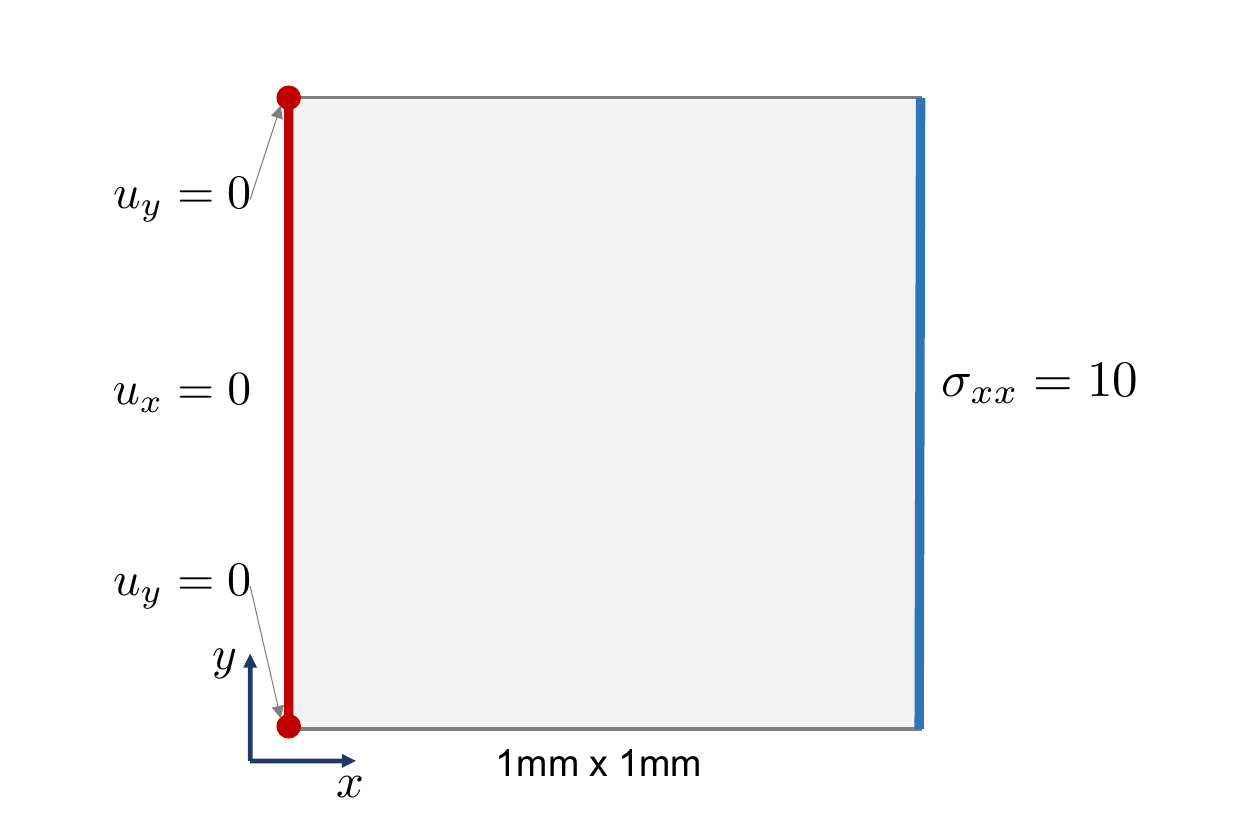}
    \end{minipage}
    &
    \begin{minipage}{.3\textwidth}
   \begin{equation}
    \begin{aligned}
    &u_x|_{x=0}=0, \\
    &u_y|_{x=0, y=0}=0, \\
    &u_y|_{x=0, y=1}=0, \\
    &\sigma_{xx}|_{x=1}=10,\sigma_{xy}|_{x=1}=0,\\
    &\sigma_{yy}|_{y=1}=0,\sigma_{xy}|_{y=1}=0,\\
\label{bc_example4}
    \end{aligned}
\end{equation} 
 \end{minipage}
    \\ \midrule
Square plate with a crack
    &
    \begin{minipage}{.3\textwidth}
      \includegraphics[width=\linewidth]{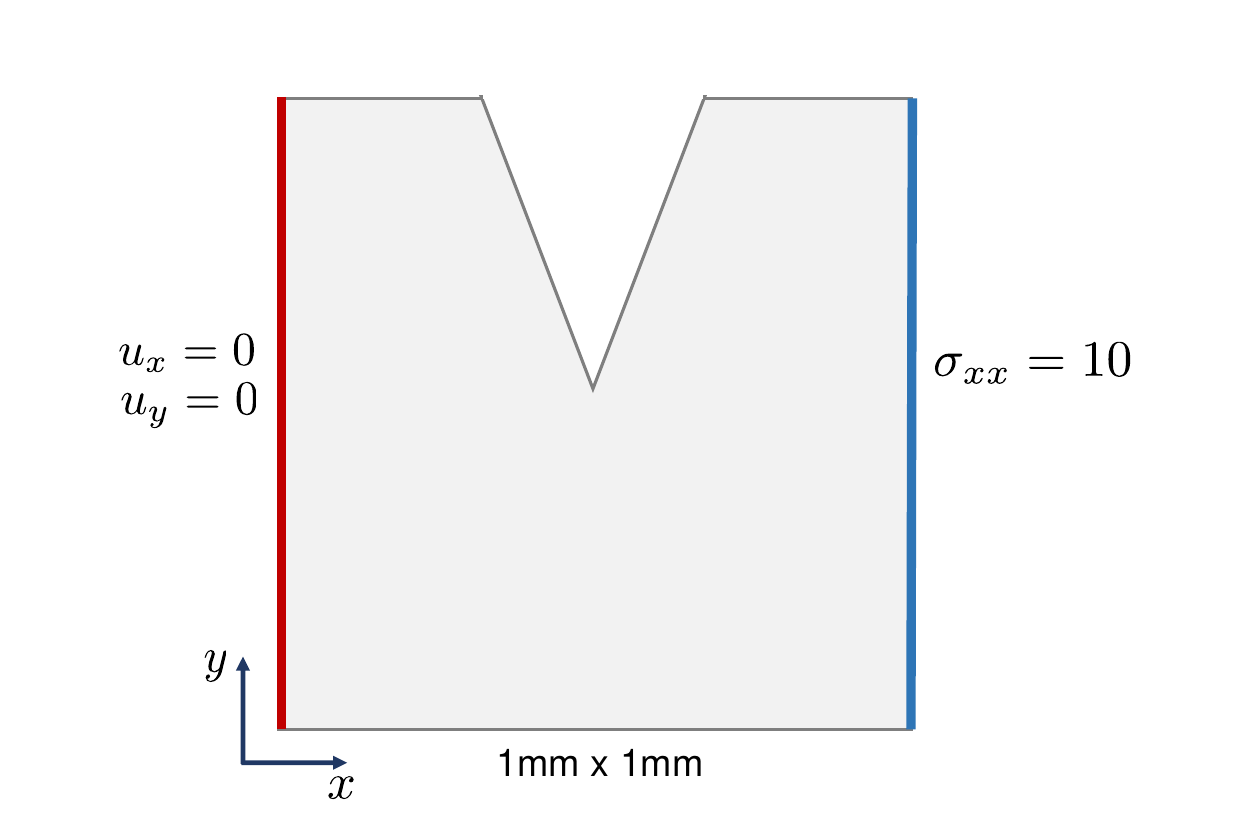}
    \end{minipage}
    &
    \begin{minipage}{.3\textwidth}
   \begin{equation}
    \begin{aligned}
    &u_x|_{x=0}=0, \\
    &u_y|_{x=0}=0, \\
    &\sigma_{xx}|_{x=1}=10,\sigma_{xy}|_{x=1}=0,\\
    &\sigma_{yy}|_{y=1}=0,\sigma_{xy}|_{y=1}=0,\\
\label{bc_example5}
    \end{aligned}
\end{equation} 
 \end{minipage}
    \\ \midrule
Cantilever beam with parabolic traction
    &
    \begin{minipage}{.3\textwidth}
      \includegraphics[width=\linewidth]{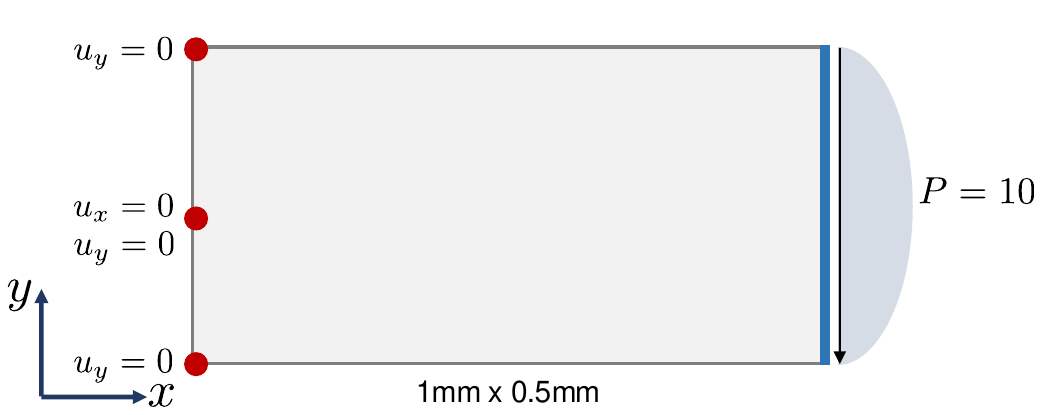}
    \end{minipage}
    &
    \begin{minipage}{.3\textwidth}
   \begin{equation}
    \begin{aligned}
    &u_x|_{x=0,y=0}=0, u_y|_{x=0,y=0}=0\\
    &u_y|_{x=0,y=-0.25}=0, \\
    &u_y|_{x=0,y=0.25}=0, \\
    &\sigma_{xx}|_{x=1}=0,\\
    &\sigma_{xy}|_{x=1}=10(0.25^2-y^2),\\
\label{bc_example6}
    \end{aligned}
\end{equation} 
 \end{minipage} \\
\bottomrule
\end{tabular}
\label{table:exps_doms_bcs}
\end{table}

For the numerical experiments, meshes are tailored to the geometry of each problem:

\begin{itemize}
    \item Square plate with constant stress: A triangular mesh extended over the square domain, with collocation points sampled at the centroids of the triangles.
    \item Square plate with a circular hole: The mesh is refined to conform to the circular boundary, maintaining centroids of triangles as collocation points.
    \item Square plate with discontinuous boundaries: A triangular mesh extended over the square domain, with collocation points sampled at the centroids of the triangles. 
    \item Square plate with point boundaries: A fine mesh generated near the point boundary for accurate modeling of boundary conditions. 
    \item Square plate with a crack: The mesh is refined to conform to the crack, maintaining centroids of triangles as collocation points.
    \item Cantilever beam with parabolic traction: A fine mesh near the fixed boundary ensures accurate modeling of displacement constraints and traction forces, with collocation points derived from triangle centroids throughout the domain.
\end{itemize}

\subsection{Baseline models}

For the experiments, the performance of the proposed PINN-FEM model is compared against three benchmark models. The first benchmark model is a vanilla PINN network with soft imposition of boundary conditions. We use total energy loss for training this model, as defined in Eq.~\eqref{energy_loss_2d_bvp}. Boundary condition loss term is added to this loss and weighted by a factor $\beta$. Therefore, the loss function for this model can be written as,

\begin{equation}
    \begin{aligned}
    J_\text{soft}(\bm \theta) = \mathcal{E}(\bm u(\bm \theta)) + \beta \sum_{\bm x_n\in \Gamma_{g}}({\bf u}(\bm x_n)-\hat{{\bf u}}(\bm x_n))^2.
\label{eq_result_nv}
    \end{aligned}
\end{equation}
It is to be noted that, the Neumann boundary conditions are automatically satisfied by loss calculated using energy formulation. Therefore, only the soft imposition of Dirichlet boundary conditions are required. The second benchmark model is a PINN network with exact imposition of boundary conditions using the theory of R-functions, introduced in \cite{sukumar2022exact}. This method calculates approximate distance function (ADF), $\phi$, to the domain boundary, $\Gamma$ for the imposition of Dirichlet boundary conditions. Let $\bf{u}_\phi$ be the PINN output with exact BC imposition using ADF. Then, the final predicted displacement, $\hat{\bf{u}}$, is given by,
\begin{equation}
    \begin{aligned}
    \hat{\bf{u}}(\bm x; \bm \theta) = \phi(\bm x) \cdot {\bf u}_{\phi}(\bm x; \bm \theta)+g,
\label{eq_result_nr}
    \end{aligned}
\end{equation}
where, $g$ is the composite Dirichlet boundary condition given by $g=w_1g_1+w_2g_2$. A detailed explanation on how the weights, $w_1$ and $w_2$, are calculated can be found in \cite{sukumar2022exact}. The third benchmark model is also a PINN network with exact imposition of boundary conditions. Here, instead of ADFs, a continuous global function is used as the Distance Function (DF) to impose the boundary conditions, similar to Eq.~\eqref{eq_result_nr}. In our experiments, we use $x$ and $y$ coordinates as the global functions for this model. The baseline models are denoted as follows for the following experiments. 
\begin{itemize}
    \item Soft: Vanilla PINN with soft boundary condition enforcement;
    \item ADF: PINN with exact boundary condition imposition using approximate distance function;
    \item DF: PINN with exact boundary condition imposition using global function as  distance function;
\end{itemize}

\subsection{Experiments}
\label{exps}

\subsubsection{Square plate with constant stress}

For the first experiment, we consider a two-dimensional square elastic plate under plane stress conditions. This example is a benchmark problem presented in \cite{li2021physics}. A uniform stretching force of 1 unit is applied in x-direction, with roller boundary condition on left and bottom edges. The domain and the detailed boundary condition equations are given in Table \ref{table:exps_doms_bcs}. For the proposed PINN-FEM network, we use a fully connected neural network with 5 hidden layers and 32 neurons per hidden layer. The input to the network comprises the location coordinates, $(x,y)$ and the output is the predicted displacement field, $(\hat{u}_x, \hat{u}_y)$. The hyperbolic (Tanh) function is used as the activation function, with a learning rate of $1E-4$. LBFGS is used as the optimizer for training the model. The collocation points for the network are the nodal points from a mesh of the domain, generated using Gmsh, with a mesh size of 0.1 mm, as detailed in Section \ref{mesh_gen}. The ground truth for this problem is the finite element solution calculated on the same mesh using Abaqus. For the three baseline models, we use a PINN network with the same architecture as that of PINN-FEM network. However, Adam optimizer is found be perform best for these models and thus is used for the training for all the baseline models. 

\begin{table}[!ht]
  \centering
  \caption{Relative error in predicting displacement fields, $u_x$ and $u_y$ for different PINN models. Soft refers to the vanilla PINN with soft BC imposition. ADF and DF impose exact BC in PINN using approximate distance function and distance function respectively. PINN-FEM is the proposed approach. Relative error, $e$ is calculated using Eq.~\eqref{eq:rel_error}.}
\begin{tabular}{l|c|cr}
\toprule
Experiment & Model  & $e(u_x) \downarrow$ & $e(u_y) \downarrow$  \\
\midrule
Square plate with constant stress & Soft & 0.10 & 0.10 \\
 & ADF & 0.01 & 0.02 \\
 & DF & 0.01 & 0.03 \\
 & PINN-FEM  & 0.01 & 0.03 \\
\midrule
Square plate with a circular hole & Soft & 0.087 & 0.13 \\
 & ADF & 0.062 & 0.12 \\
 & PINN-FEM  & 0.02 & 0.06 \\
\midrule
Square plate with discontinuous boundaries & Soft & 0.17 & 0.09 \\
 & ADF & 0.03 & 0.01 \\
 & PINN-FEM  & 0.02 & 0.02 \\
\midrule
Square plate with point boundaries & Soft & 0.07 & 0.11 \\
 & PINN-FEM  & 0.005 & 0.09 \\
\midrule
Square plate with a crack & Soft & 0.22 & 0.30 \\
 & ADF & 0.07 & 0.13 \\
 & DF & 0.08 & 0.15 \\
 & PINN-FEM  & 0.05 & 0.06 \\
\midrule
Cantilever beam with parabolic traction& Soft & 0.56 & 0.56 \\
 & PINN-FEM  & 0.0009 & 0.0003 \\
\bottomrule
\end{tabular}
\label{table:rel_error}
\end{table}

The performance of the models are evaluated by calculating the relative error, $e$, in predicting the displacement fields. The relative error is given by \begin{equation}
e(\bm u(\bm x; \bm \theta)) = \dfrac{\|\widehat{\bm{u}}(\bm x; \bm \theta) - \bm u(\bm x)\|_{\ell^1}}{\|\bm u(\bm x)\|_{\ell^1}}, 
\label{eq:rel_error}
\end{equation}
where $\ell^1$ is the L1-norm of the variable of interest. Table \ref{table:rel_error} shows the relative errors in predicting $u_x$ and $u_y$ for all the experiments. The results of this experiment for all the models are presented in Table \ref{table:rel_error} as Exp 1. We observe that all the models with exact imposition of boundary conditions have similar performance, with same relative errors. Each of these models outperforms the vanilla PINN, which uses soft imposition of Dirichlet boundary conditions. This is also evident from Figure \ref{fig:exp1_pred}, which shows the distribution of the predicted displacement fields and the corresponding error distribution for the three benchmark models and the proposed approach. For comparison, it also incliudes the ground truth obtained using the finite element simulations.
     
\begin{table}[!ht]
  \centering
    \caption{The distribution of predicted displacement fields, $\hat{u}_x$ and $\hat{u}_y$, and the corresponding error for the square elastic plate with uniform stretching force, under plane stress conditions, with boundary conditions as given in Eq.~\eqref{bc_example1}.}
  \begin{tabular}{ |c | c | c |c | c| }
    \hline
    PINN Model & $\hat{u}_x$ & $\hat{u}_y$ & $|\hat{u}_x-u_x|$ & $|\hat{u}_y-u_y|$ \\ \hline
    Ground Truth
    &
    \begin{minipage}{.19\textwidth}
      \includegraphics[width=0.9\linewidth]{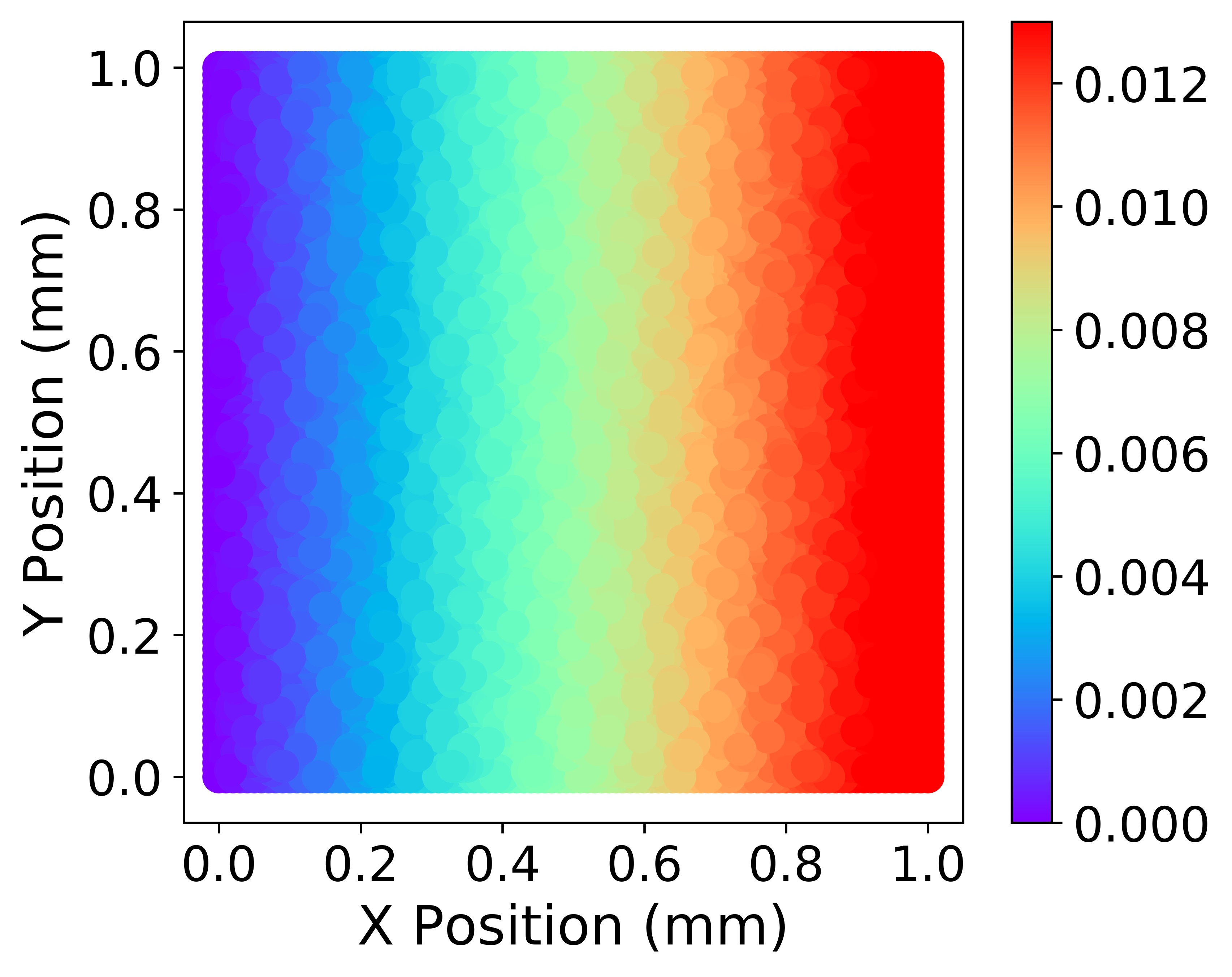}
    \end{minipage}
    &
   \begin{minipage}{.19\textwidth}
      \includegraphics[width=0.9\linewidth]{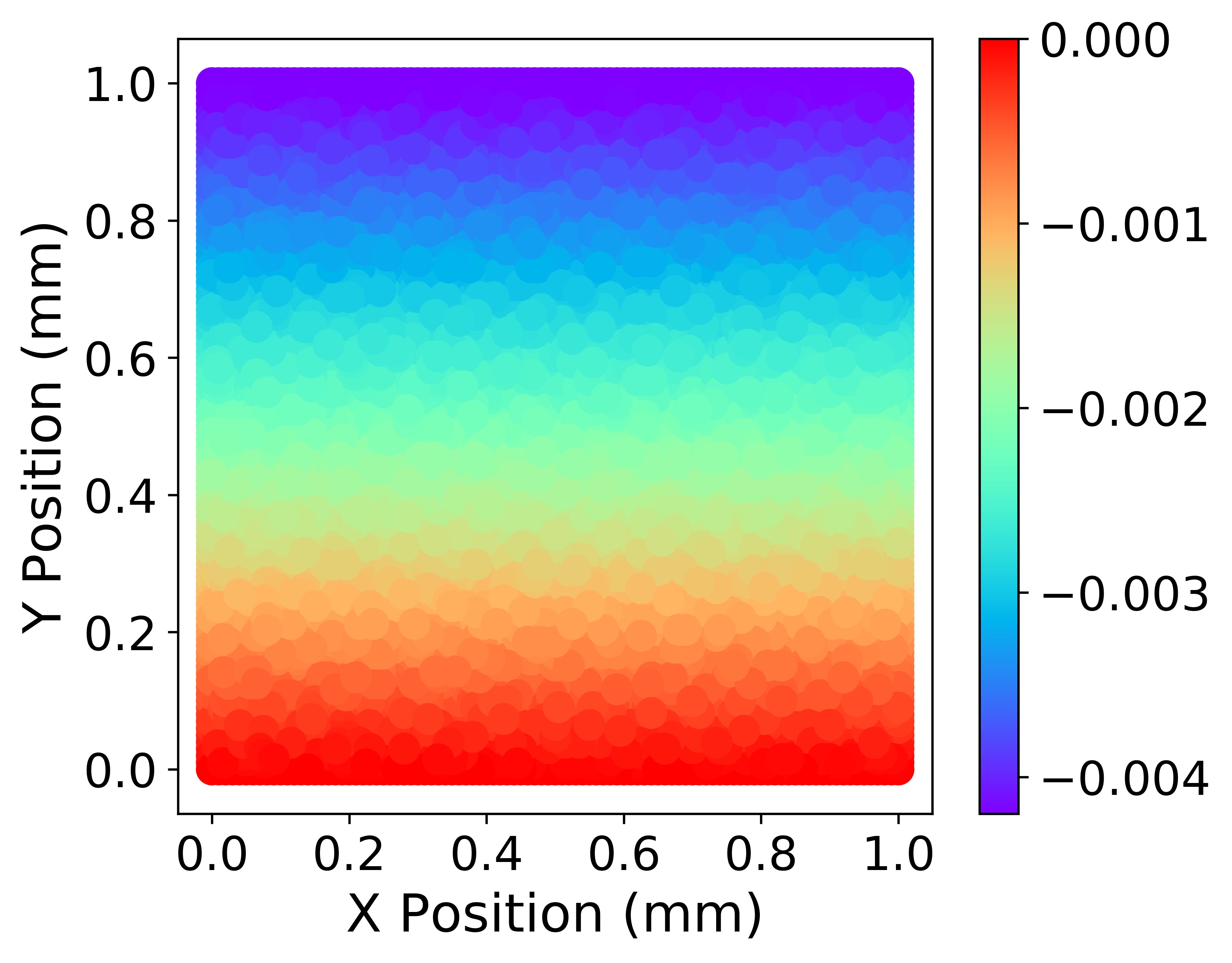}
    \end{minipage}
    & 
     & 
    \\ \hline
    Soft 
    &
    \begin{minipage}{.19\textwidth}
      \includegraphics[width=0.9\linewidth]{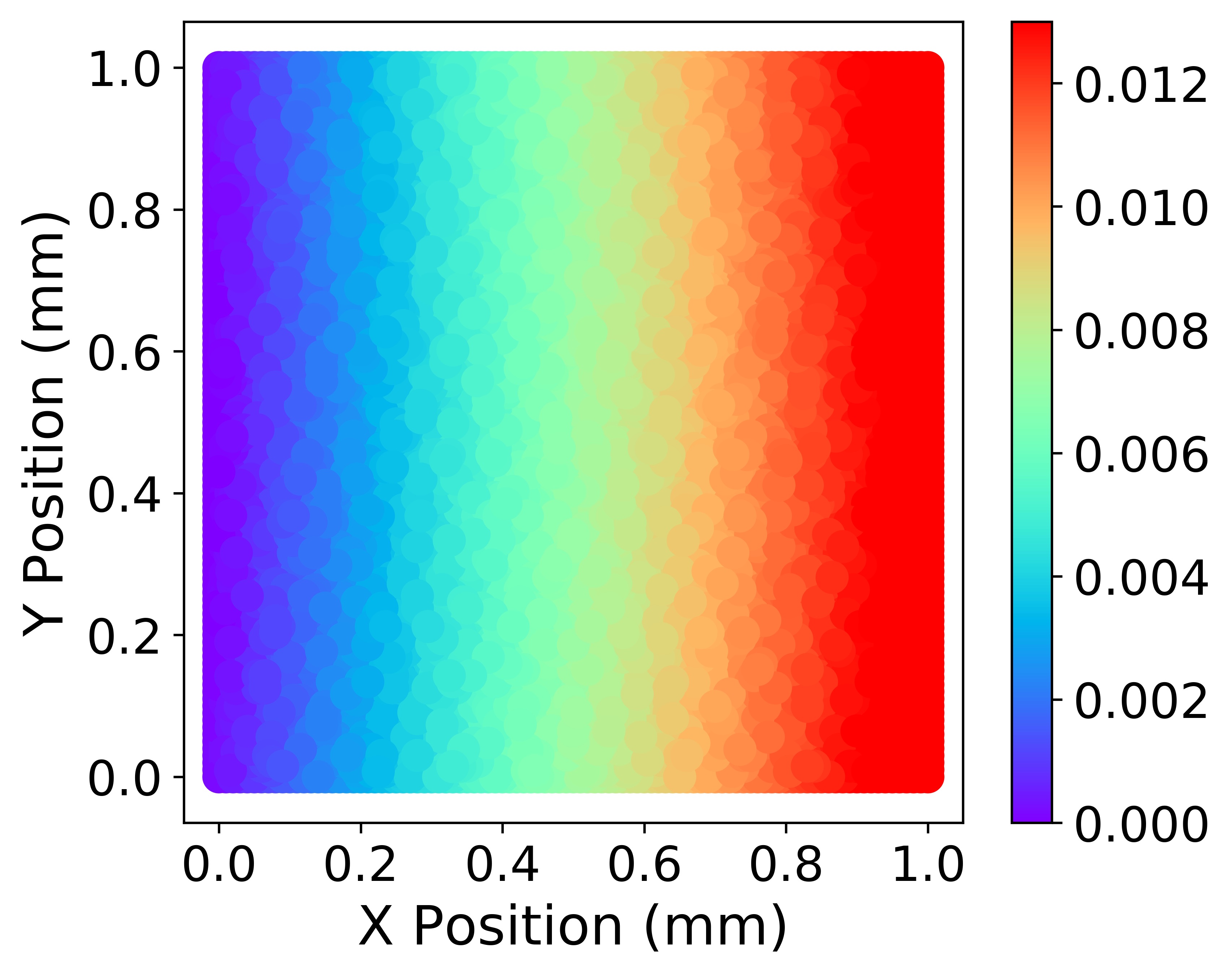}
    \end{minipage}
    &
   \begin{minipage}{.19\textwidth}
      \includegraphics[width=0.9\linewidth]{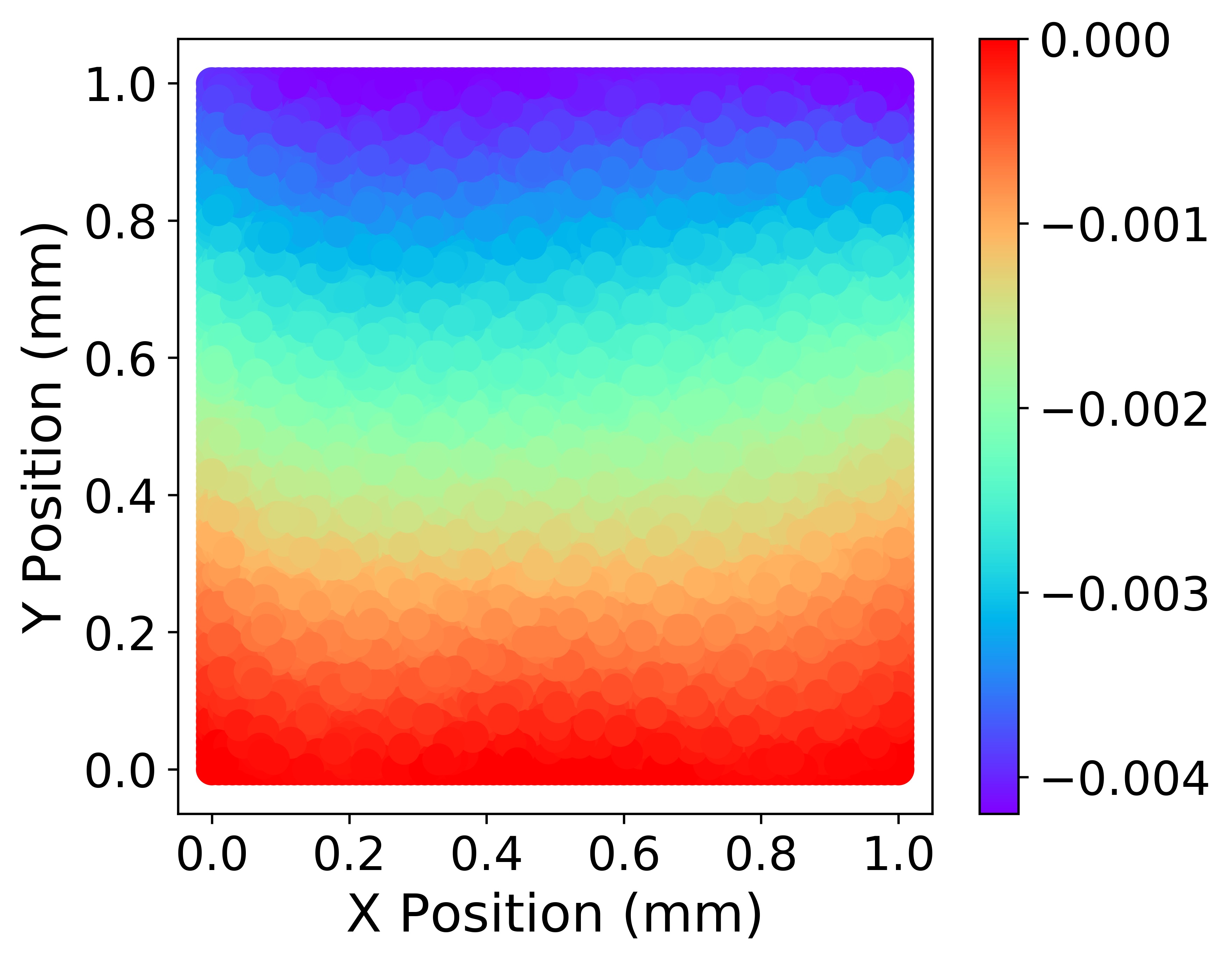}
    \end{minipage}
    & 
   \begin{minipage}{.19\textwidth}
      \includegraphics[width=0.9\linewidth]{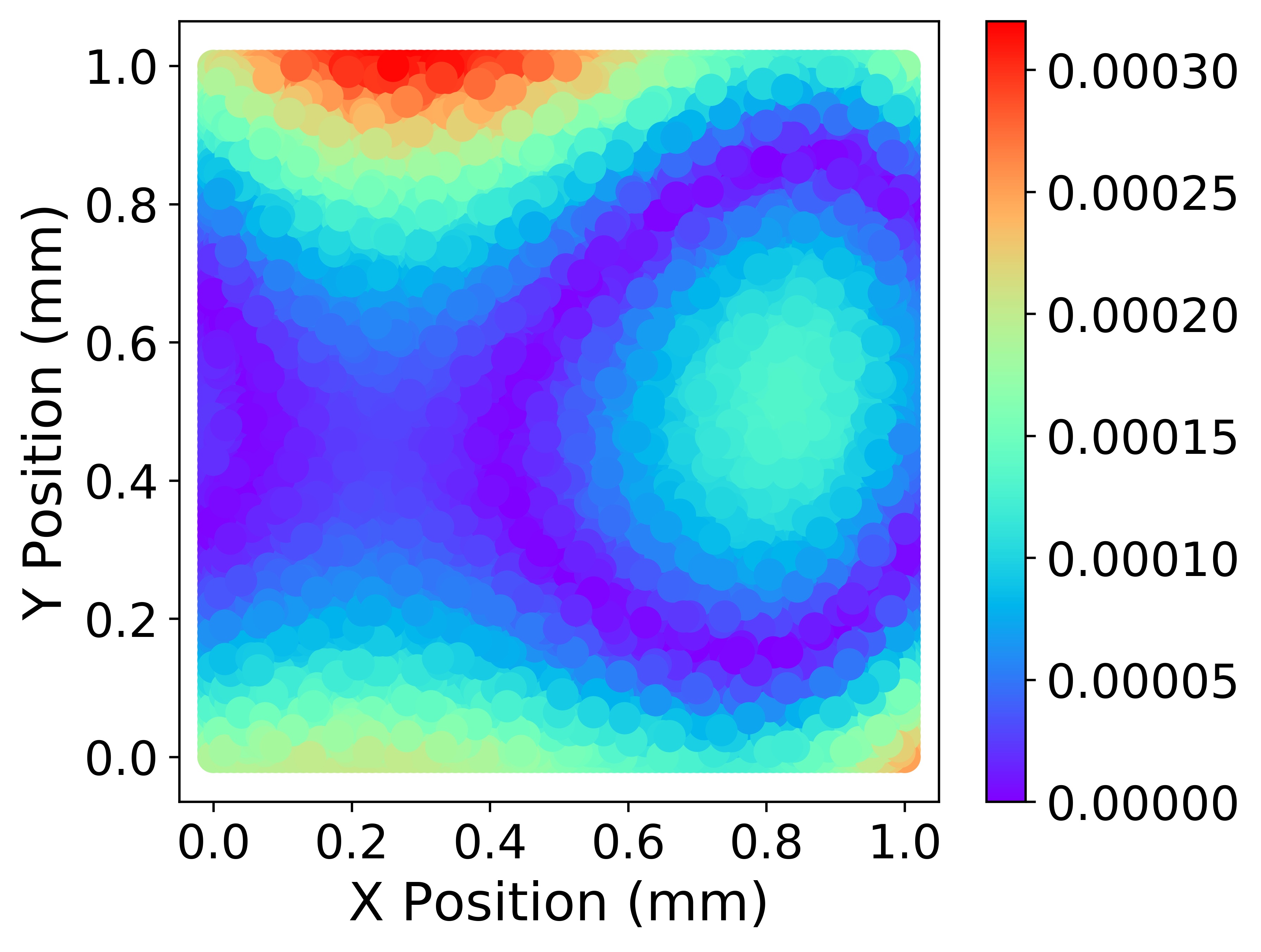}
    \end{minipage}
     & 
   \begin{minipage}{.19\textwidth}
      \includegraphics[width=0.9\linewidth]{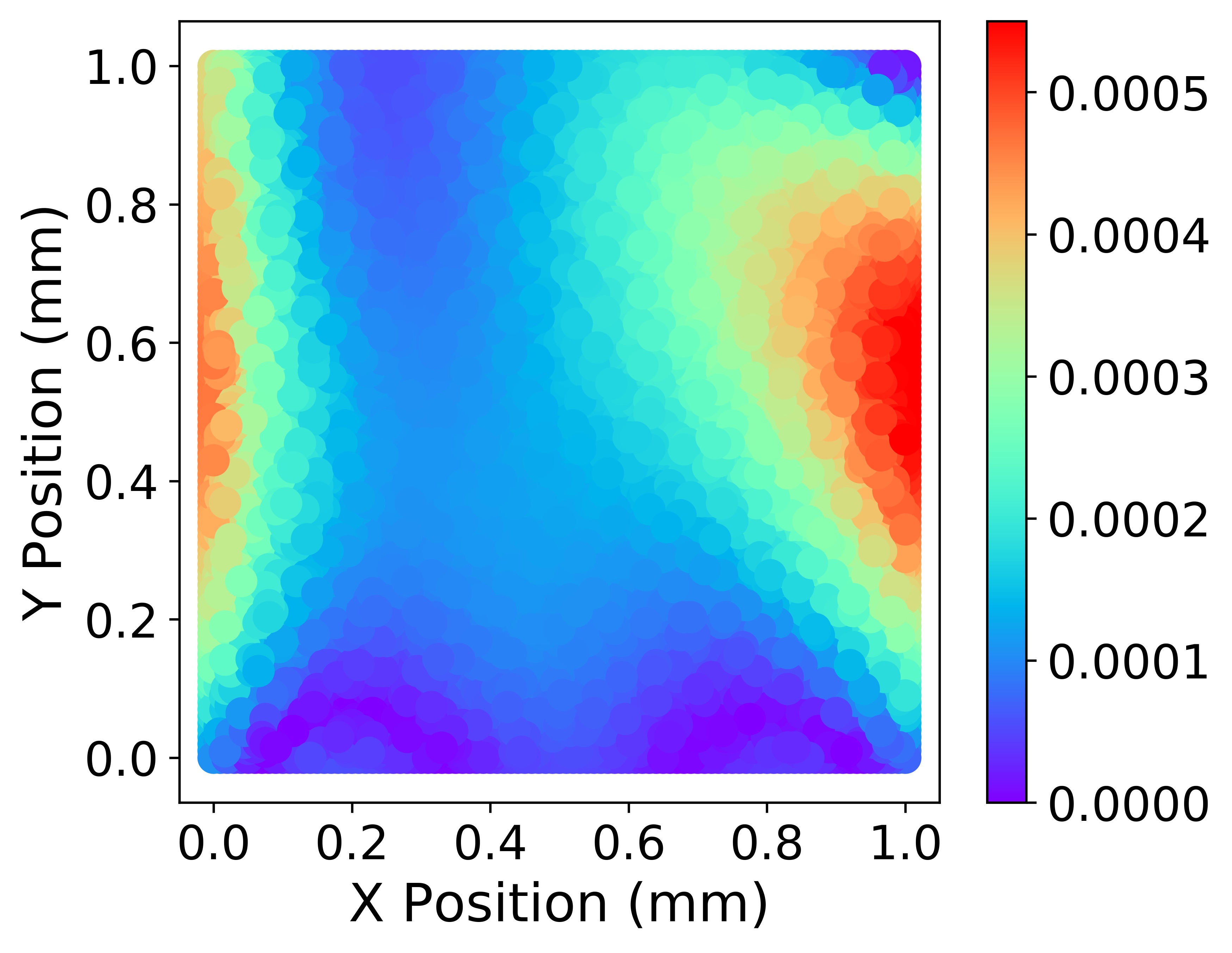}
    \end{minipage}
    \\ \hline
    ADF
    &
    \begin{minipage}{.19\textwidth}
      \includegraphics[width=0.9\linewidth]{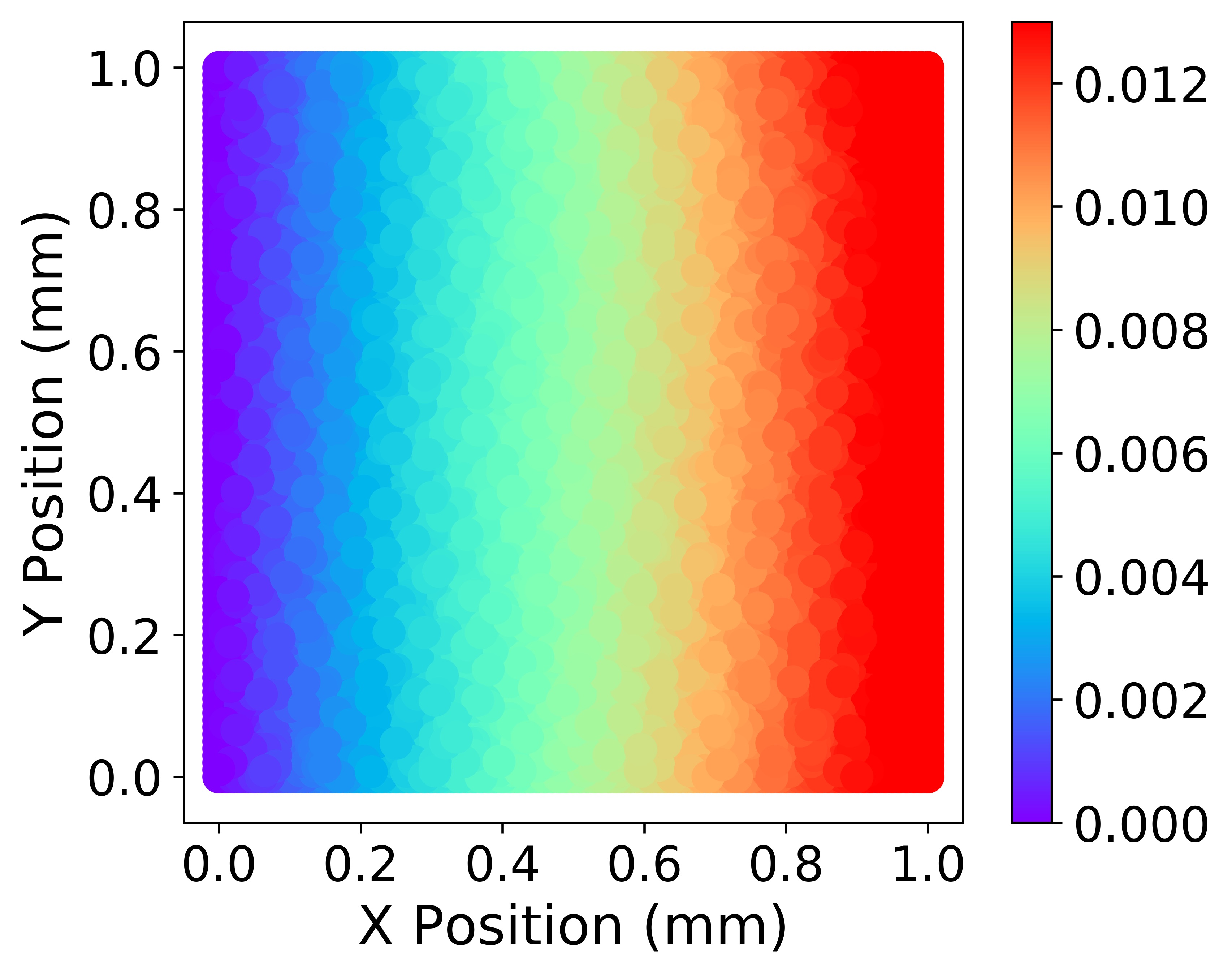}
    \end{minipage}
    &
   \begin{minipage}{.19\textwidth}
      \includegraphics[width=0.9\linewidth]{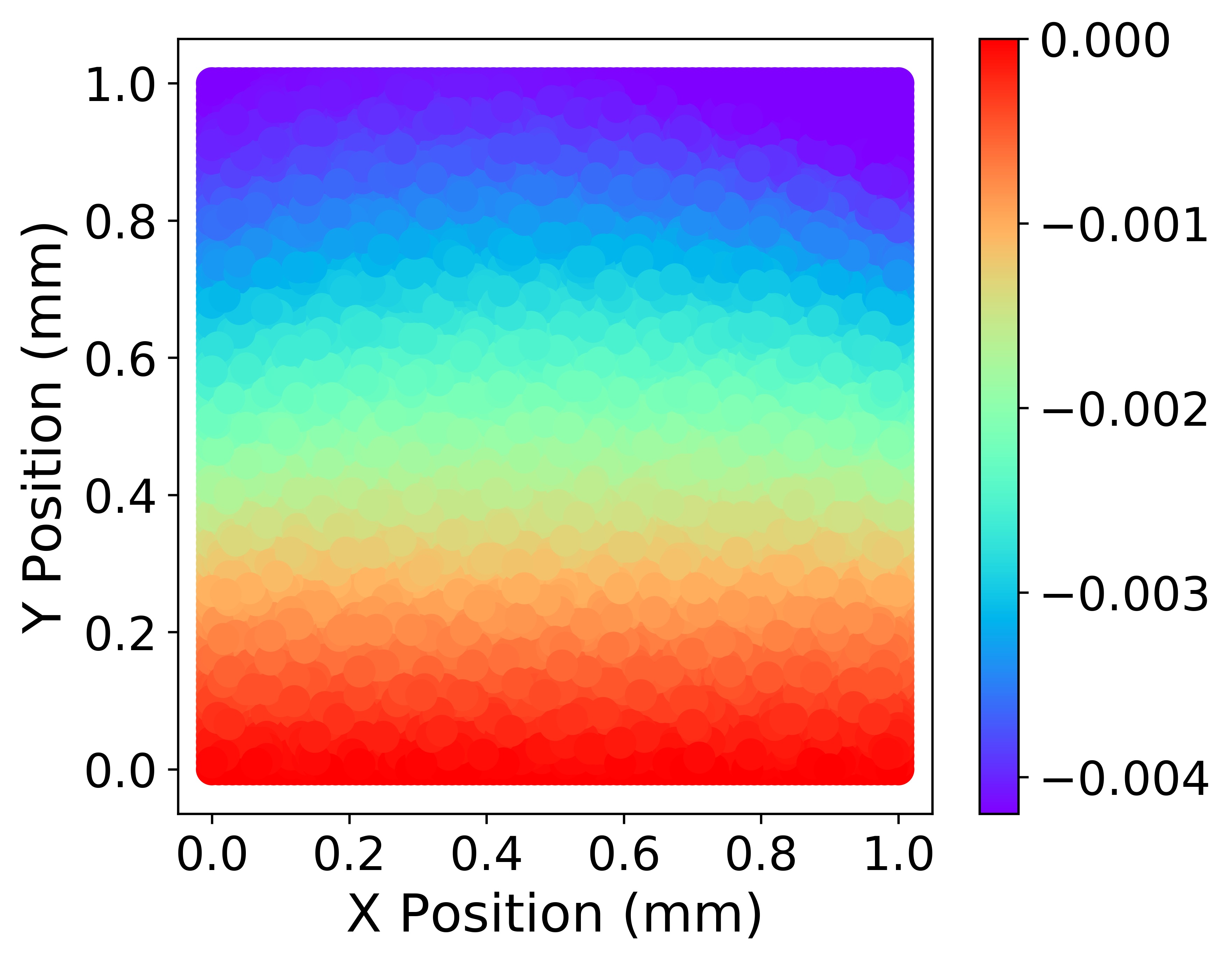}
    \end{minipage}
    & 
   \begin{minipage}{.19\textwidth}
      \includegraphics[width=0.9\linewidth]{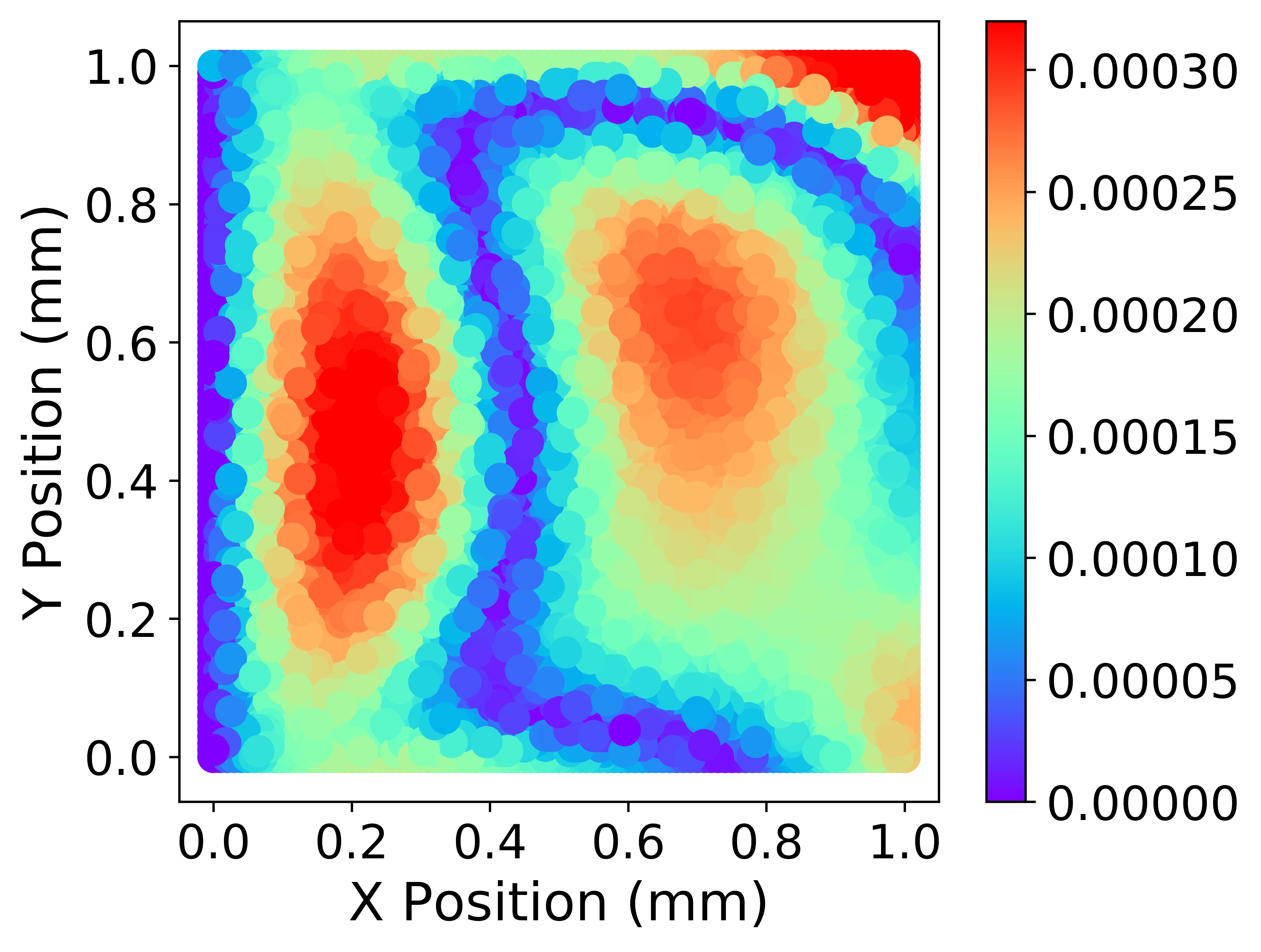}
    \end{minipage}
     & 
   \begin{minipage}{.19\textwidth}
      \includegraphics[width=0.9\linewidth]{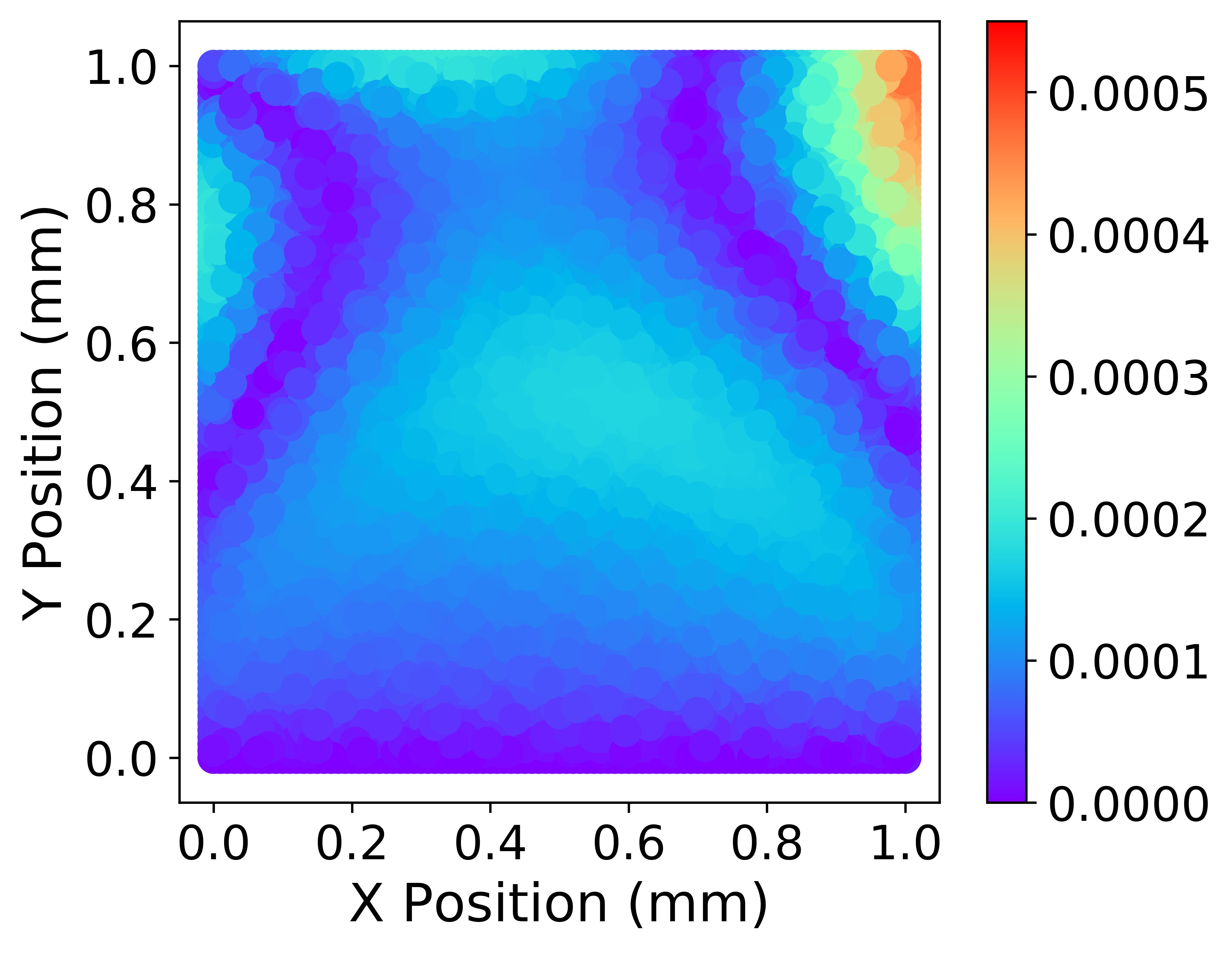}
    \end{minipage}
    \\ \hline
    DF
    &
    \begin{minipage}{.19\textwidth}
      \includegraphics[width=0.9\linewidth]{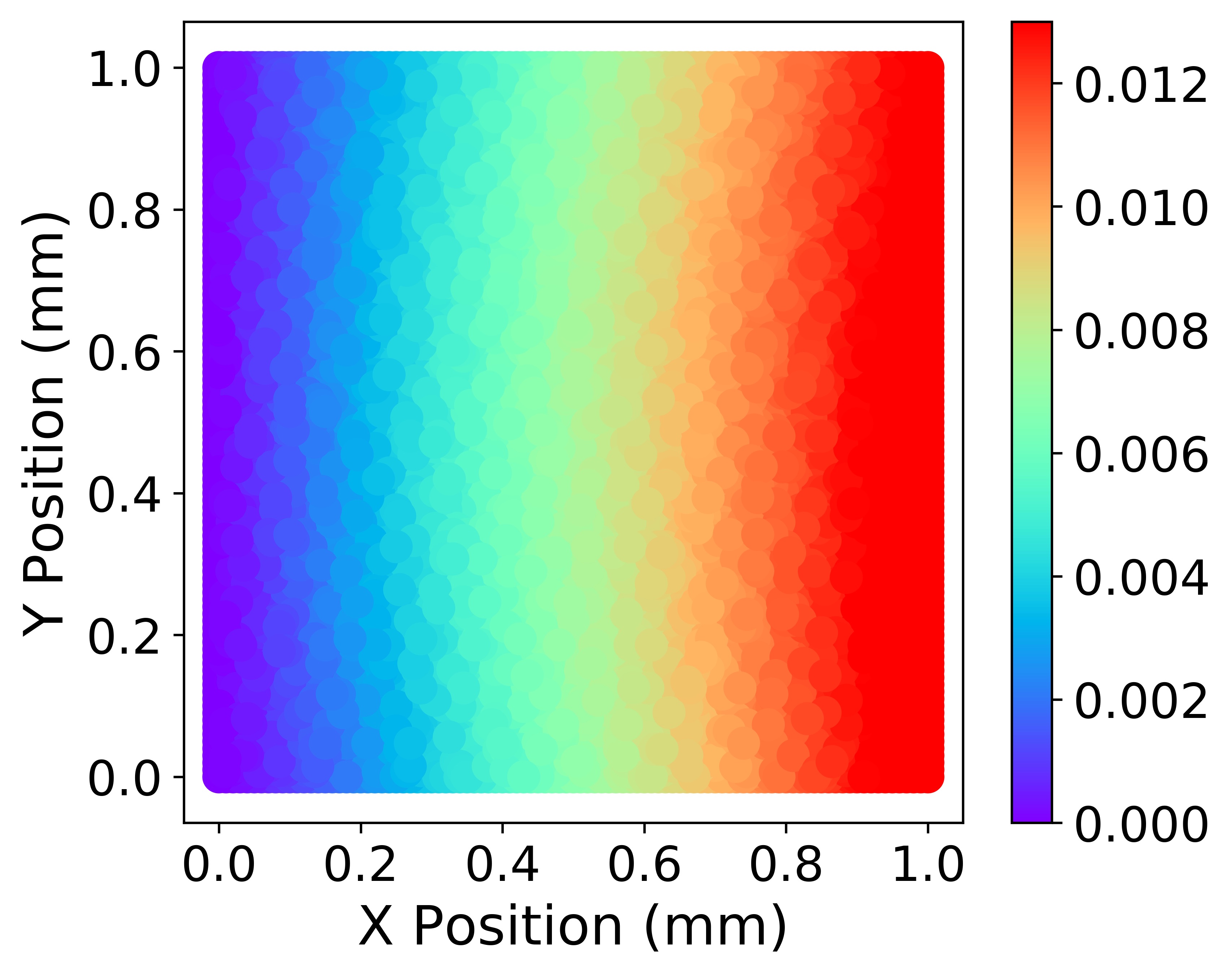}
    \end{minipage}
    &
   \begin{minipage}{.19\textwidth}
      \includegraphics[width=0.9\linewidth]{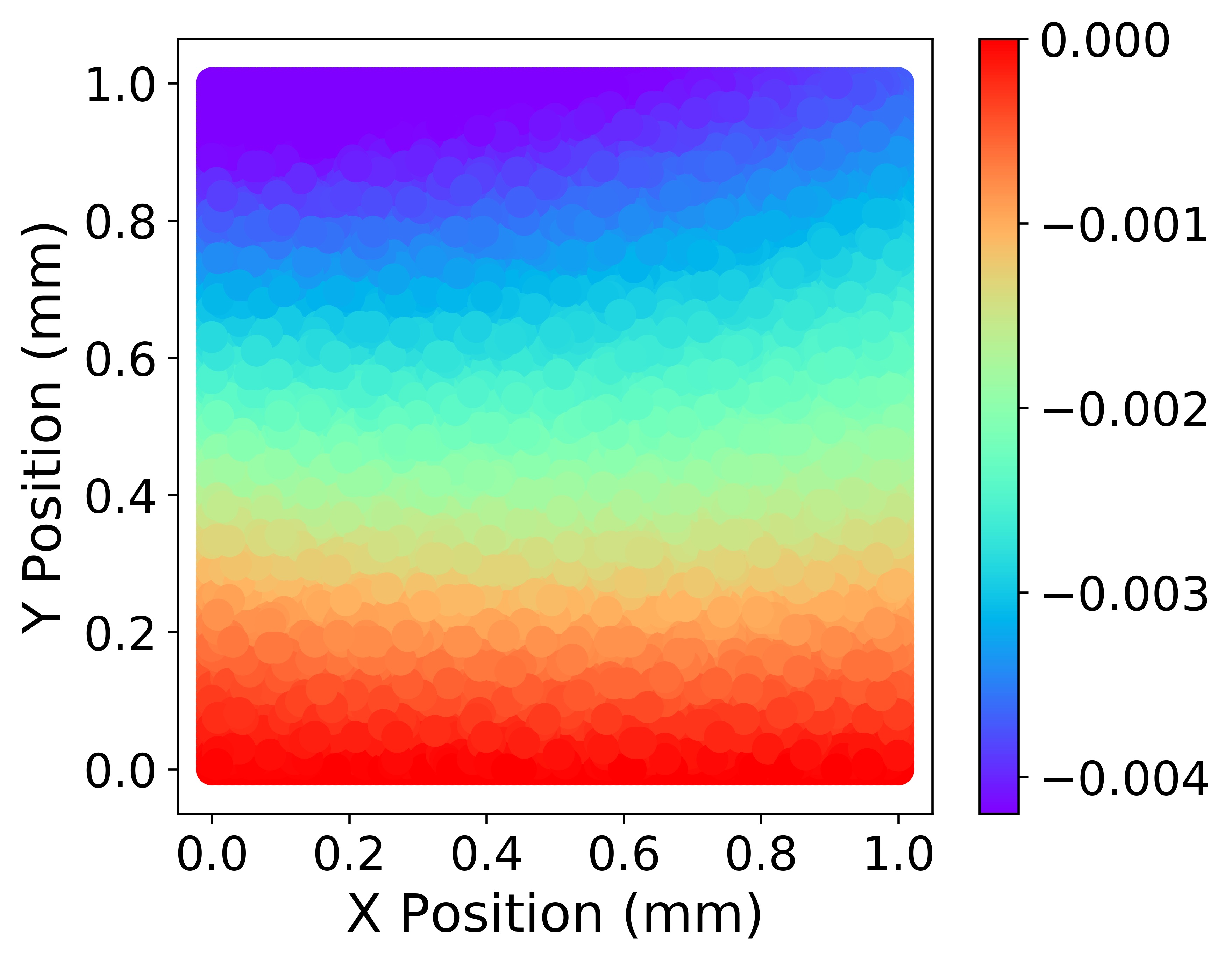}
    \end{minipage}
    & 
   \begin{minipage}{.19\textwidth}
      \includegraphics[width=0.9\linewidth]{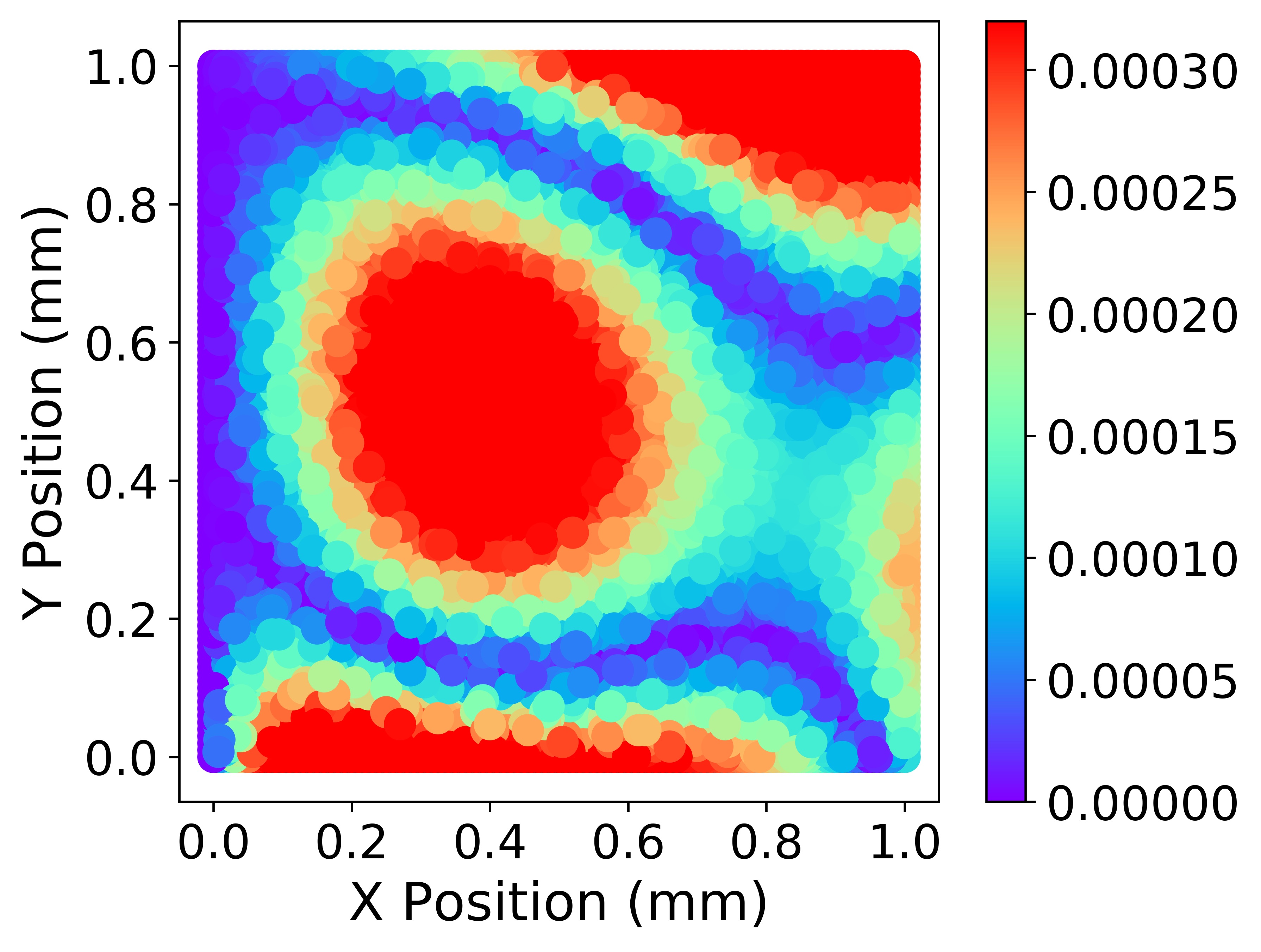}
    \end{minipage}
     & 
   \begin{minipage}{.19\textwidth}
      \includegraphics[width=0.9\linewidth]{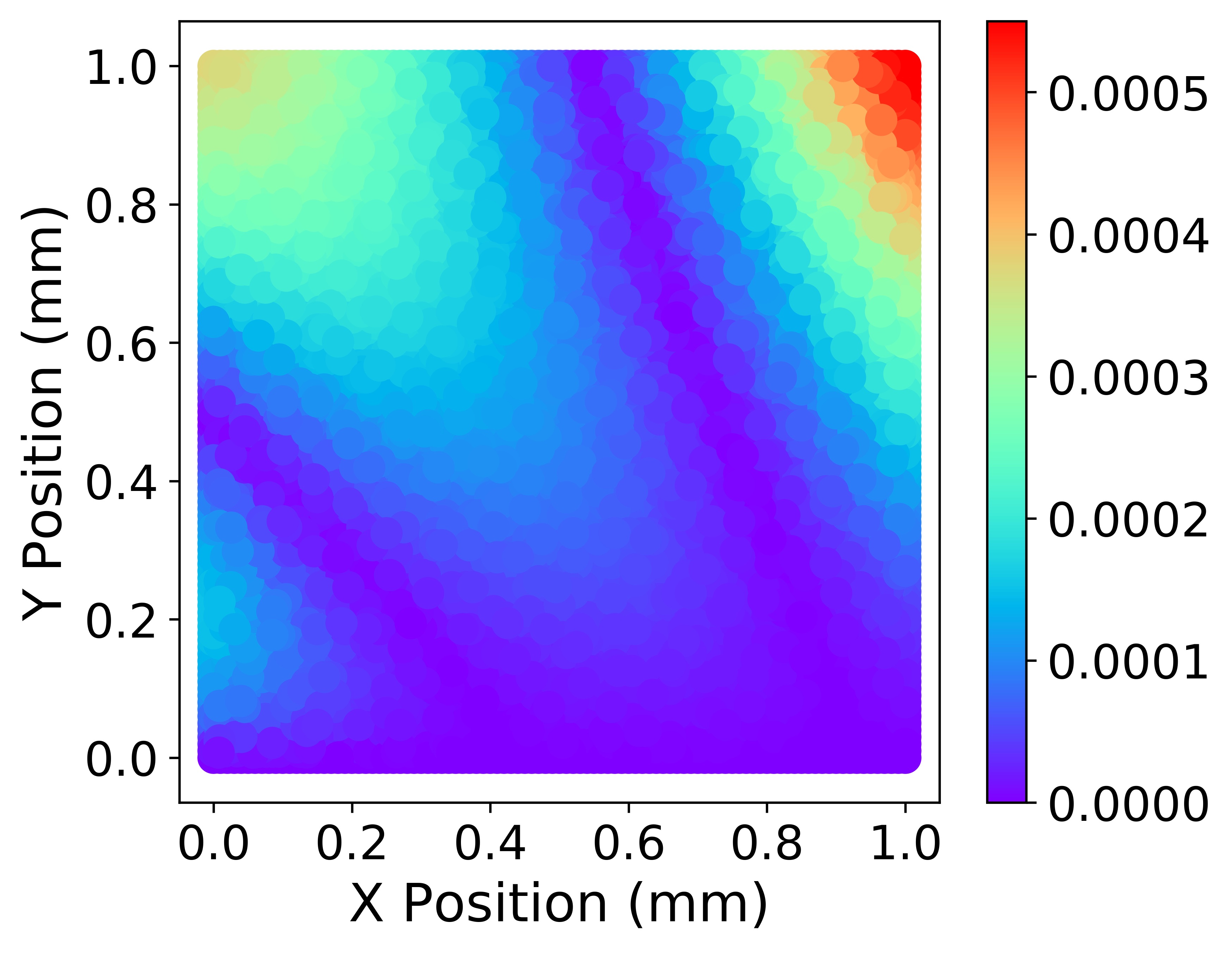}
    \end{minipage}
    \\ \hline
    PINN-FEM
    &
    \begin{minipage}{.19\textwidth}
      \includegraphics[width=0.9\linewidth]{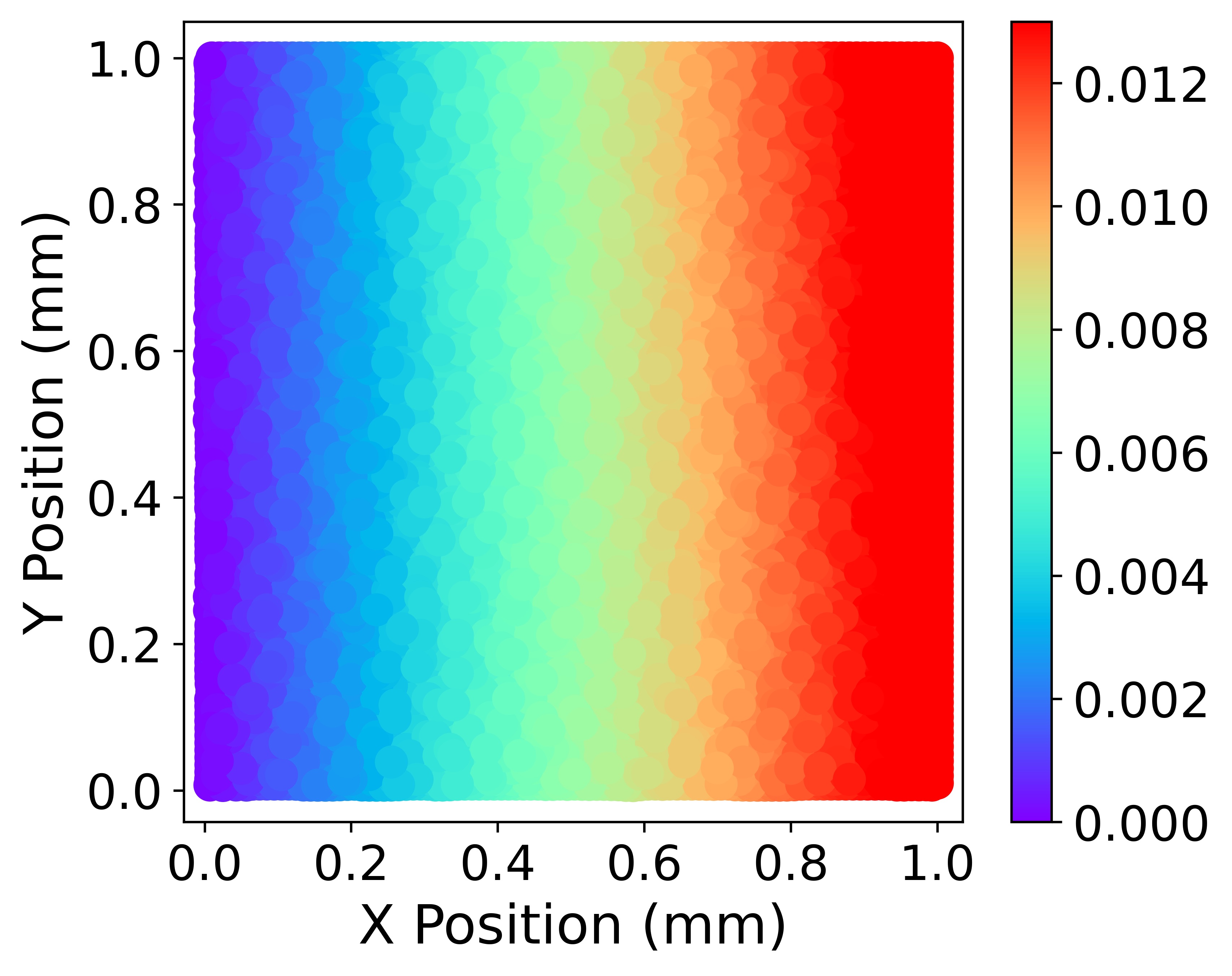}
    \end{minipage}
    &
   \begin{minipage}{.19\textwidth}
      \includegraphics[width=0.9\linewidth]{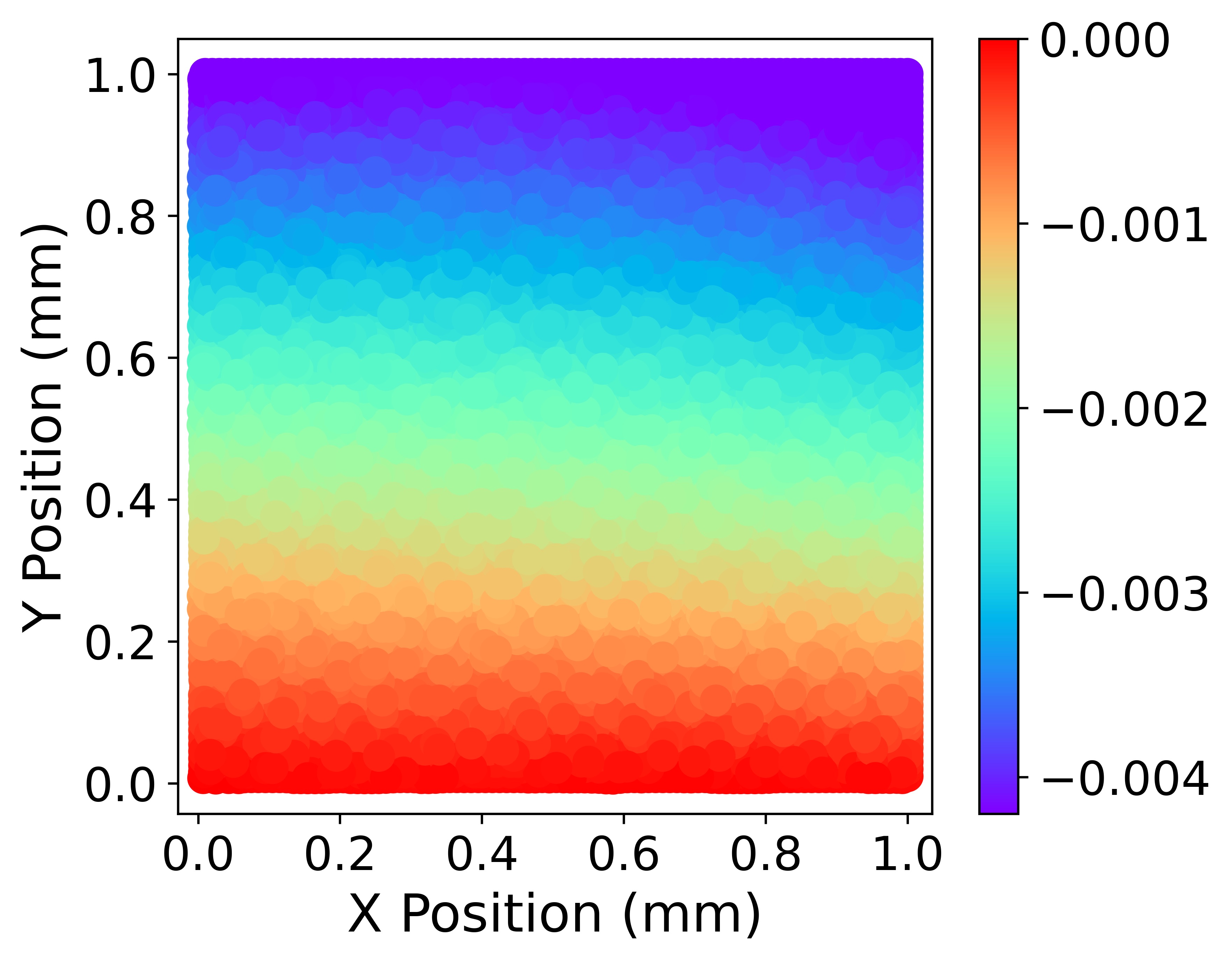}
    \end{minipage}
    & 
   \begin{minipage}{.19\textwidth}
      \includegraphics[width=0.9\linewidth]{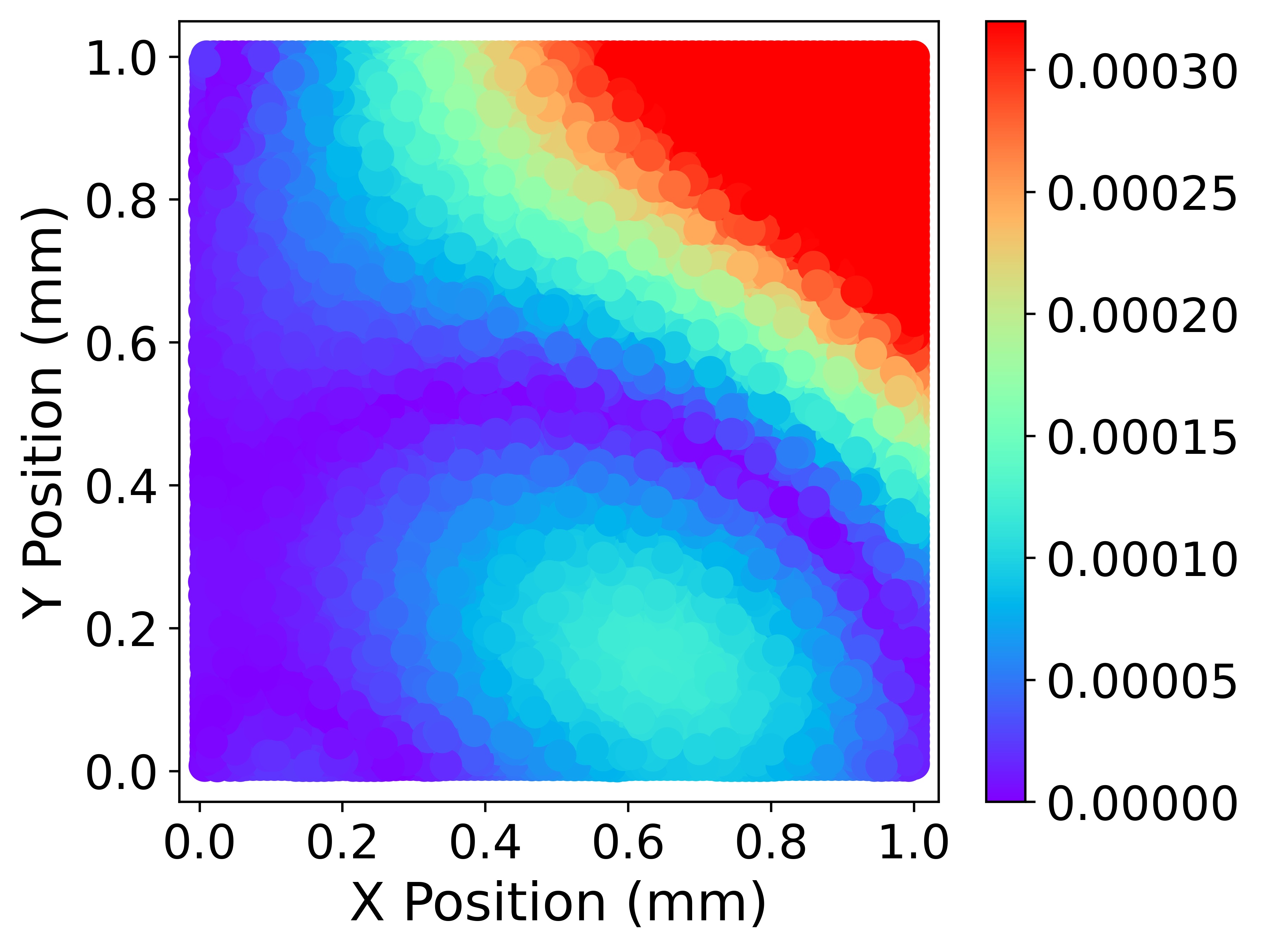}
    \end{minipage}
     & 
   \begin{minipage}{.19\textwidth}
      \includegraphics[width=0.9\linewidth]{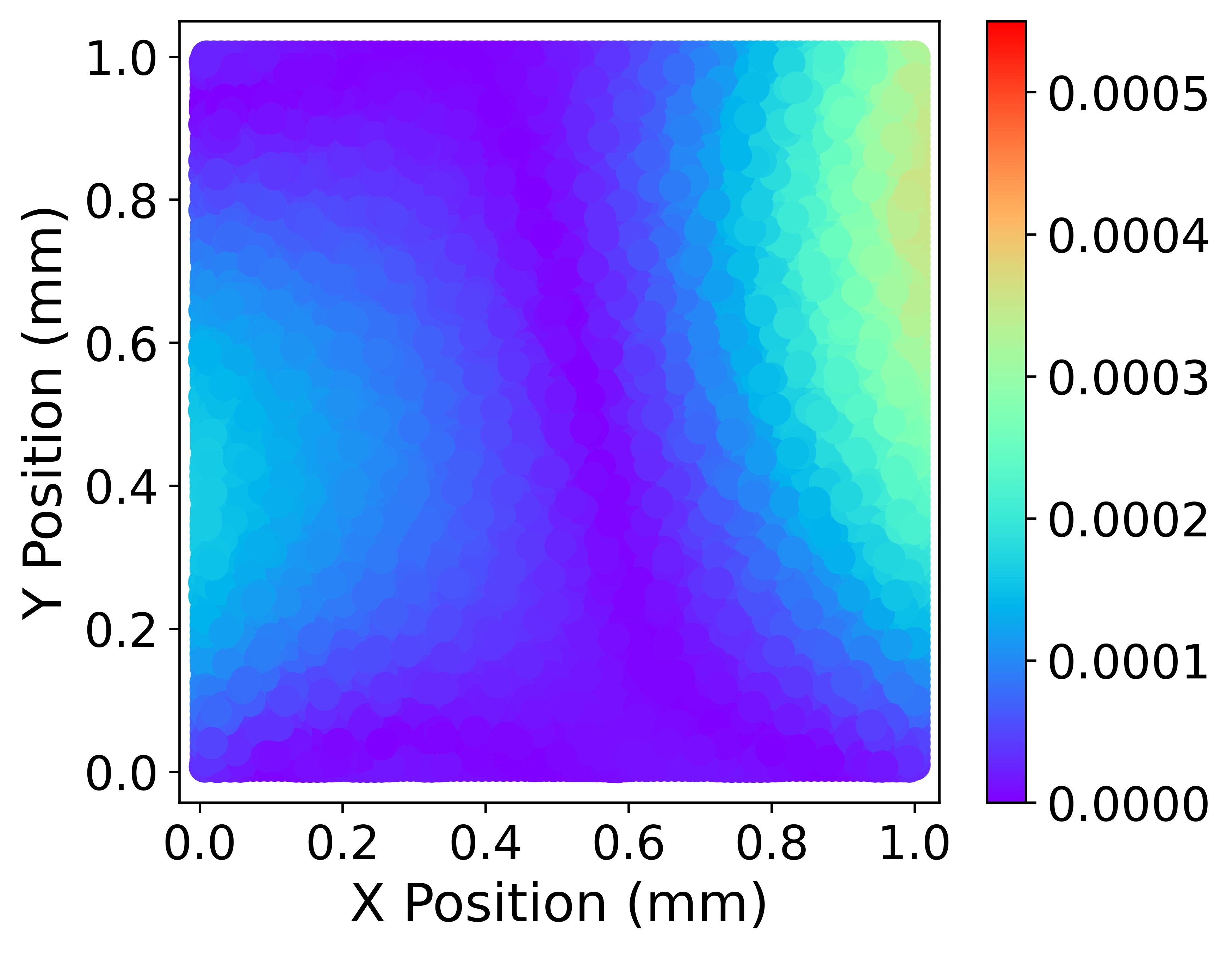}
    \end{minipage}
    \\ \hline
  \end{tabular}
  \label{fig:exp1_pred}
\end{table}

\subsubsection{Square plate with a circular hole}

In this example, we consider a square plate with a circular hole of radius $0.2$, centered at $(0.5,0.5)$. A uniform stretching force of 10 units is applied in the x-direction. The left half of the circular hole is fixed. The detailed boundary condition equations can be found in Eq.~\eqref{bc_example2}. Table \ref{table:exps_doms_bcs} also shows the domain with the prescribed boundary conditions. We use the same PINN architectures, model parameters, and optimizers as the first experiment. We cannot train the PINN model with distance function (DF) for this experiment, as the boundary conditions are discontinuous and thus, a continuous global function using the coordinates for strong imposition of Dirichlet boundary conditions cannot be defined. For the collocation points, we generate the nodes from the mesh created using Gmsh, which is also used for finite element simulations in Abaqus to get the ground truth for this problem. The relative errors, $e$, for the three models, Soft, ADF, and PINN-FEM are shown in Table \ref{table:rel_error}. We observe that the proposed approach performs significantly better than the baseline models. This is further evident in Table \ref{fig:exp_circle_pred} which shows the distribution of the predicted response and the corresponding error compared to the ground truth for all the models. For both the baseline models, Soft and ADF, while the boundary conditions at the left half of the circular hole are accurately predicted, it fails to learn the displacement near the boundary where the boundary conditions change significantly. This is evident from the large errors near the right half of the circular hole, which are more pronounced for vanilla PINN  than the PINN with ADF. However, these errors are much lower for the proposed model.   

\begin{table}[!ht]
  \centering
    \caption{The distribution of predicted displacement fields, $\hat{u}_x$ and $\hat{u}_y$, and the corresponding error for the square elastic plate with a circular hole in the center, with uniform stretching force, under plane stress conditions, with boundary conditions as given in Eq.~\eqref{bc_example2}.}
  \begin{tabular}{ |c | c | c |c | c| }
    \hline
    PINN Model & $\hat{u}_x$ & $\hat{u}_y$ & $|\hat{u}_x-u_x|$ & $|\hat{u}_y-u_y|$ \\ \hline
    Ground Truth
    &
    \begin{minipage}{.19\textwidth}
      \includegraphics[width=0.9\linewidth]{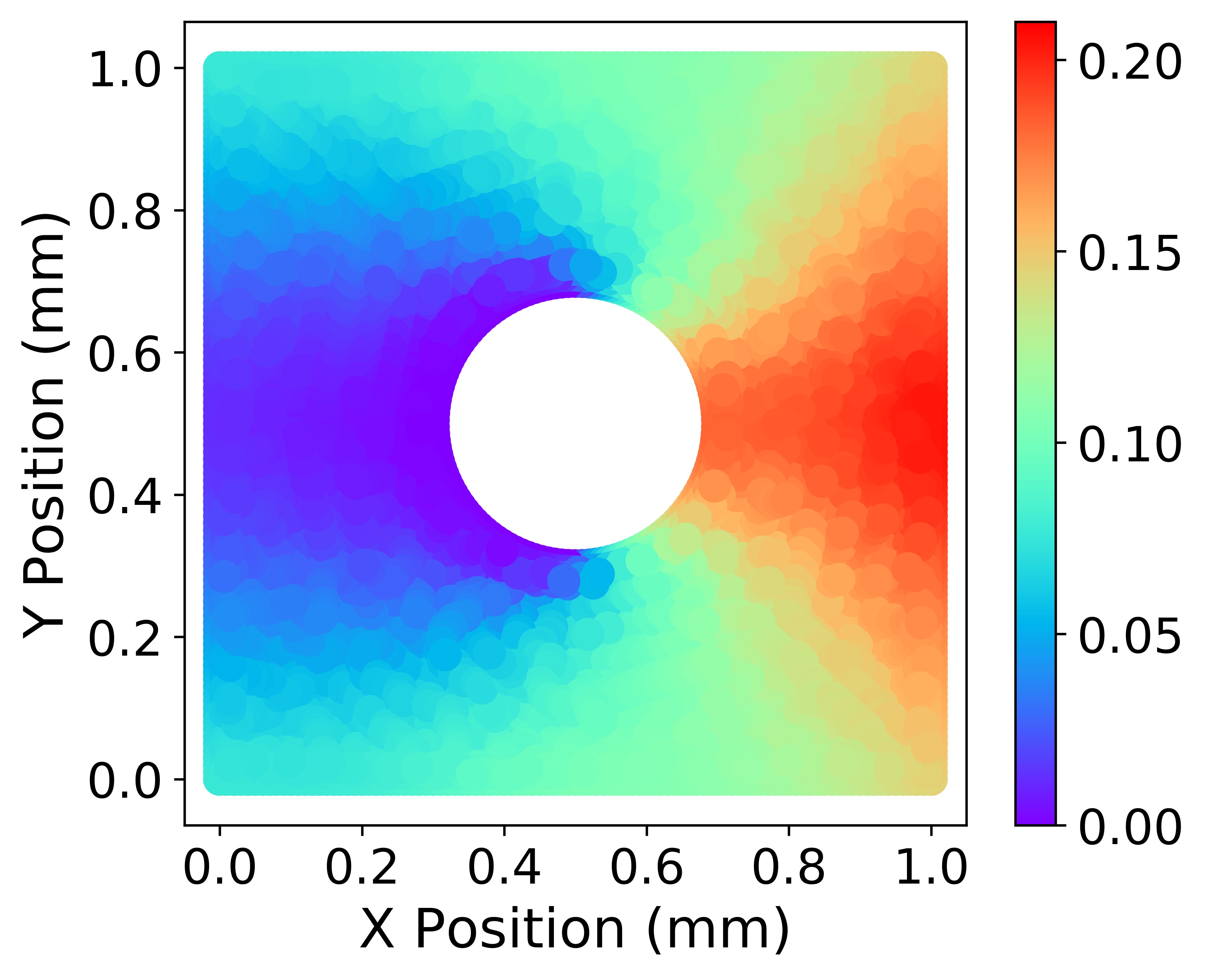}
    \end{minipage}
    &
   \begin{minipage}{.19\textwidth}
      \includegraphics[width=0.9\linewidth]{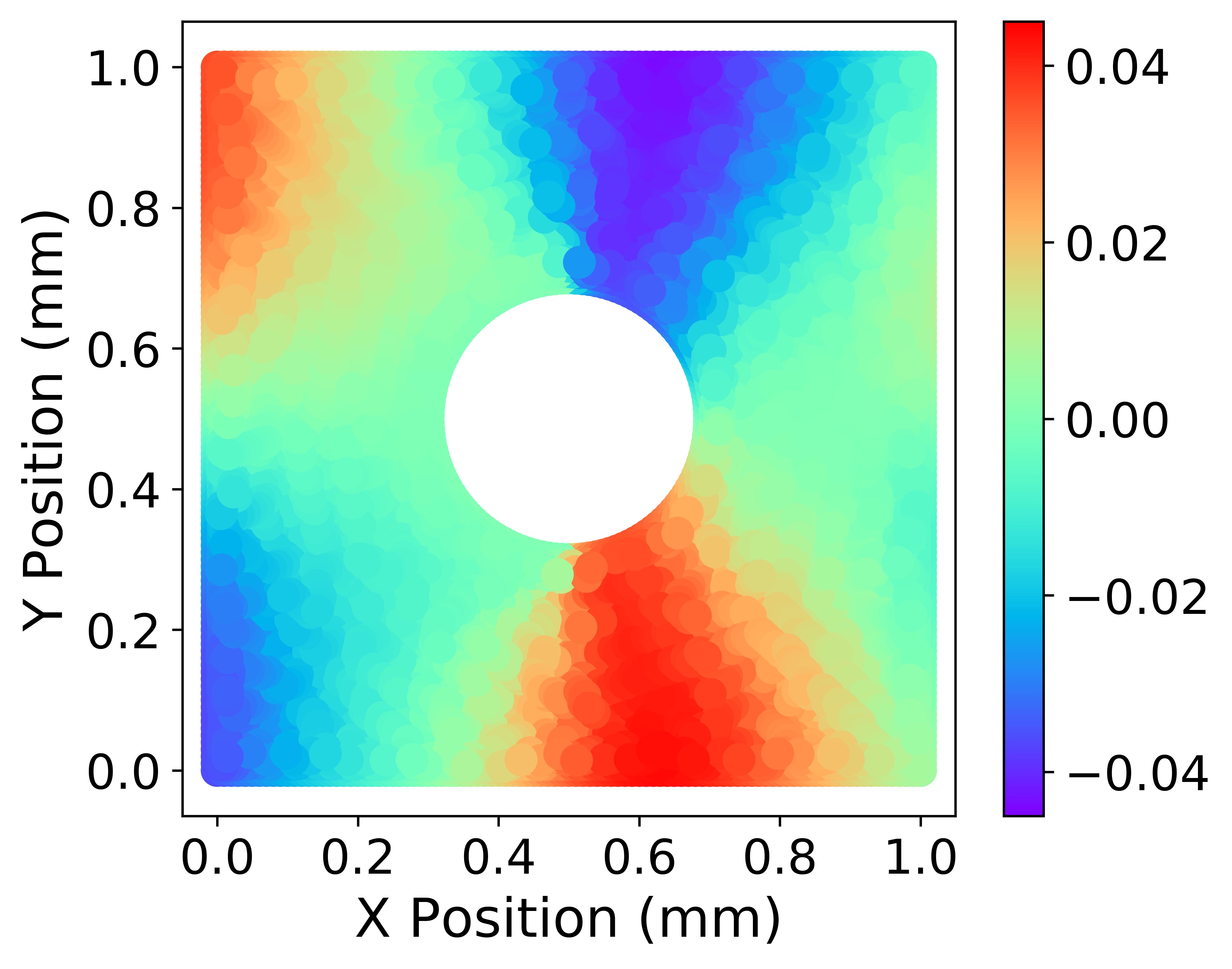}
    \end{minipage}
    & 
     & 
    \\ \hline
    Soft
    &
    \begin{minipage}{.19\textwidth}
      \includegraphics[width=0.9\linewidth]{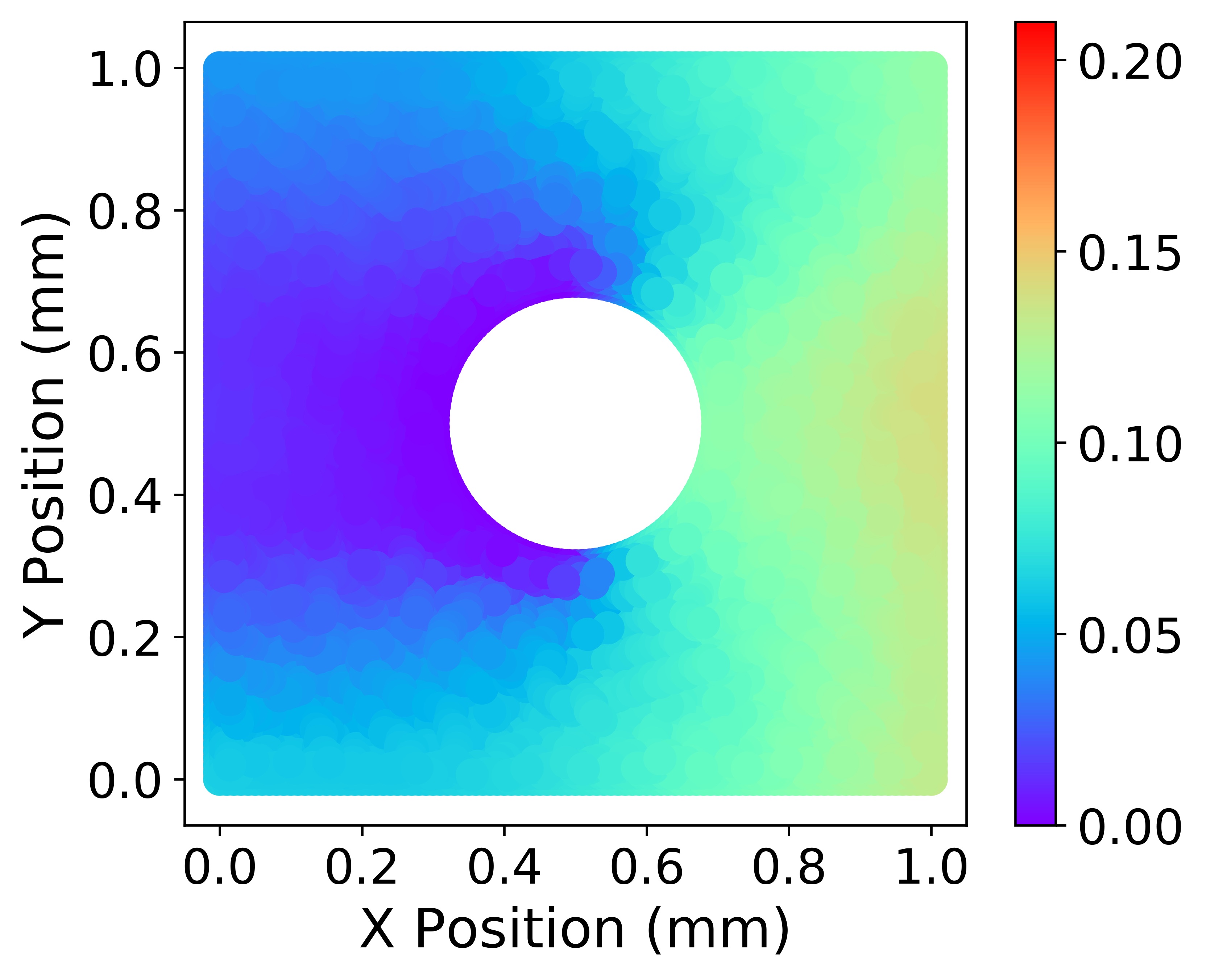}
    \end{minipage}
    &
   \begin{minipage}{.19\textwidth}
      \includegraphics[width=0.9\linewidth]{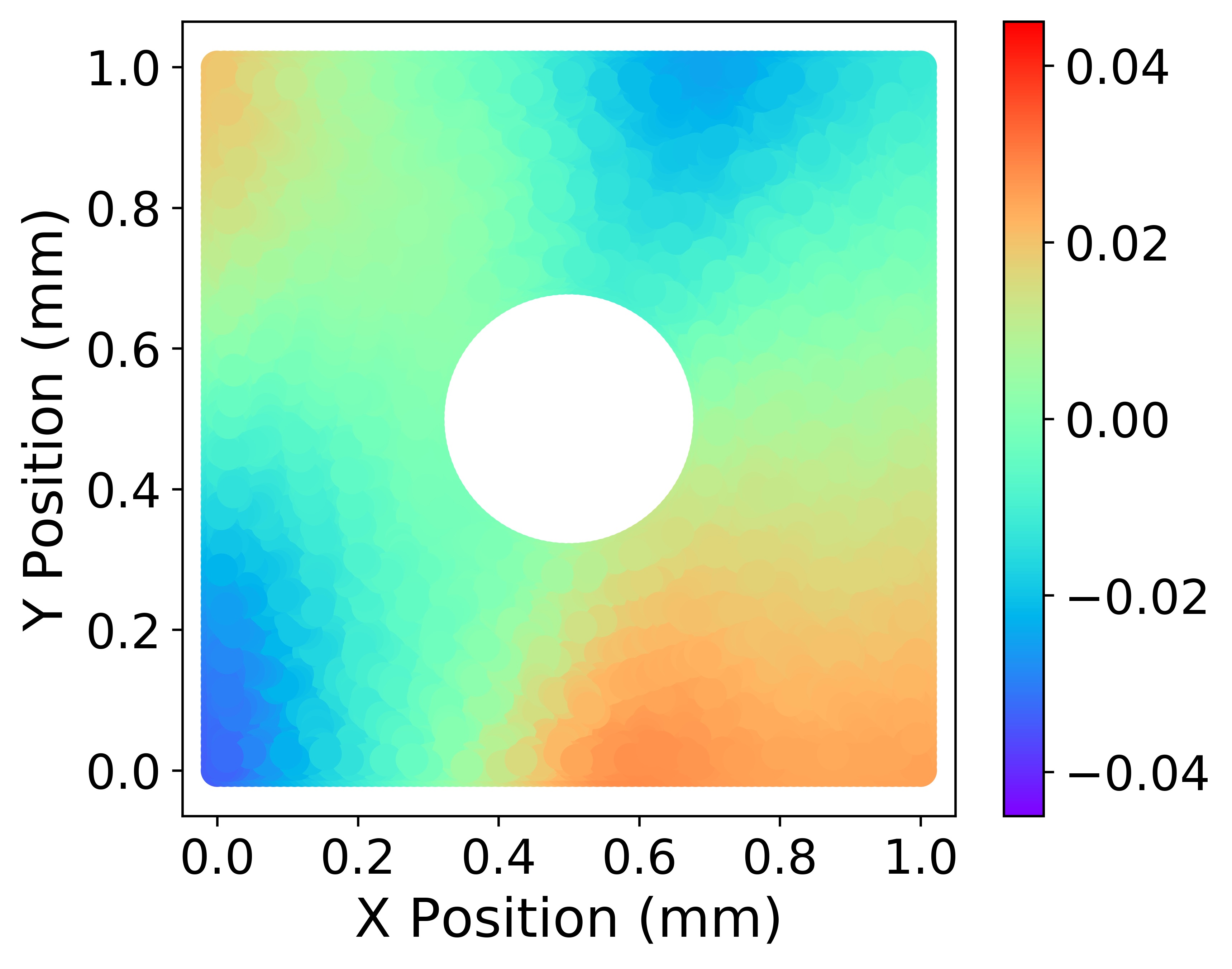}
    \end{minipage}
    & 
   \begin{minipage}{.19\textwidth}
      \includegraphics[width=0.9\linewidth]{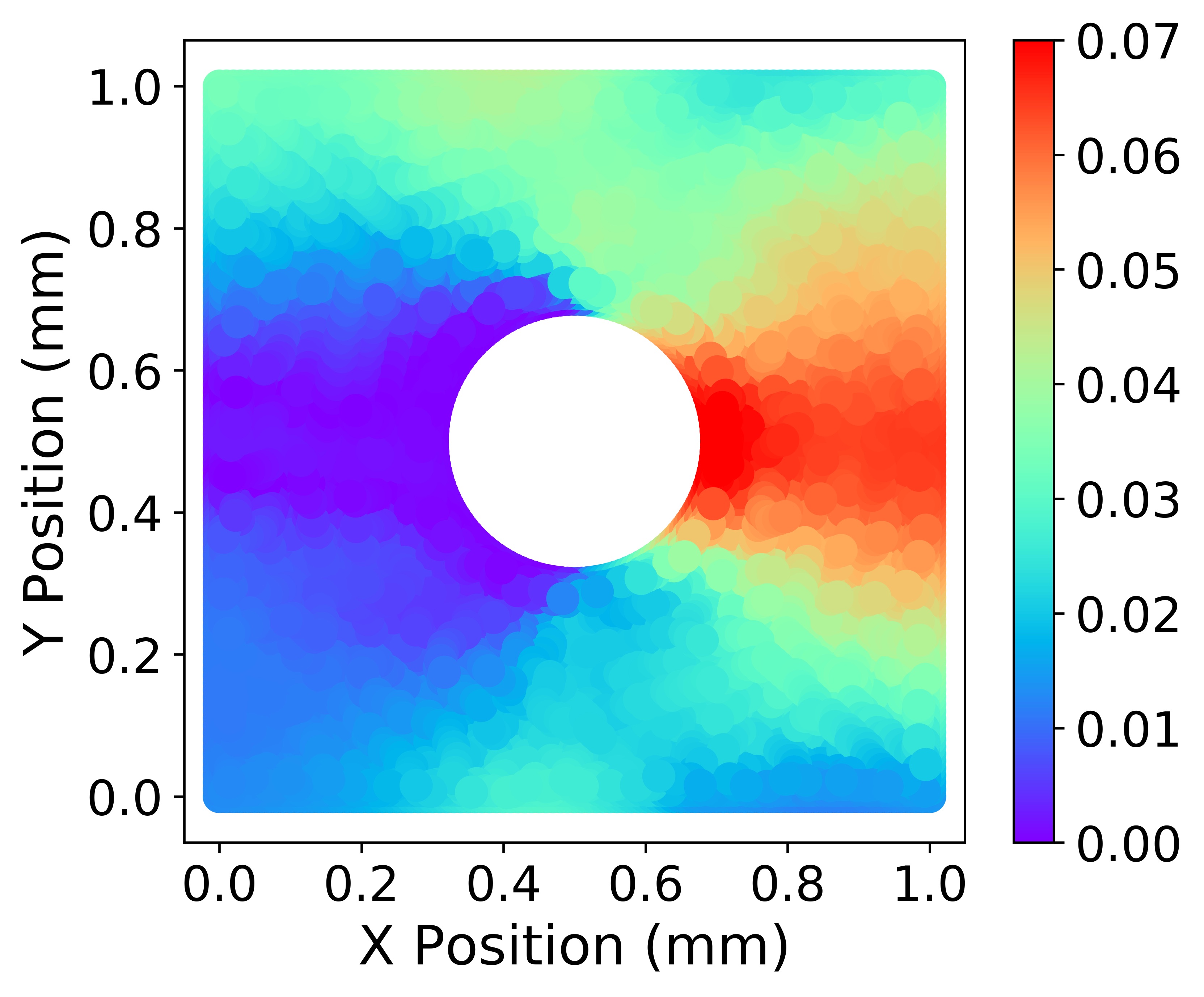}
    \end{minipage}
     & 
   \begin{minipage}{.19\textwidth}
      \includegraphics[width=0.9\linewidth]{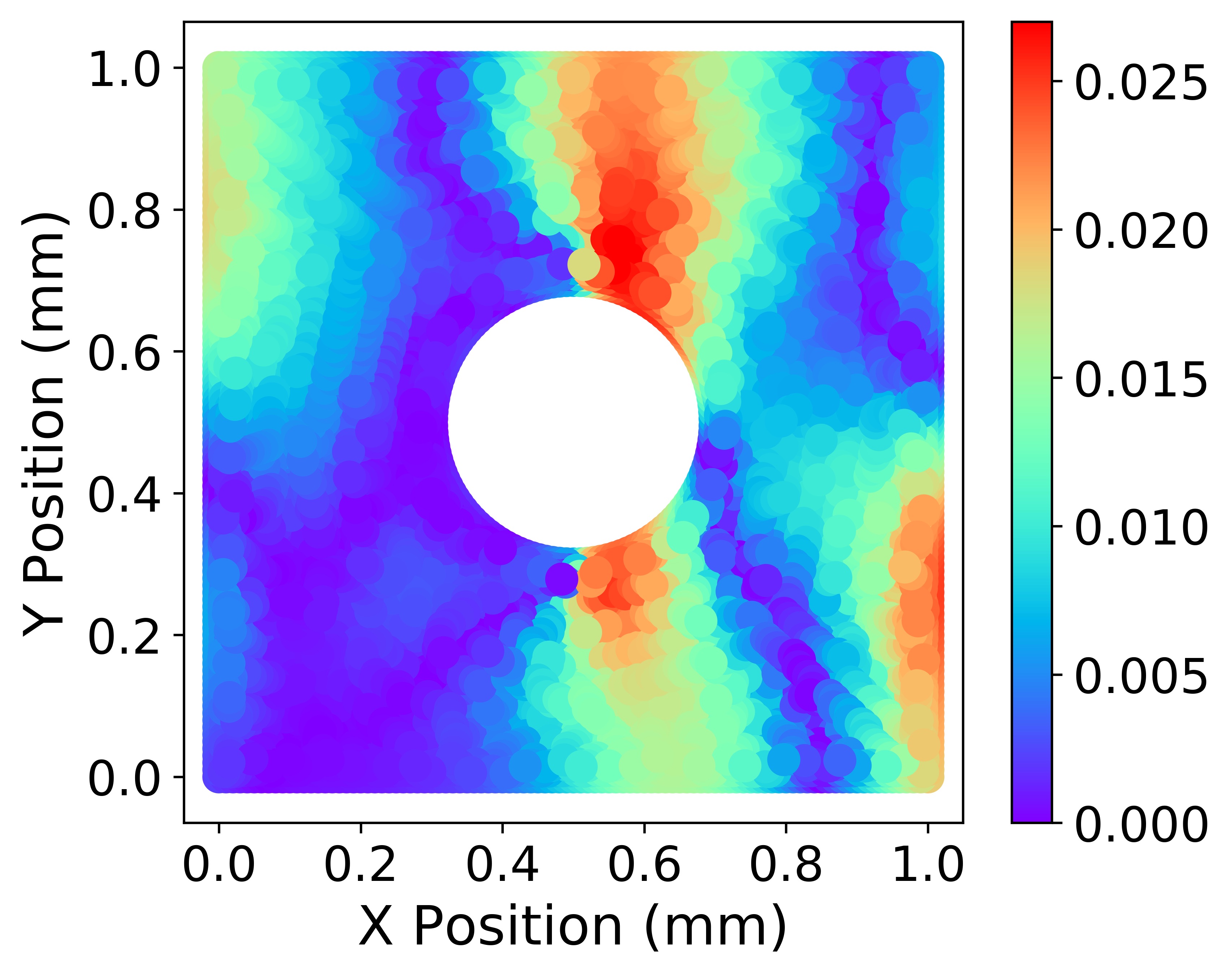}
    \end{minipage}
    \\ \hline
    ADF
    &
    \begin{minipage}{.19\textwidth}
      \includegraphics[width=0.9\linewidth]{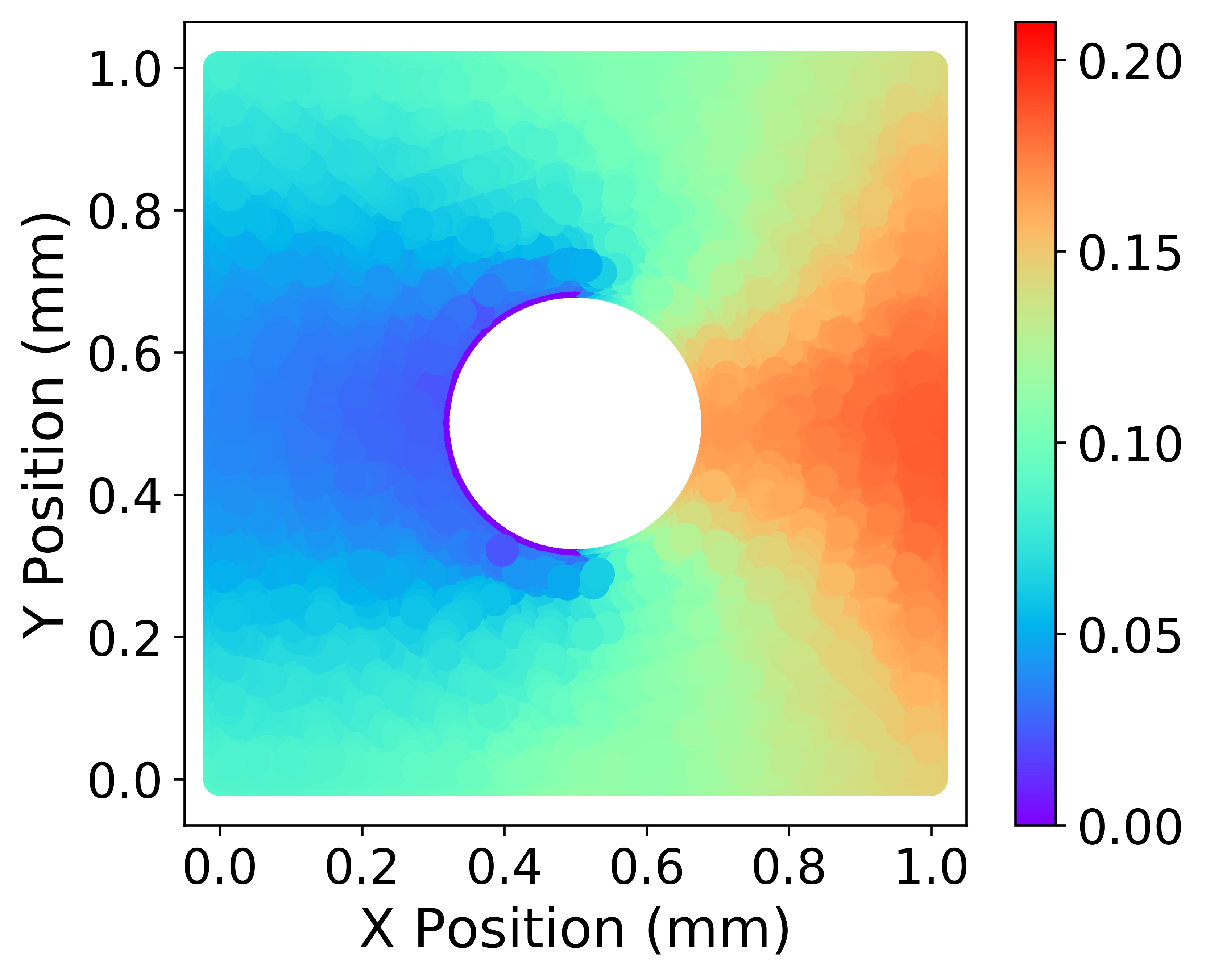}
    \end{minipage}
    &
   \begin{minipage}{.19\textwidth}
      \includegraphics[width=0.9\linewidth]{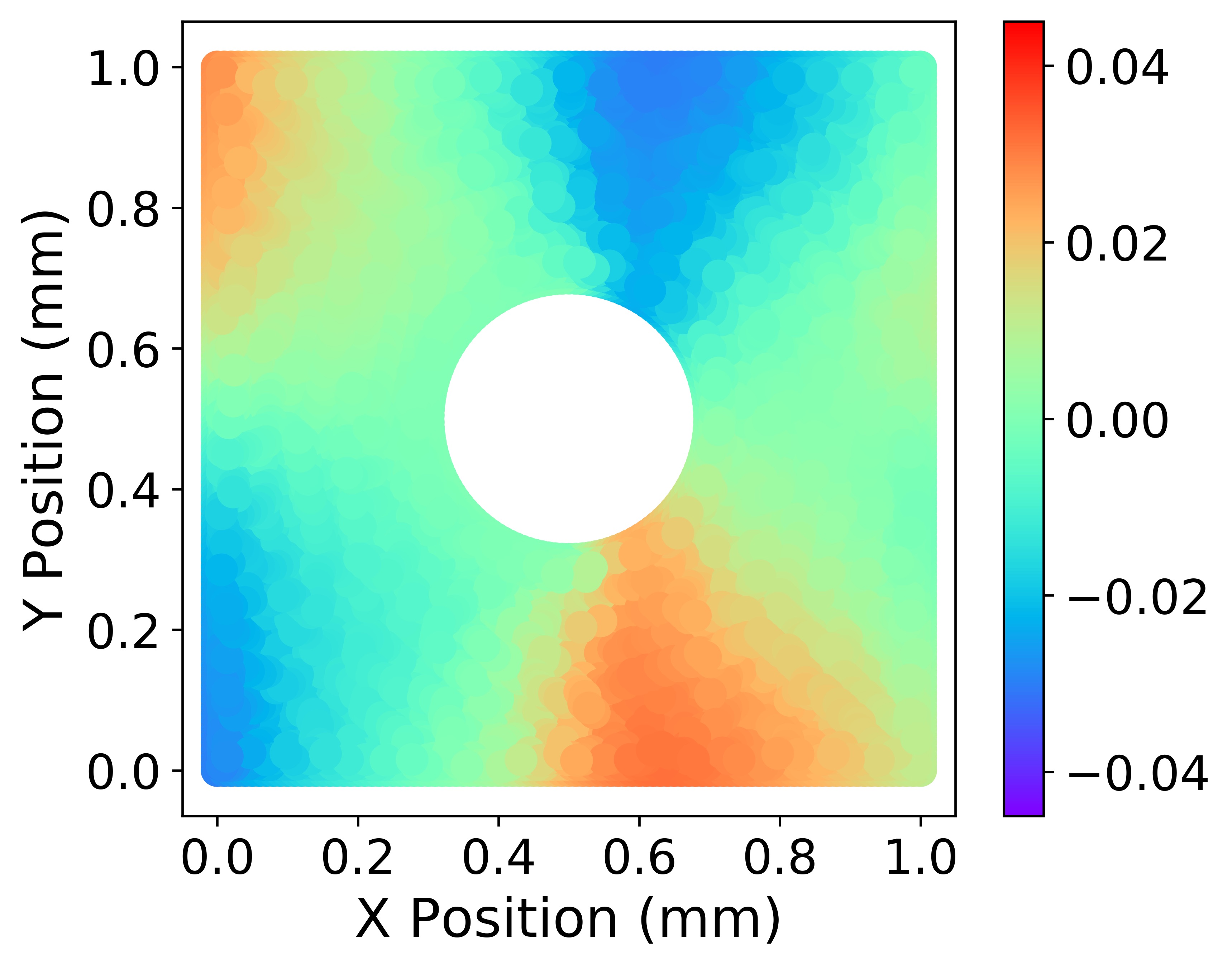}
    \end{minipage}
    & 
   \begin{minipage}{.19\textwidth}
      \includegraphics[width=0.9\linewidth]{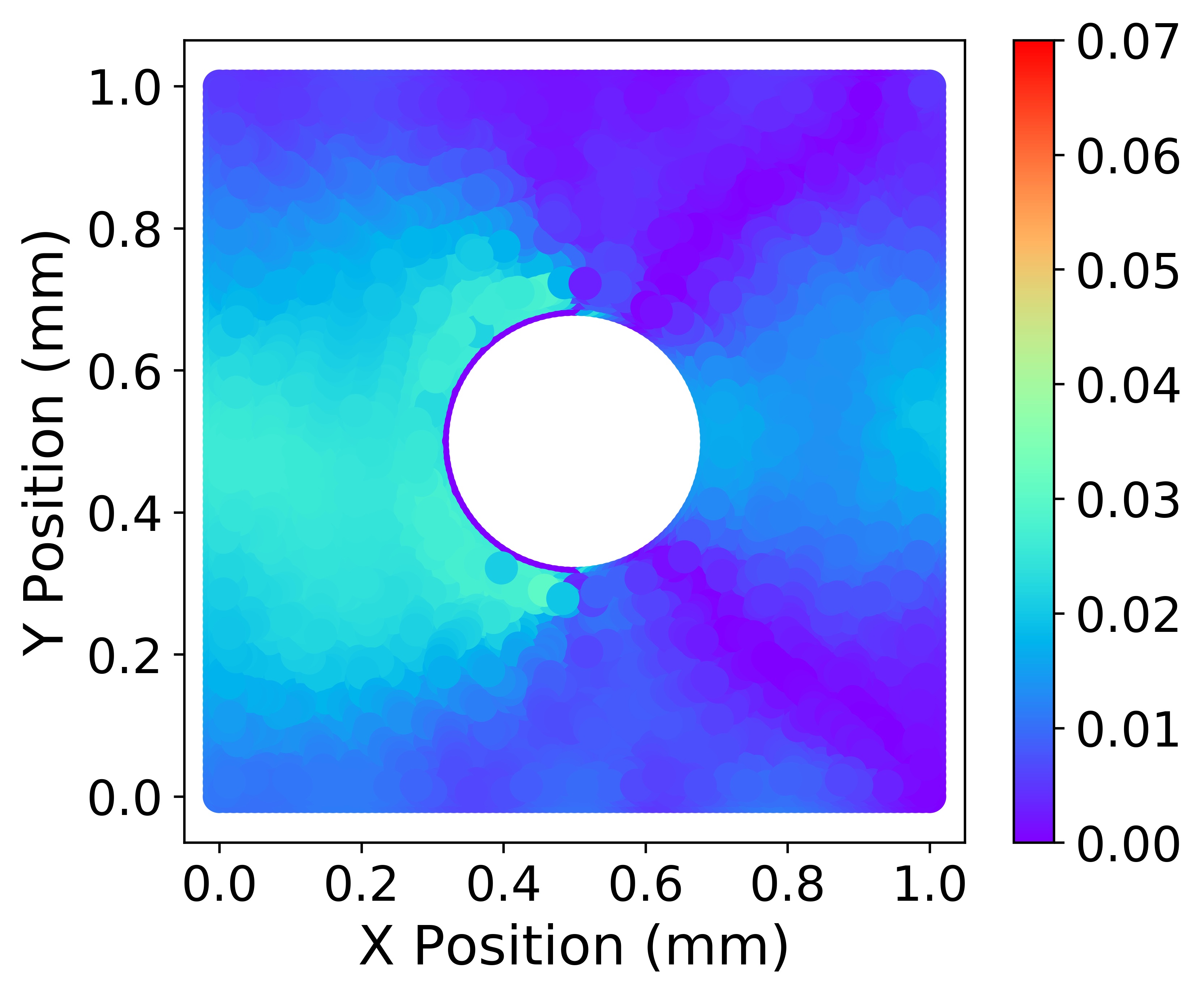}
    \end{minipage}
     & 
   \begin{minipage}{.19\textwidth}
      \includegraphics[width=0.9\linewidth]{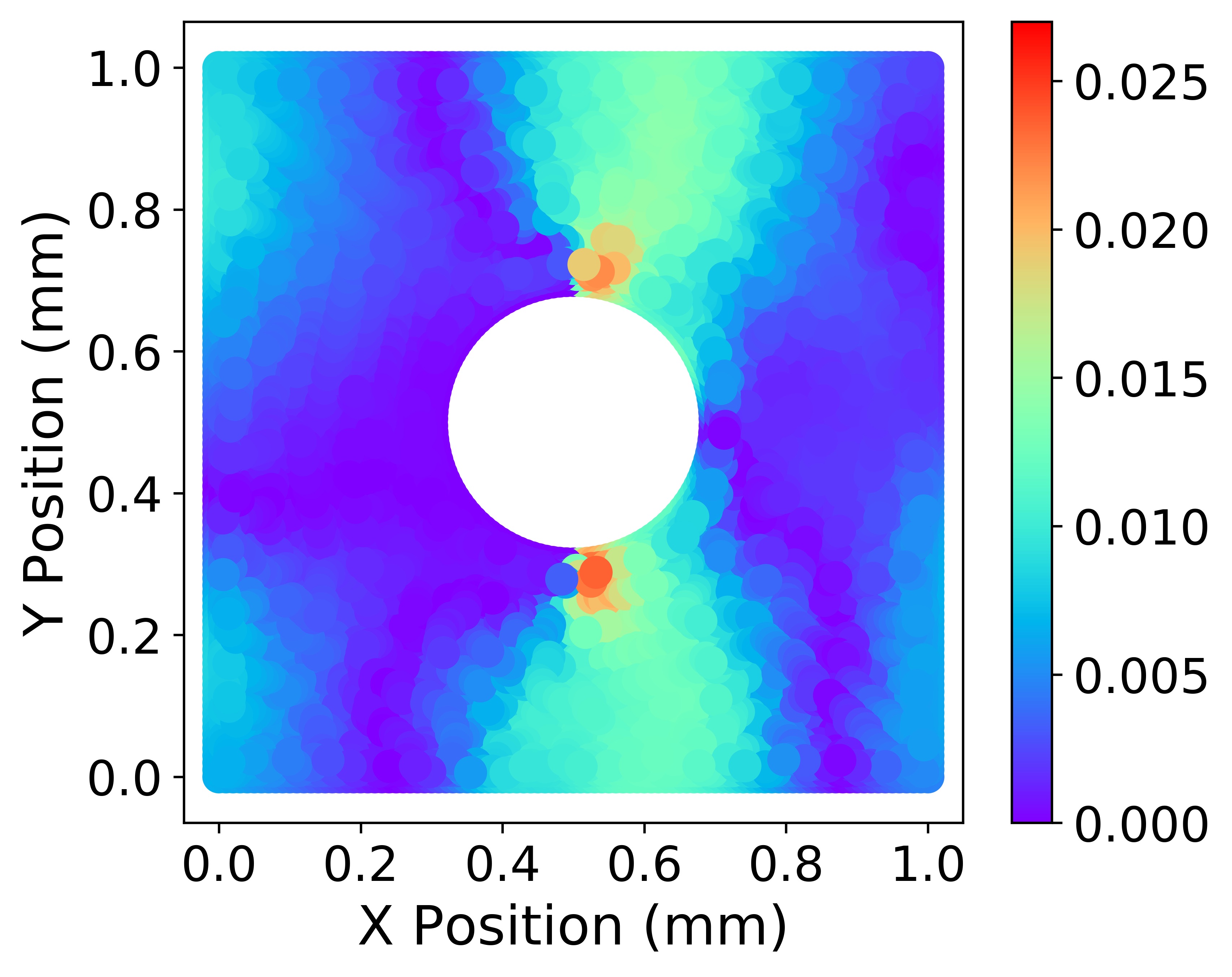}
    \end{minipage}
    \\ \hline
    PINN-FEM
    &
    \begin{minipage}{.19\textwidth}
      \includegraphics[width=0.9\linewidth]{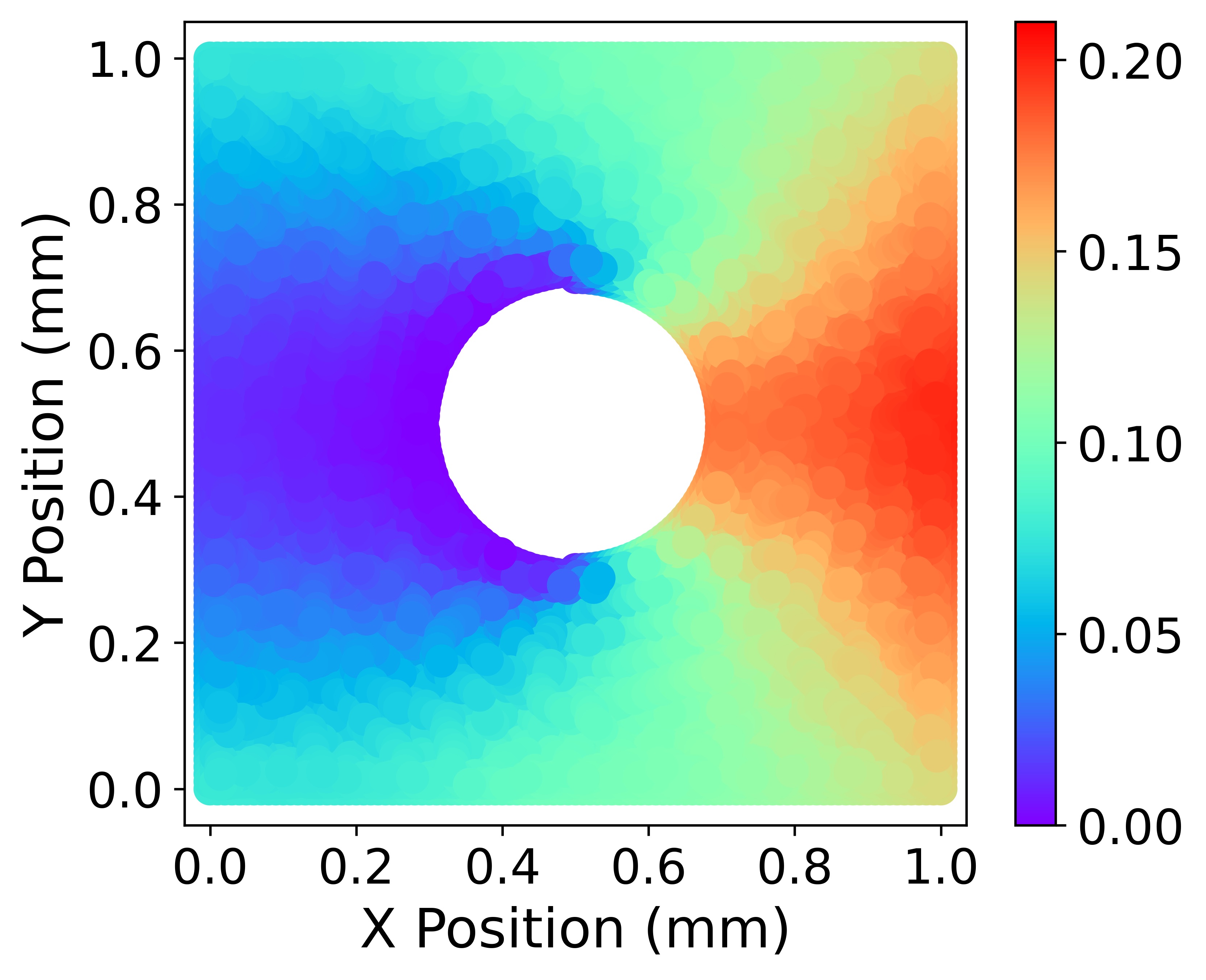}
    \end{minipage}
    &
   \begin{minipage}{.19\textwidth}
      \includegraphics[width=0.9\linewidth]{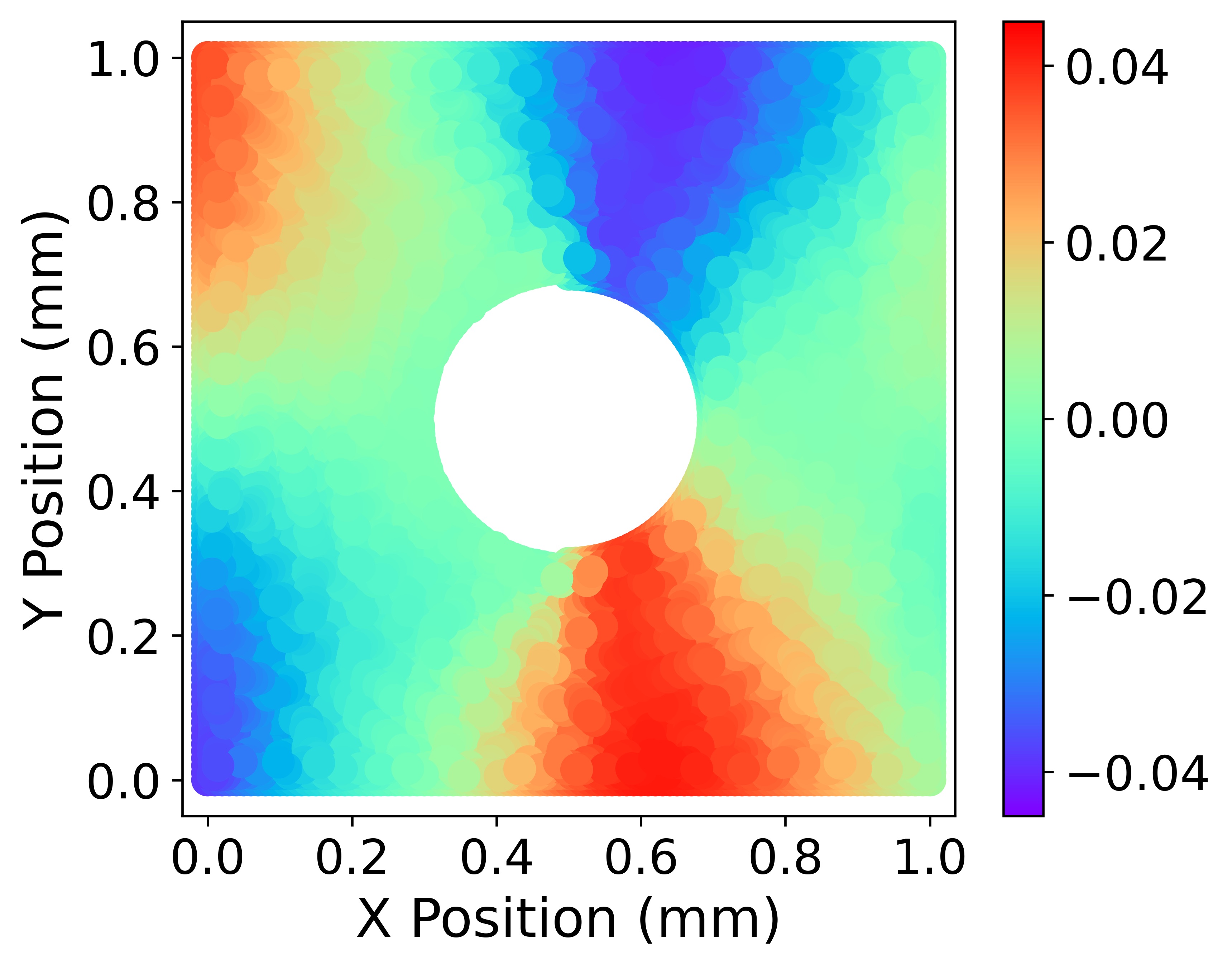}
    \end{minipage}
    & 
   \begin{minipage}{.19\textwidth}
      \includegraphics[width=0.9\linewidth]{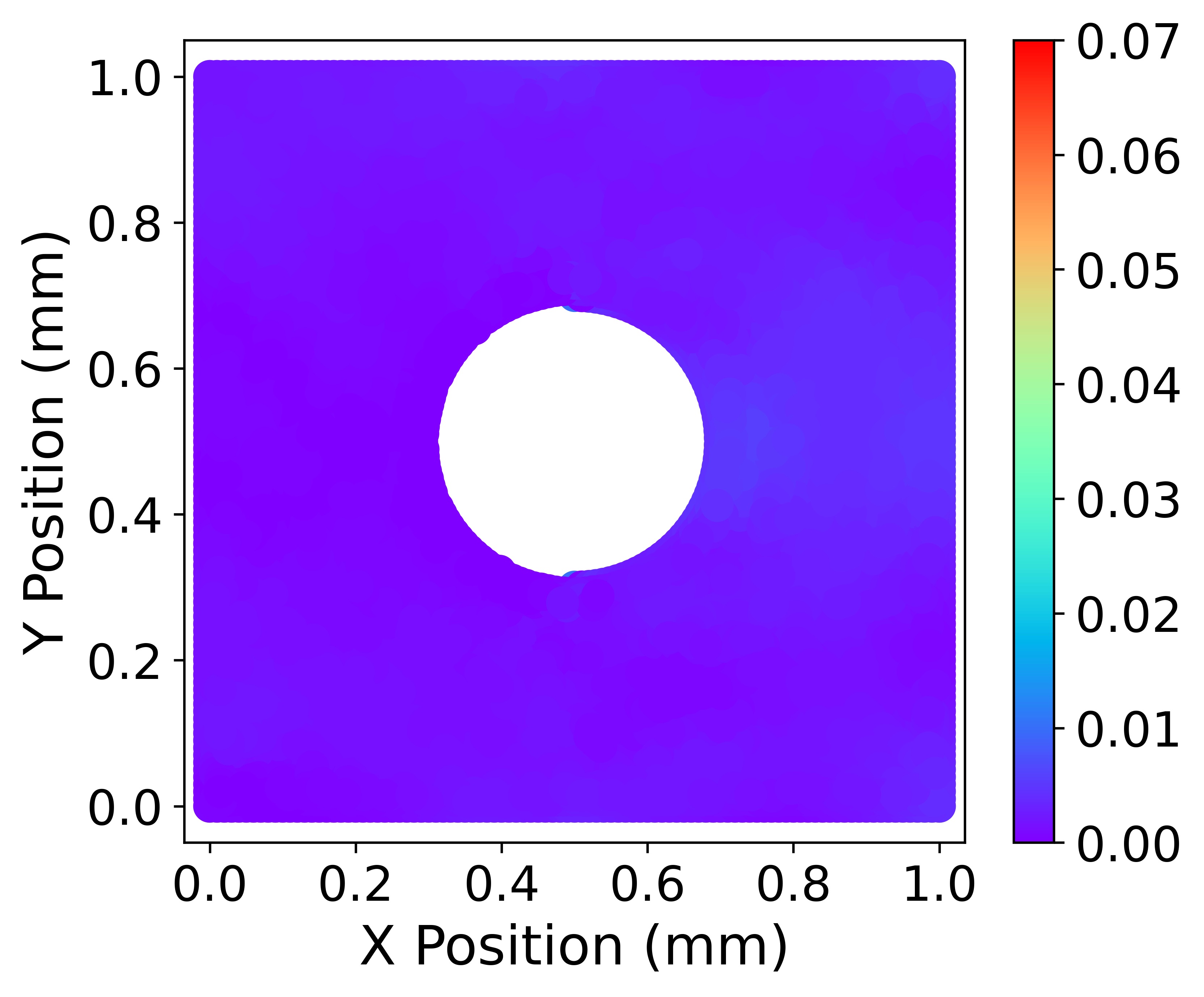}
    \end{minipage}
     & 
   \begin{minipage}{.19\textwidth}
      \includegraphics[width=0.9\linewidth]{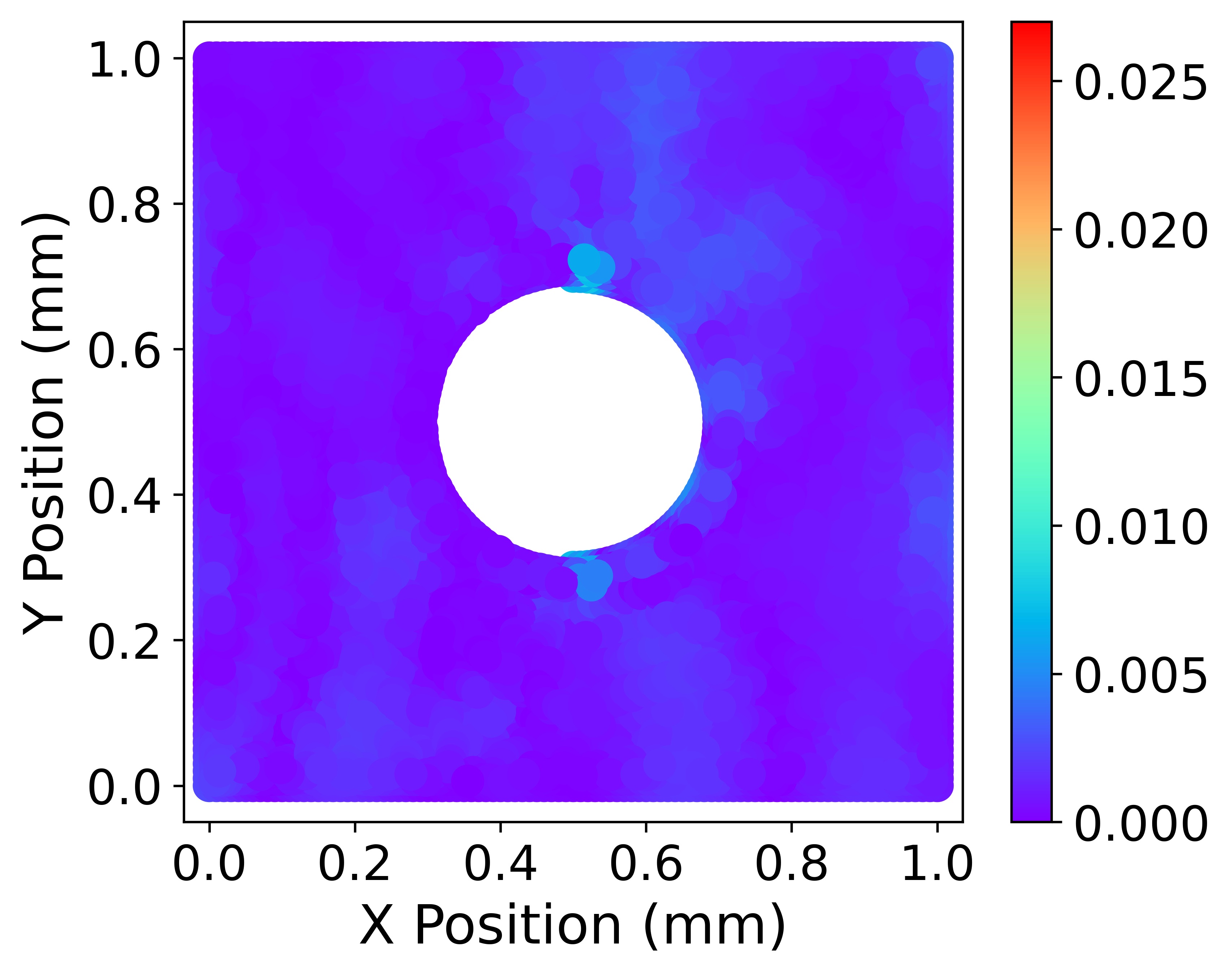}
    \end{minipage}
    \\ \hline
  \end{tabular}
  \label{fig:exp_circle_pred}
\end{table}

\subsubsection{Square plate with discontinuous boundaries}

In this example, we consider a square elastic plate similar to the first experiment, with a uniform stretching force of 10 units applied in the x-direction. Fixed boundary conditions are applied on just the lower half of the left edge of the square. The detailed boundary condition equations are given in Eq.~\eqref{bc_example3}. Table \ref{table:exps_doms_bcs} also shows the domain with the prescribed boundary conditions. We use the same PINN architectures, model parameters and optimizers as the first experiment. Similar to the previous example, we do not consider the benchmark model, PINN with DF, for this experiment as the boundary conditions are discontinuous. As in the previous experiments, the mesh generated using Gmsh is used for the collocation points and for finite element simulations in Abaqus to get the ground truth for this problem. The relative errors, $e$, for the three models-Soft, ADF, and PINN-FEM-are shown in Table \ref{table:rel_error}. We observe that the proposed approach performs significantly better than the baseline models. This is also evident in Table \ref{fig:exp_halfbc_pred}, which shows the distribution of the predicted response and the corresponding error compared  to the ground truth for all the models. For both the baseline models, Soft and ADF, while the fixed boundary conditions at the lower half of the left edge are accurately predicted, they fail to learn the displacement for points near the boundary where the boundary conditions change significantly. This is evident from the large errors near the middle of the left edge. However, the proposed model is able to accurately predicts the significant changes in the boundary conditions, as evident by the low error values, especially near the points where the boundary condition changes on the left edge.

\begin{table}[!ht]
  \centering
    \caption{The distribution of predicted displacement fields, $\hat{u}_x$ and $\hat{u}_y$, and the corresponding error, for the square elastic plate with discontinuous boundaries, with fixed Dirichlet boundary conditions for just the lower half of the left edge, as given in Eq.~\eqref{bc_example3}.}
  \begin{tabular}{ |c | c | c |c | c| }
    \hline
    PINN Model & $\hat{u}_x$ & $\hat{u}_y$ & $|\hat{u}_x-u_x|$ & $|\hat{u}_y-u_y|$ \\ \hline
    Ground Truth
    &
    \begin{minipage}{.19\textwidth}
      \includegraphics[width=0.9\linewidth]{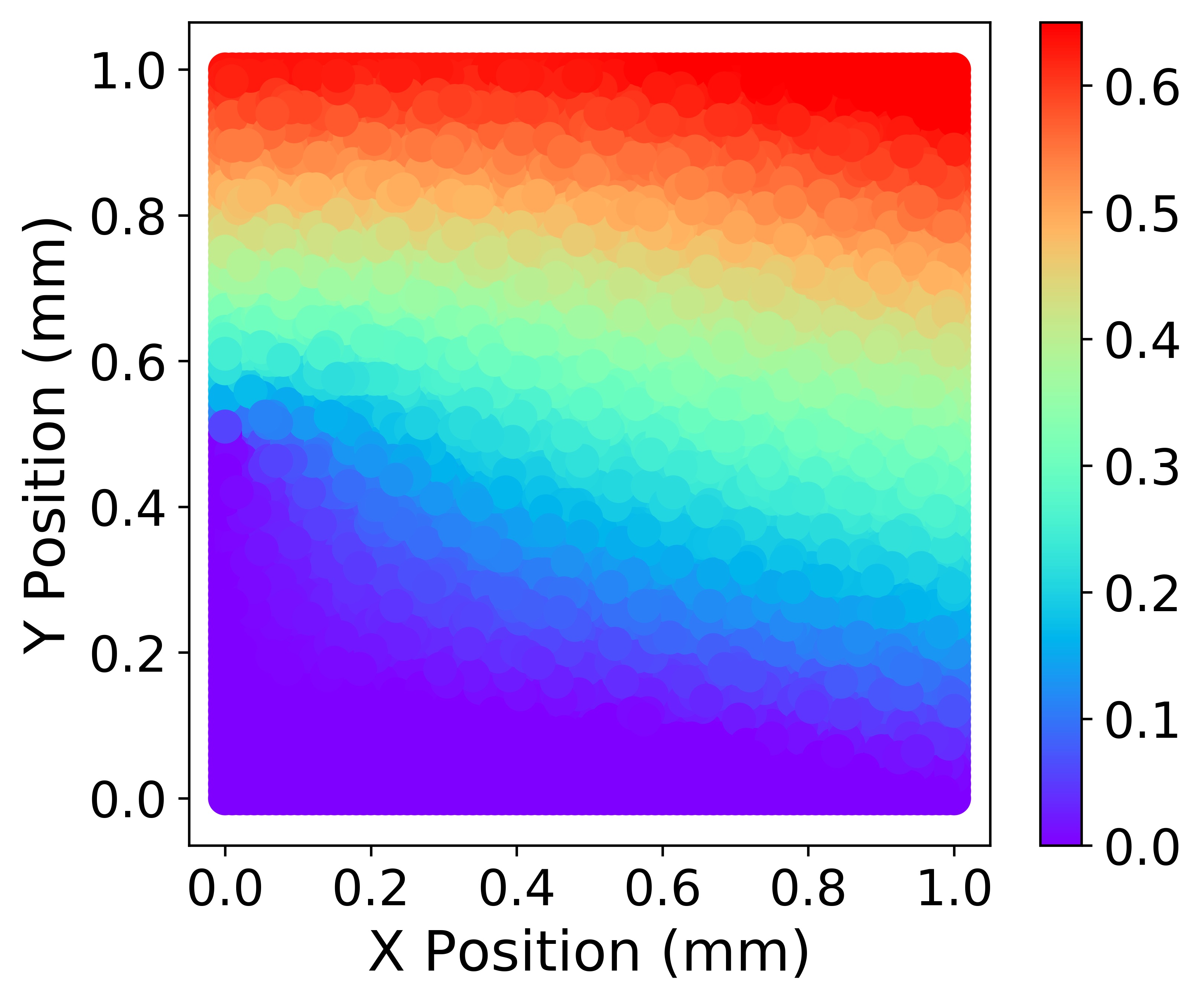}
    \end{minipage}
    &
   \begin{minipage}{.19\textwidth}
      \includegraphics[width=0.9\linewidth]{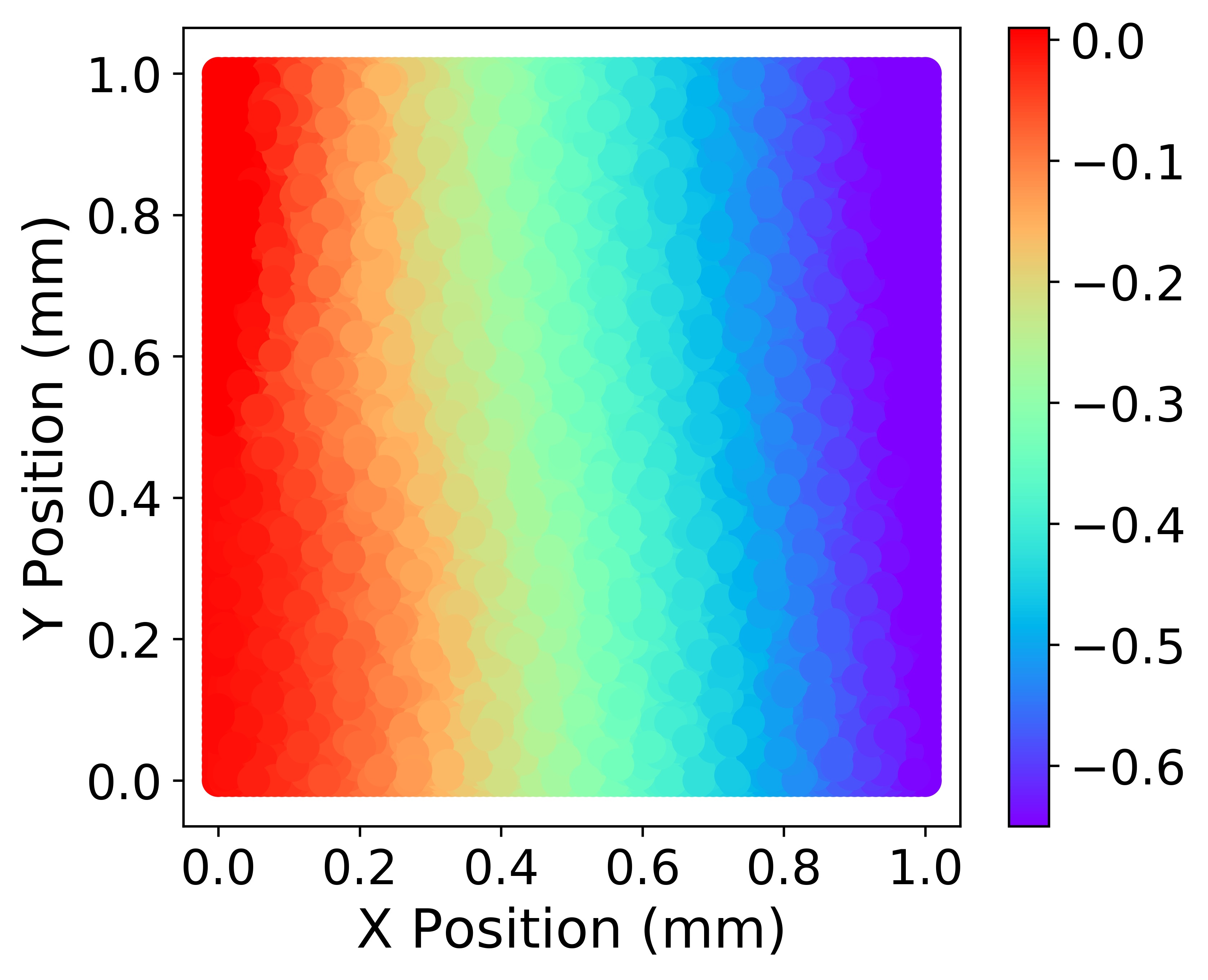}
    \end{minipage}
    & 
     & 
    \\ \hline
    Soft
    &
    \begin{minipage}{.19\textwidth}
      \includegraphics[width=0.9\linewidth]{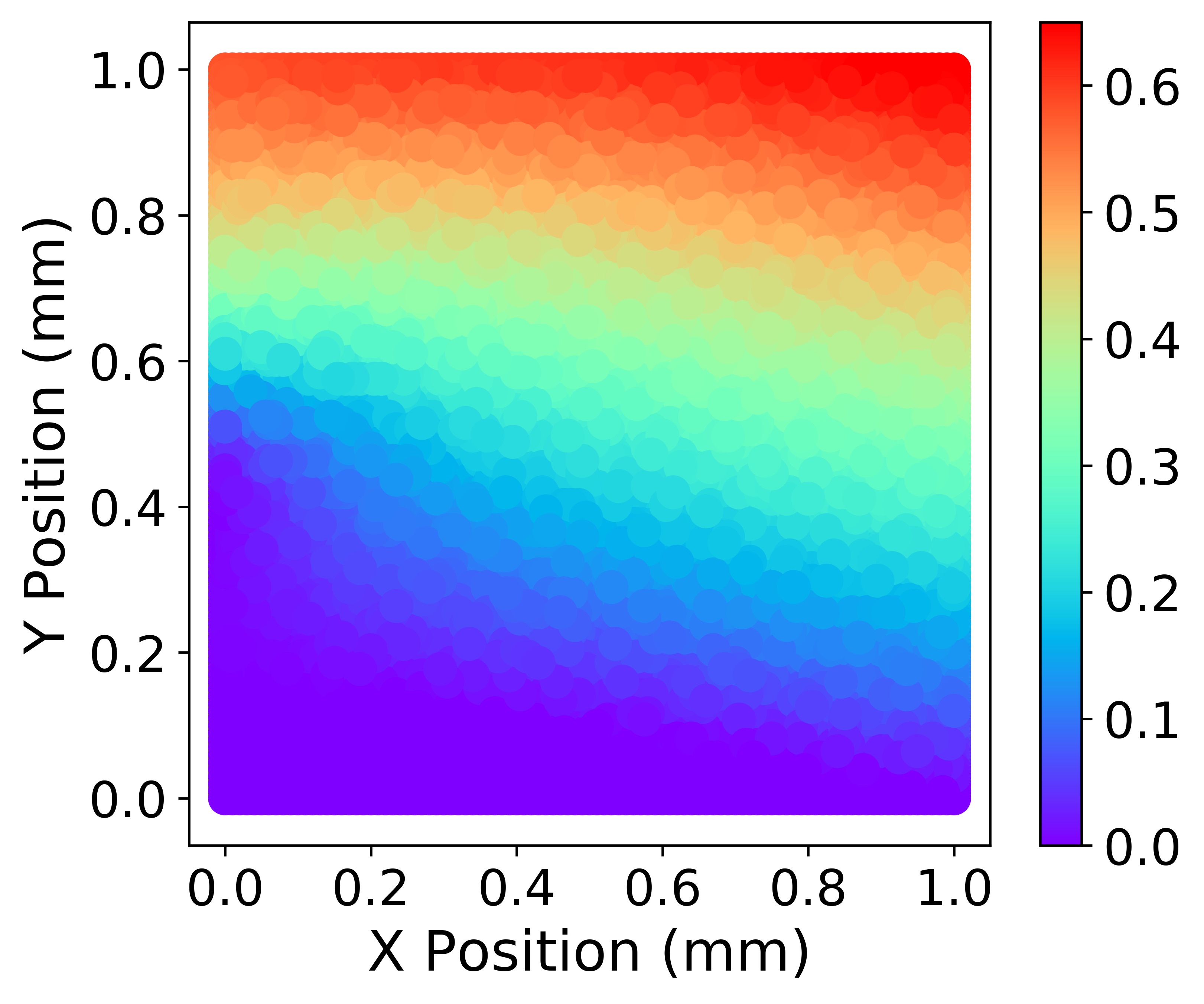}
    \end{minipage}
    &
   \begin{minipage}{.19\textwidth}
      \includegraphics[width=0.9\linewidth]{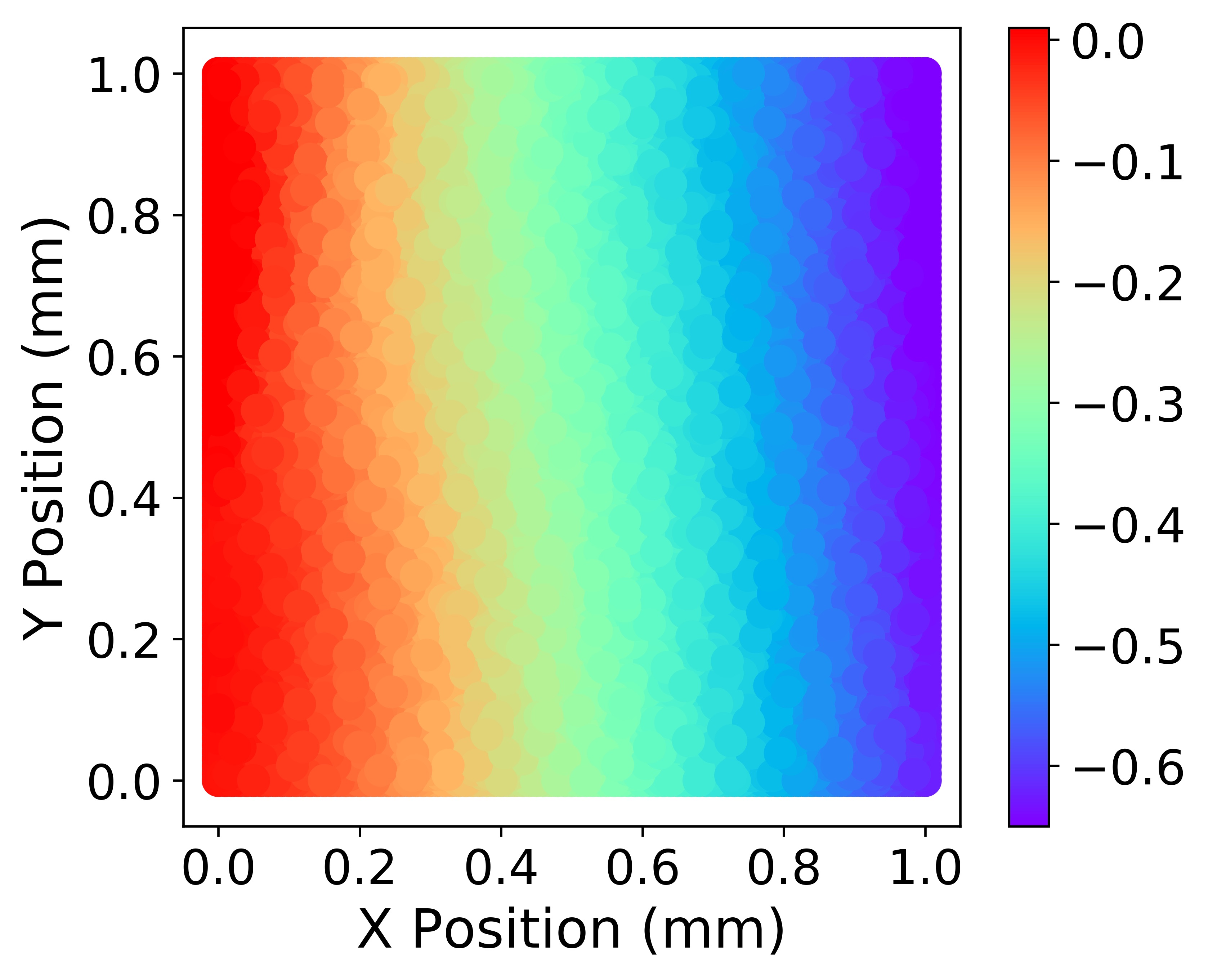}
    \end{minipage}
    & 
   \begin{minipage}{.19\textwidth}
      \includegraphics[width=0.9\linewidth]{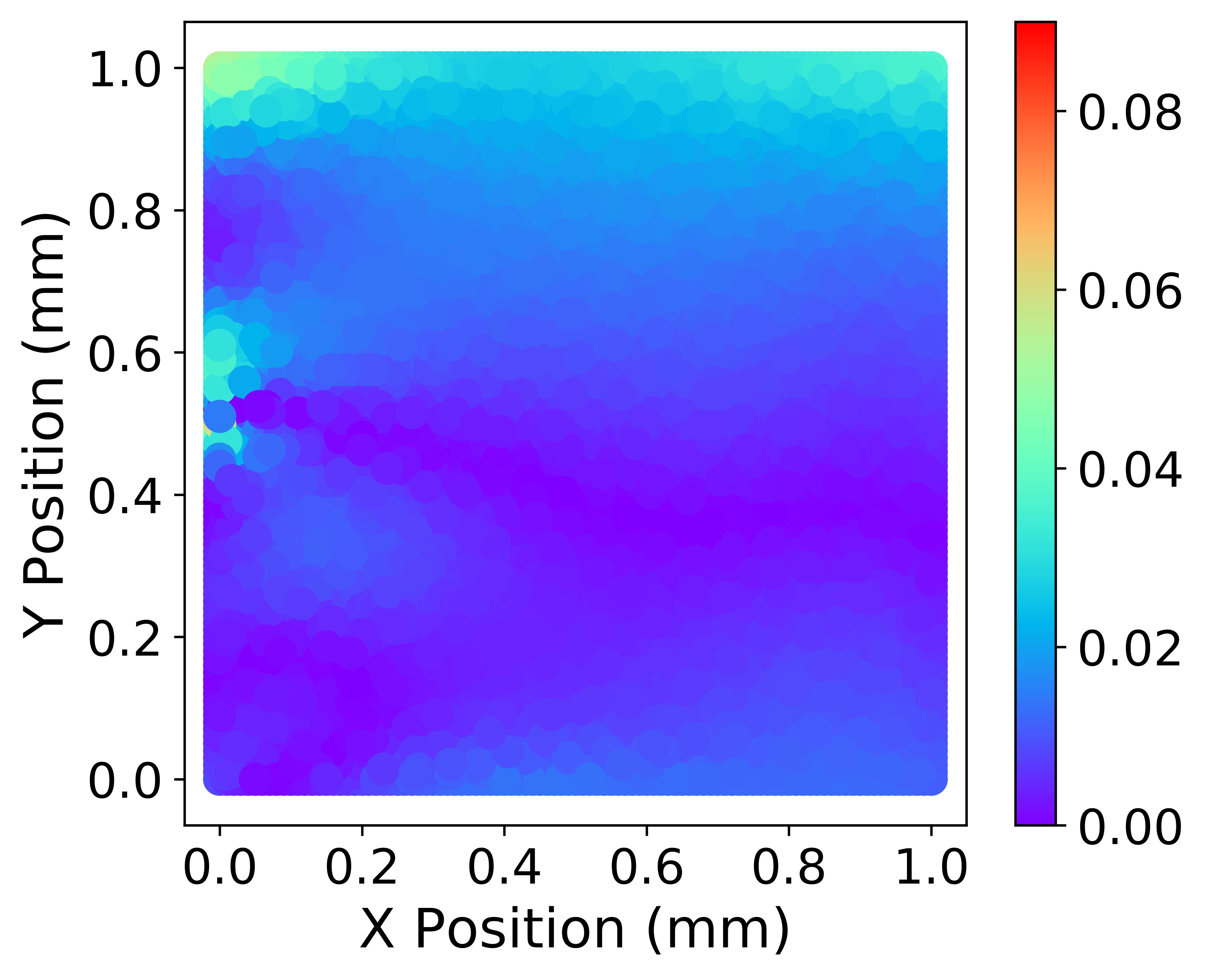}
    \end{minipage}
     & 
   \begin{minipage}{.19\textwidth}
      \includegraphics[width=0.9\linewidth]{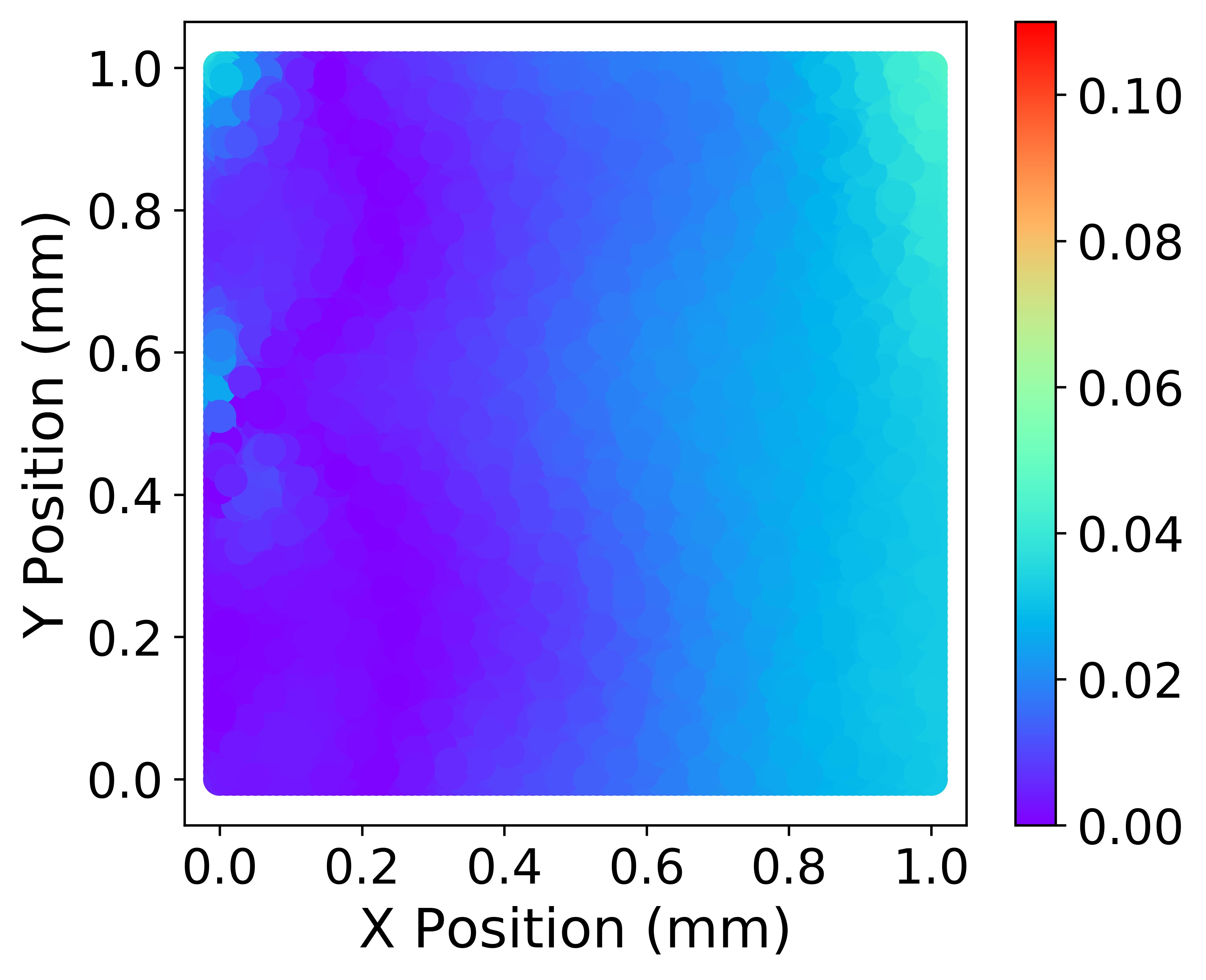}
    \end{minipage}
    \\ \hline
    ADF
    &
    \begin{minipage}{.19\textwidth}
      \includegraphics[width=0.9\linewidth]{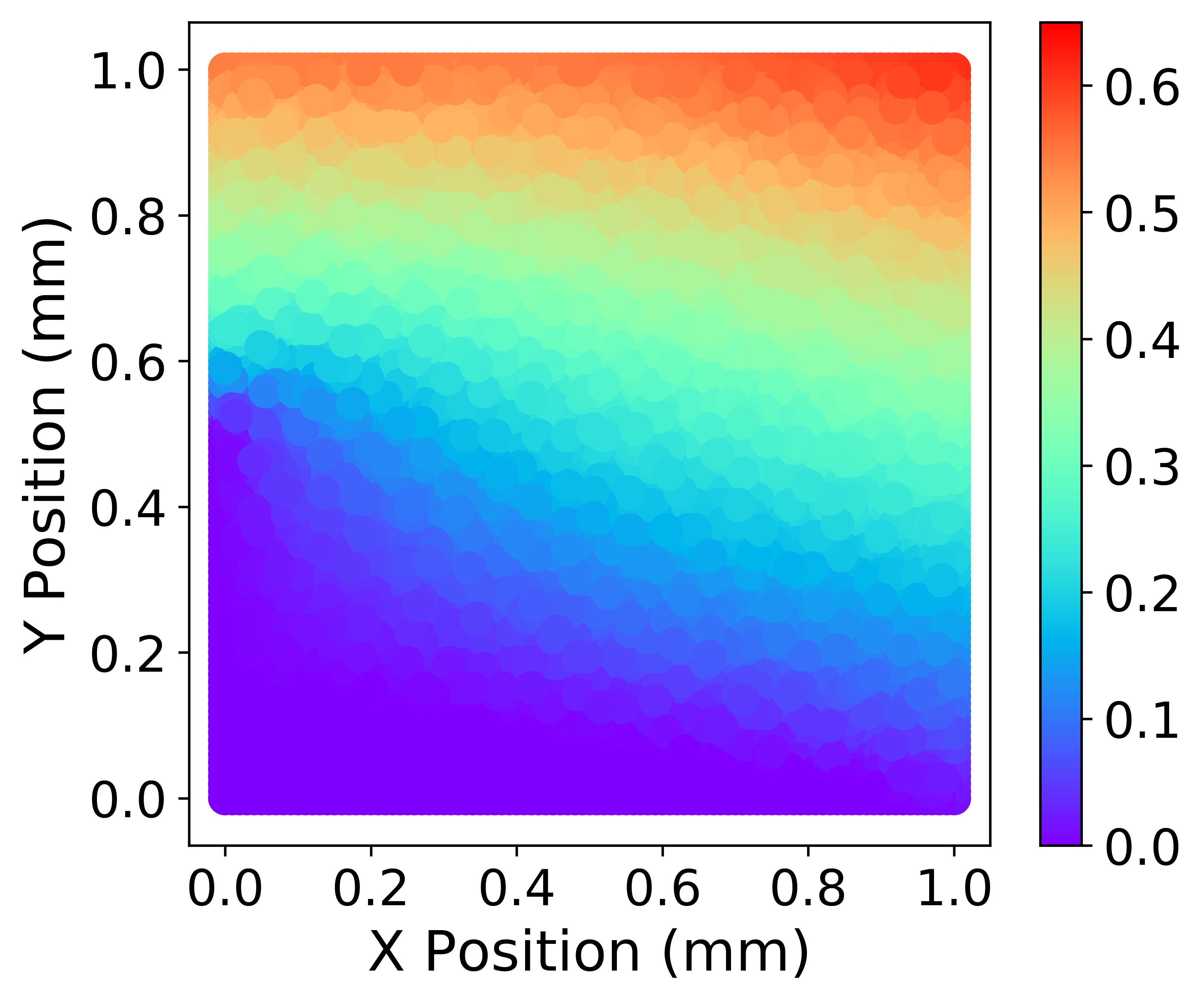}
    \end{minipage}
    &
   \begin{minipage}{.19\textwidth}
      \includegraphics[width=0.9\linewidth]{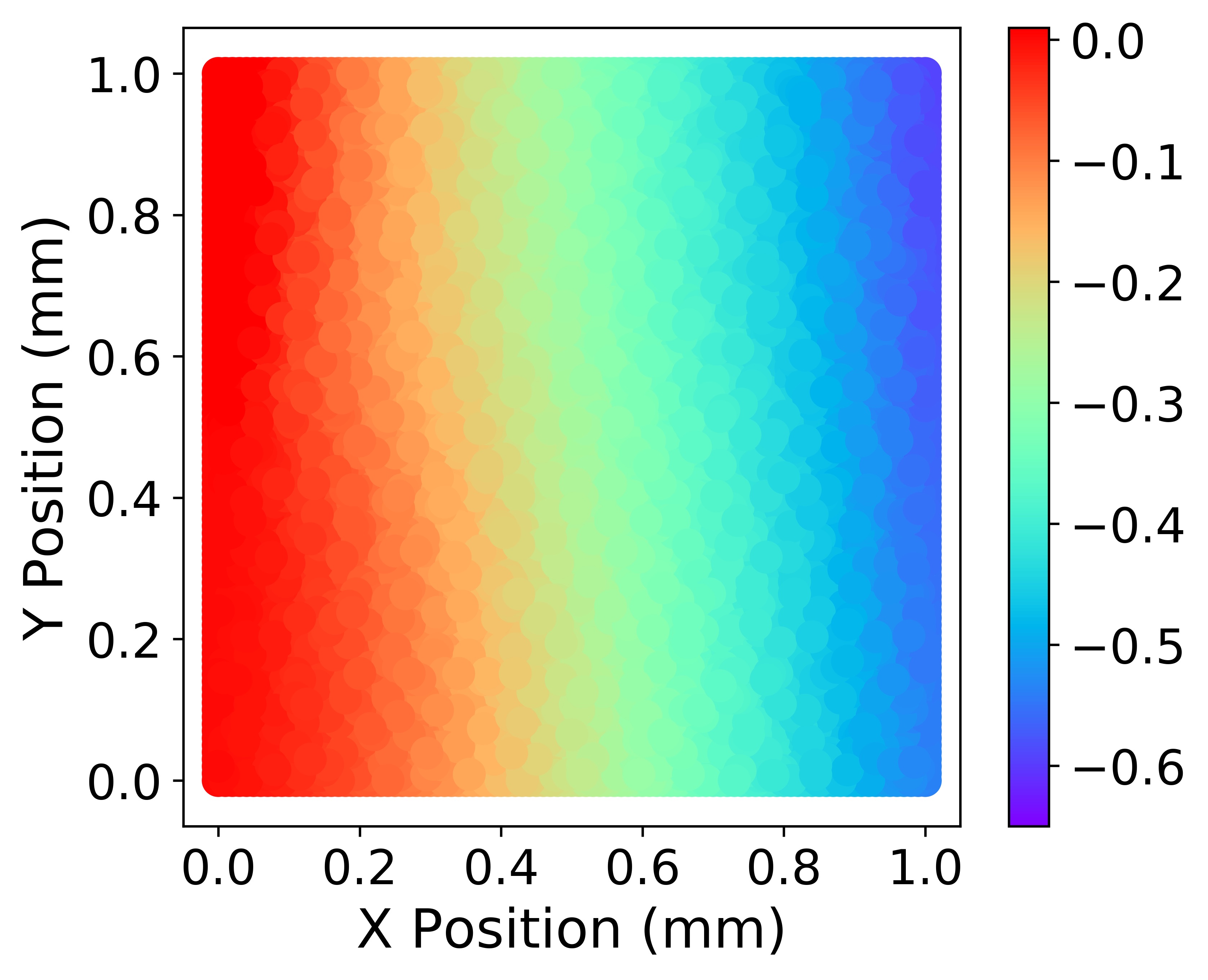}
    \end{minipage}
    & 
   \begin{minipage}{.19\textwidth}
      \includegraphics[width=0.9\linewidth]{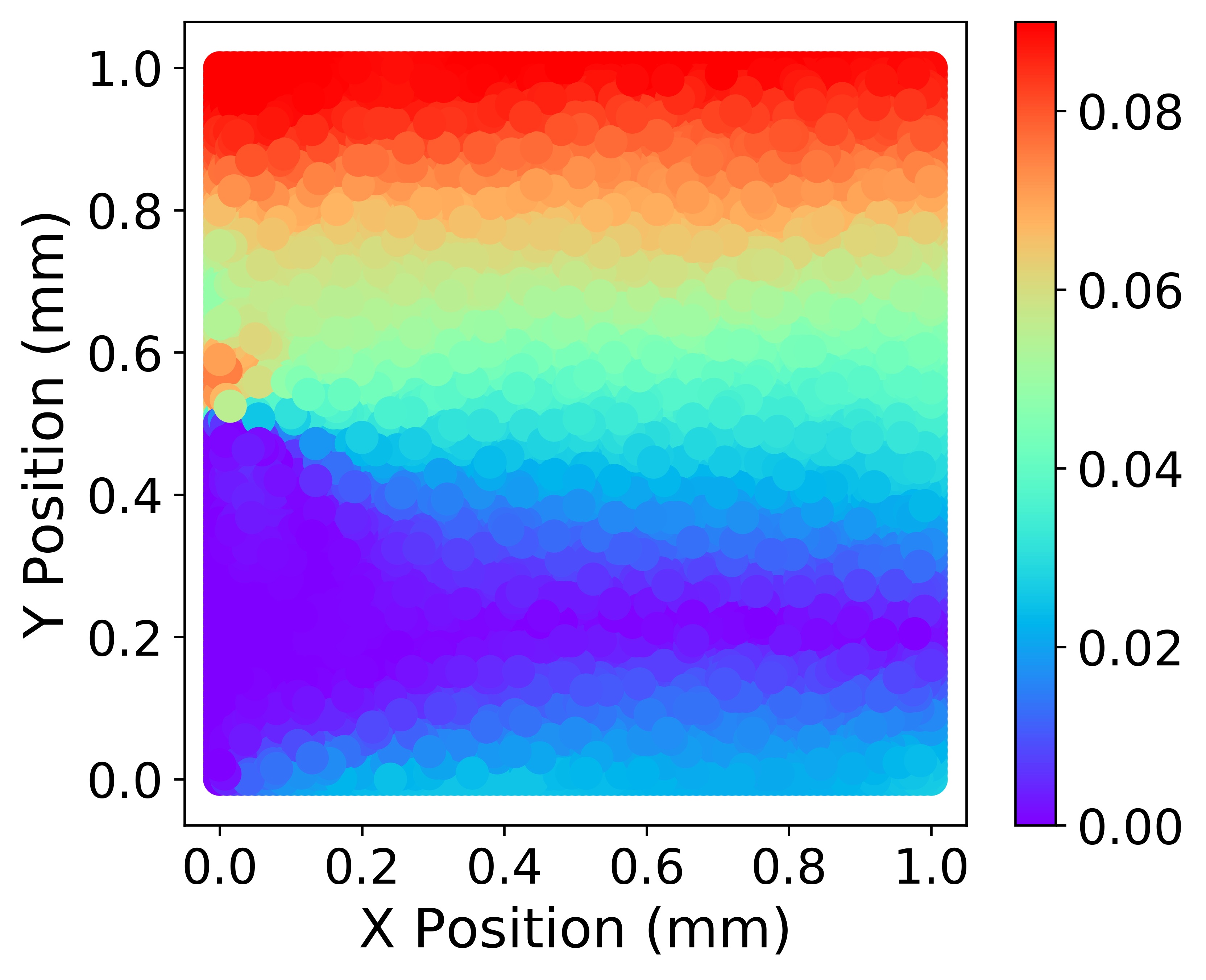}
    \end{minipage}
     & 
   \begin{minipage}{.19\textwidth}
      \includegraphics[width=0.9\linewidth]{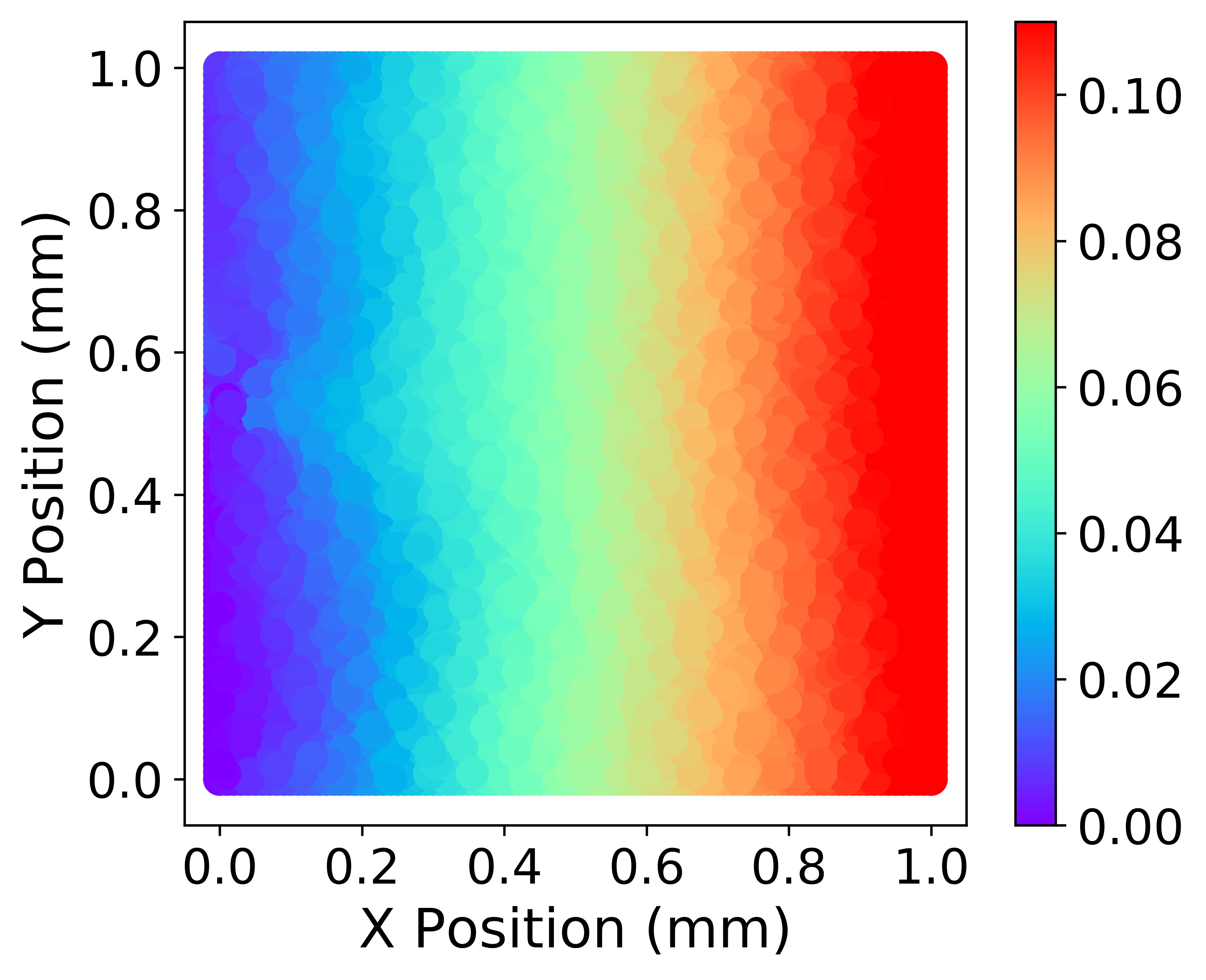}
    \end{minipage}
    \\ \hline
    PINN-FEM
    &
    \begin{minipage}{.19\textwidth}
      \includegraphics[width=0.9\linewidth]{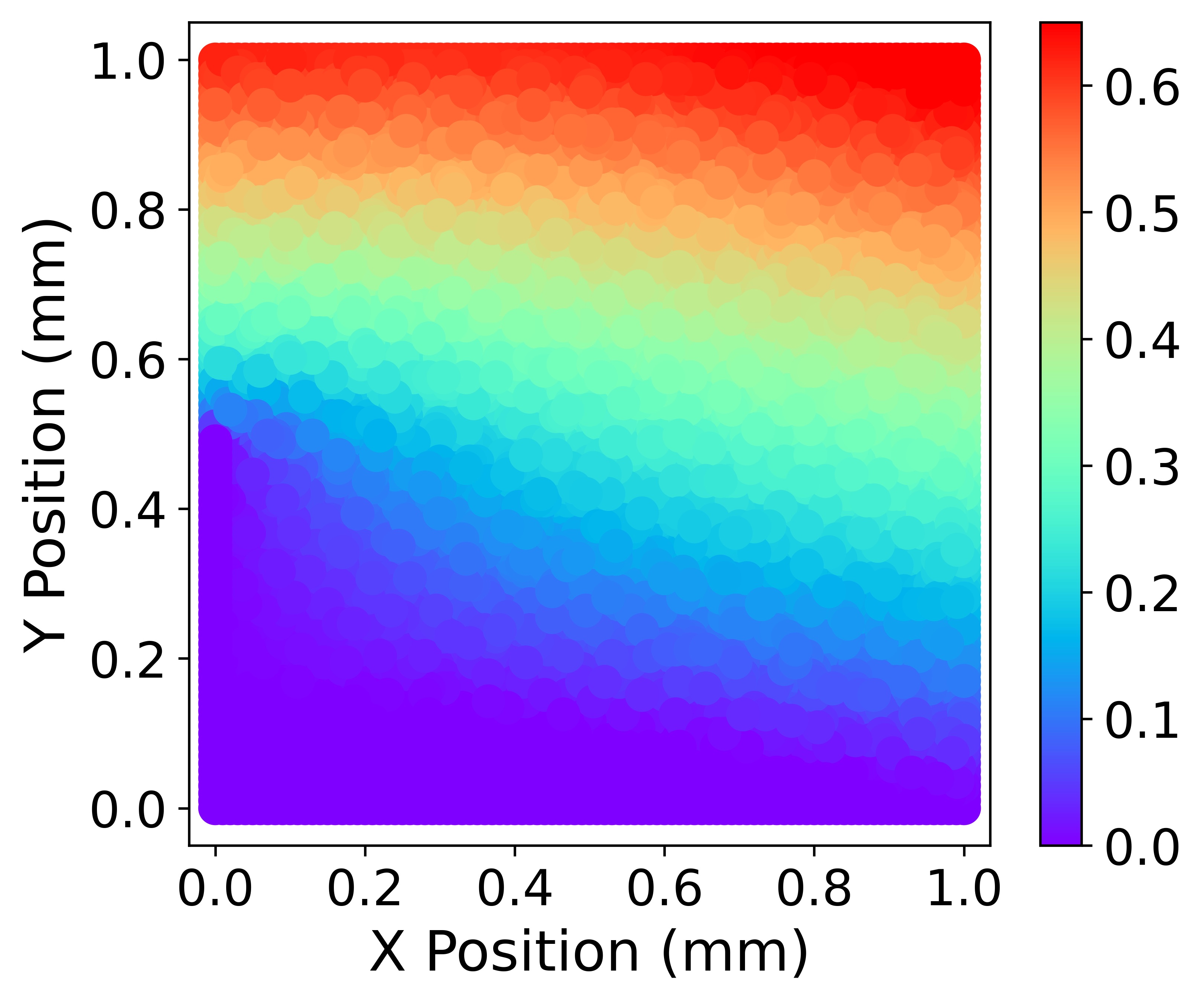}
    \end{minipage}
    &
   \begin{minipage}{.19\textwidth}
      \includegraphics[width=0.9\linewidth]{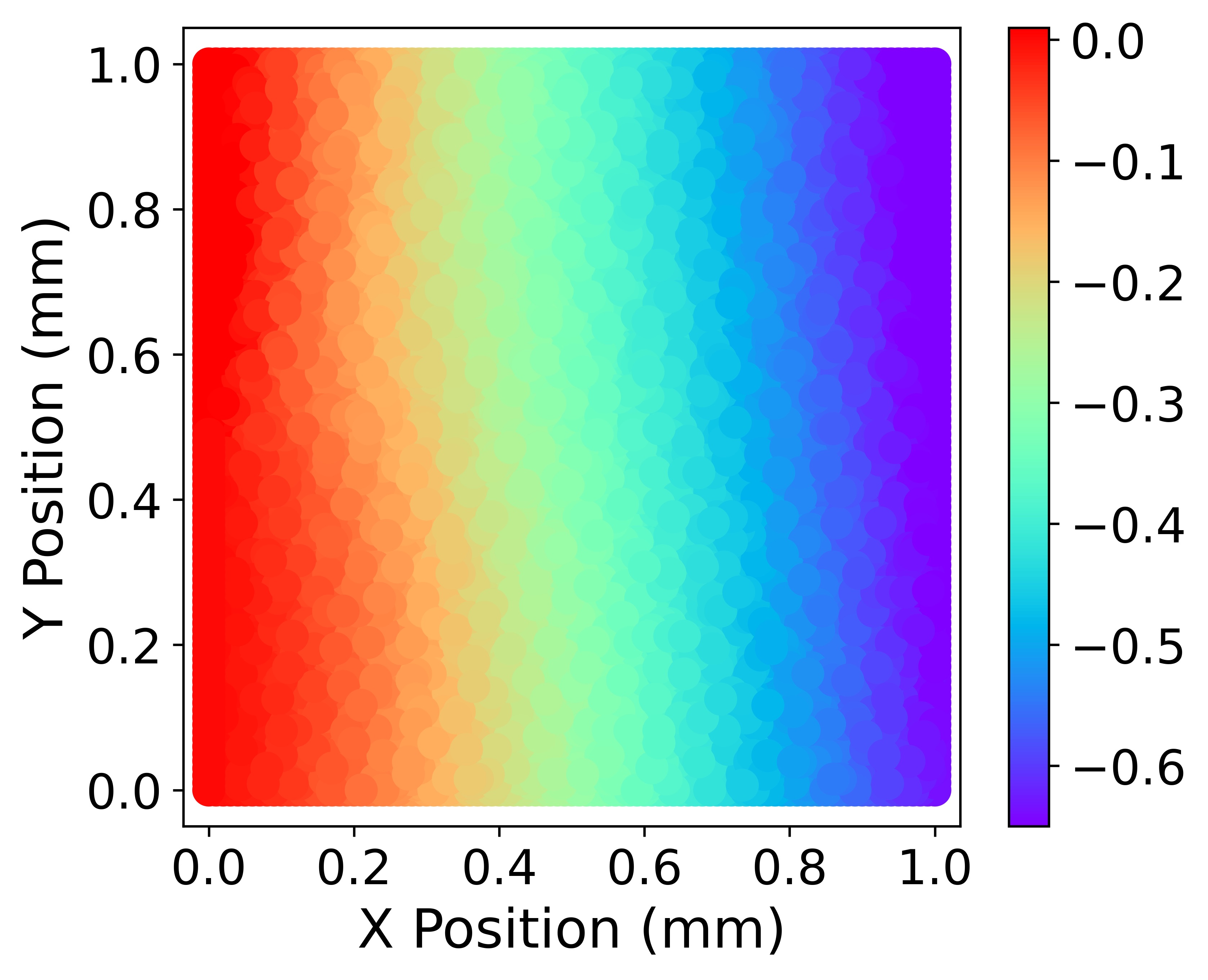}
    \end{minipage}
    & 
   \begin{minipage}{.19\textwidth}
      \includegraphics[width=0.9\linewidth]{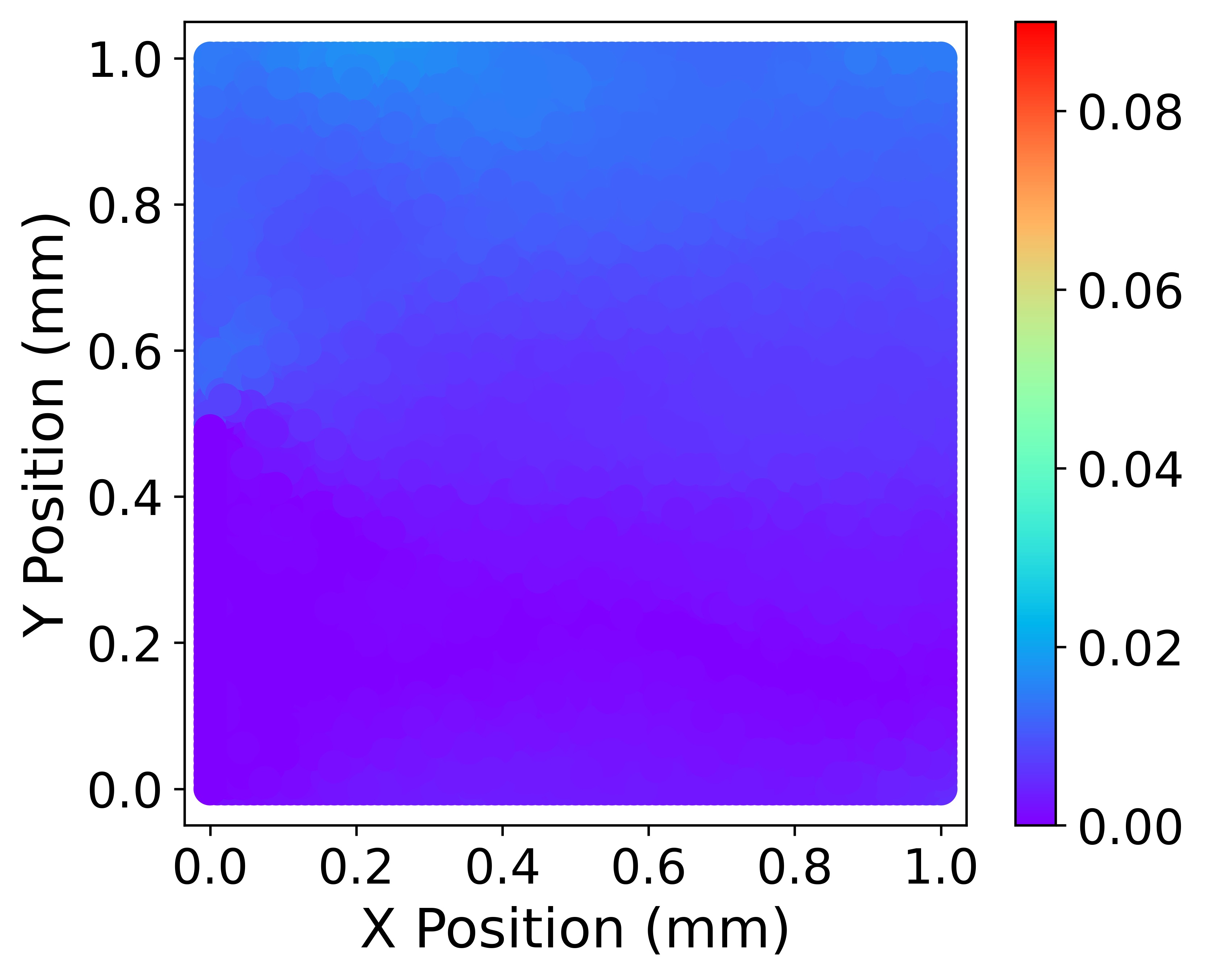}
    \end{minipage}
     & 
   \begin{minipage}{.19\textwidth}
      \includegraphics[width=0.9\linewidth]{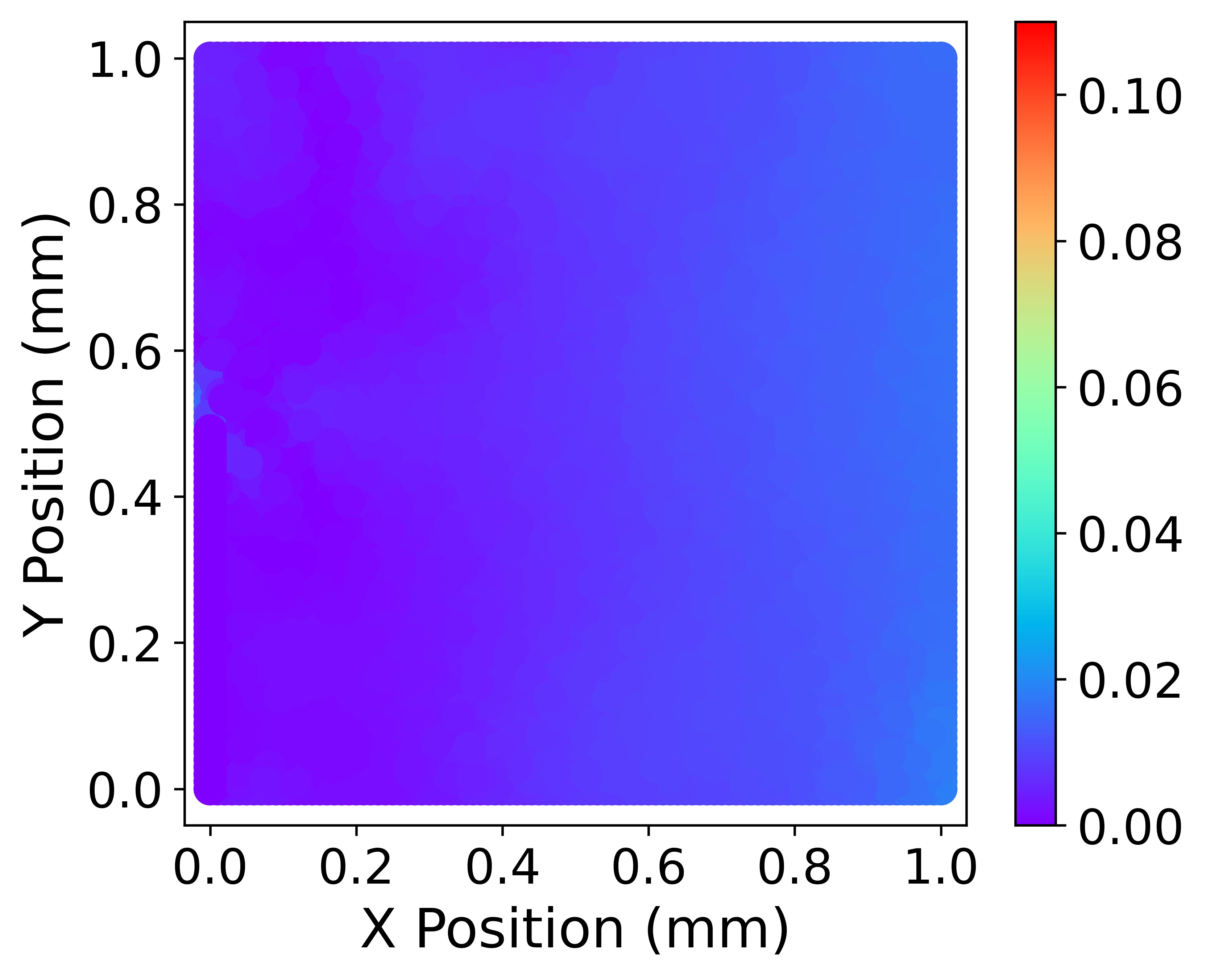}
    \end{minipage}
    \\ \hline
  \end{tabular}
  \label{fig:exp_halfbc_pred}
\end{table}

\subsubsection{Square plate with point boundaries}

For the fourth experiment, we consider the square elastic plate from the first experiment and apply a roller boundary condition on the left edge. However, fixed boundary conditions are applied at two points, $(0,0)$ and $(0,1)$. Similar to the previous experiments, a uniform stretching force of 10 units is applied on the right edge. The details of the domain and equations for boundary conditions are provided in Table \ref{table:exps_doms_bcs}. Approximate distance functions cannot be calculated for point boundary conditions. Therefore, apart from PINN with DF, we also cannot train the baseline model, PINN with ADF, for this experiment. Hence, the only baseline model considered here is the vanilla PINN network with soft imposition of boundary conditions. Vanilla PINN and PINN-FEM models are compared against the ground truth, which is the finite element solution generated using Abaqus for a mesh size of 0.1 mm. The nodal points from the same mesh are used as the collocation points for PINN-FEM. The relative errors, $e$, for both the models are shown in Table \ref{table:rel_error}. We observe that the proposed approach performs better by an order of magnitude than the baseline model. This is also evident in Table \ref{fig:exp_pointbc_pred}, which shows the distribution of the predicted response and the corresponding error with respect to the ground truth for all the models. The vanilla PINN fails to predict the point boundary conditions, which are successfully captured by our model.

\begin{table}[!ht]
  \centering
    \caption{The distribution of predicted displacement fields, $\hat{u}_x$ and $\hat{u}_y$, and the corresponding error for the square elastic plate with point boundary conditions as given in Eq.~\eqref{bc_example4}.}
  \begin{tabular}{ |c | c | c |c | c| }
    \hline
    PINN Model & $\hat{u}_x$ & $\hat{u}_y$ & $|\hat{u}_x-u_x|$ & $|\hat{u}_y-u_y|$ \\ \hline
    Ground Truth
    &
    \begin{minipage}{.19\textwidth}
      \includegraphics[width=0.9\linewidth]{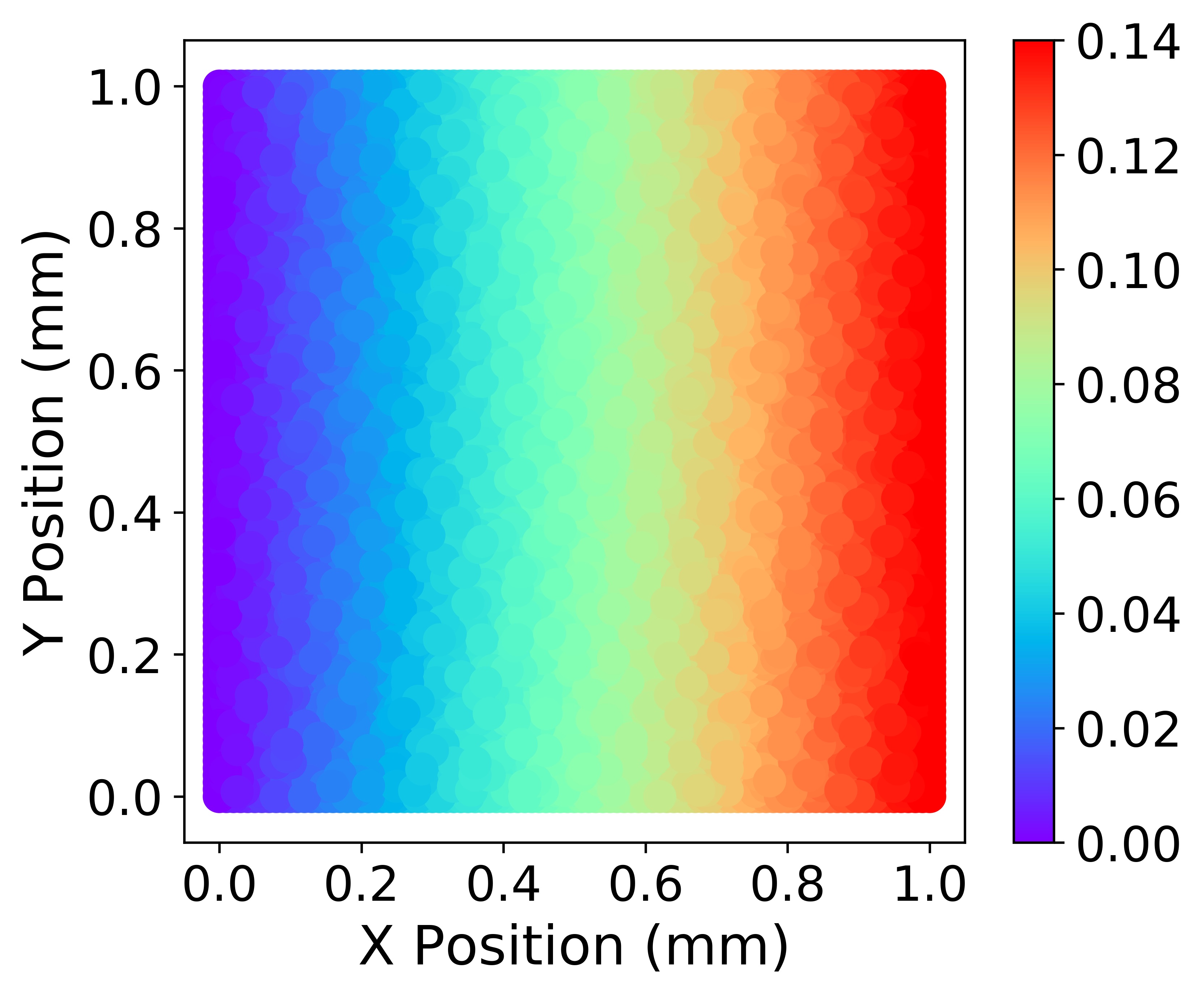}
    \end{minipage}
    &
   \begin{minipage}{.19\textwidth}
      \includegraphics[width=0.9\linewidth]{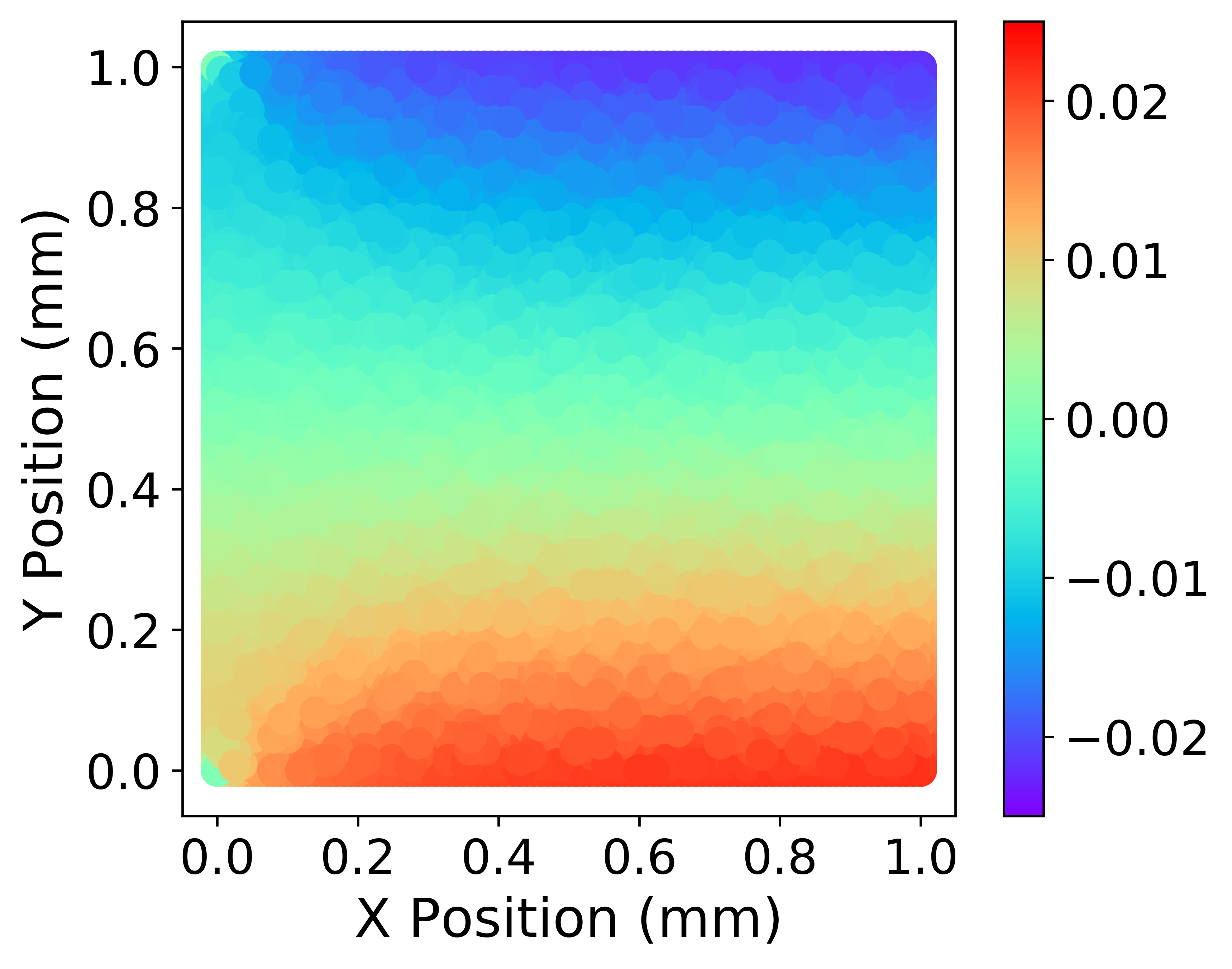}
    \end{minipage}
    & 
     & 
    \\ \hline
    Soft
    &
    \begin{minipage}{.19\textwidth}
      \includegraphics[width=0.9\linewidth]{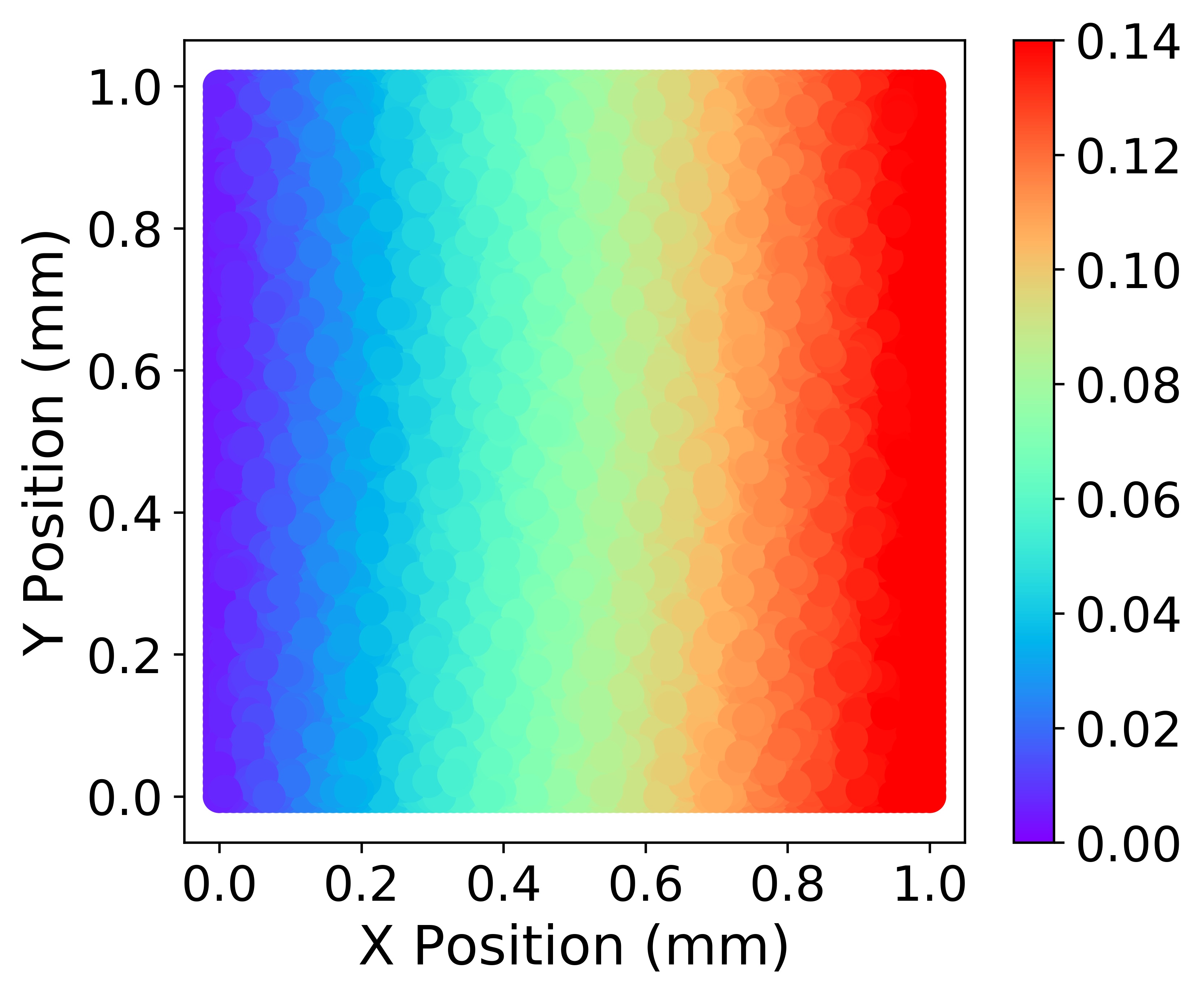}
    \end{minipage}
    &
   \begin{minipage}{.19\textwidth}
      \includegraphics[width=0.9\linewidth]{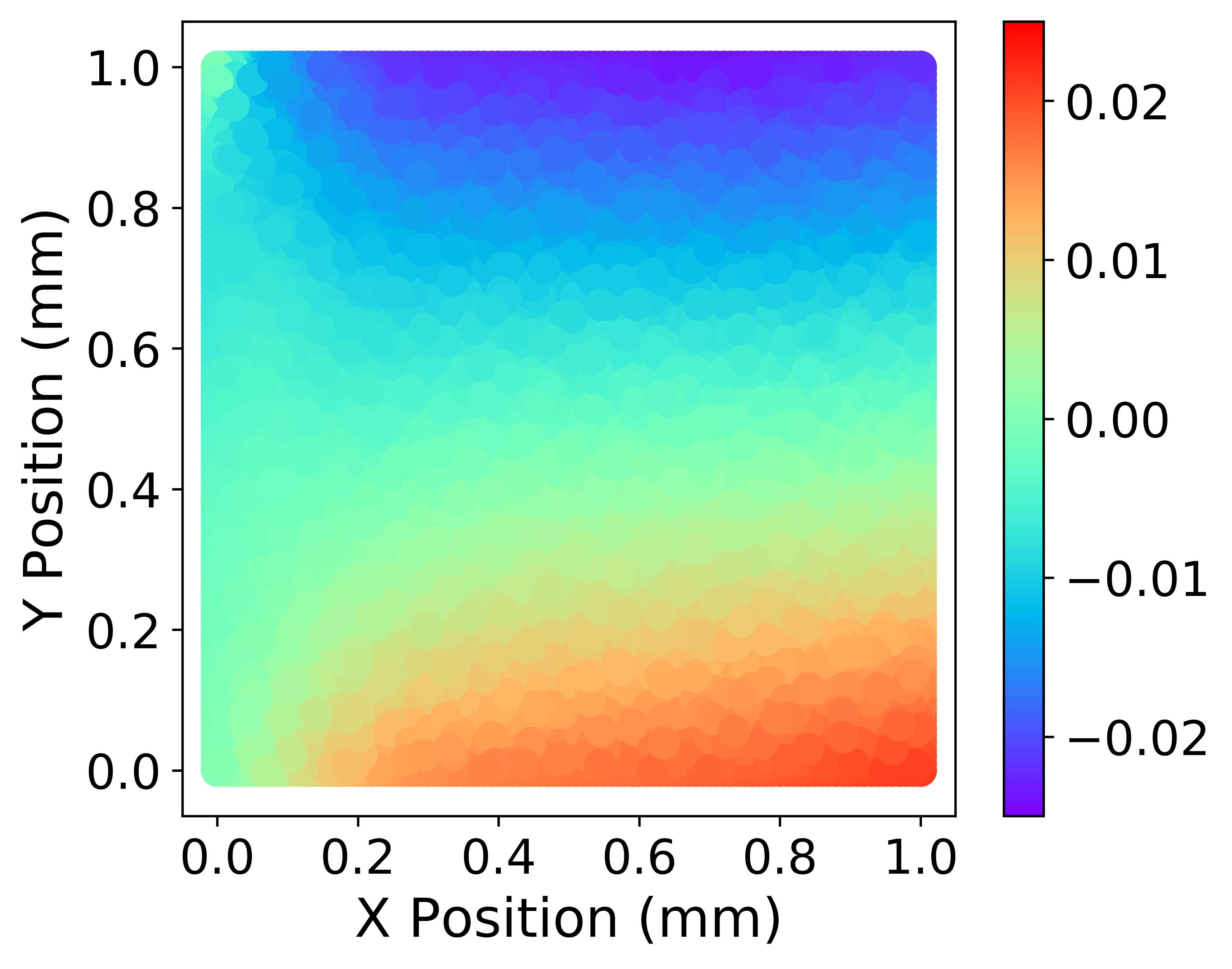}
    \end{minipage}
    & 
   \begin{minipage}{.19\textwidth}
      \includegraphics[width=0.9\linewidth]{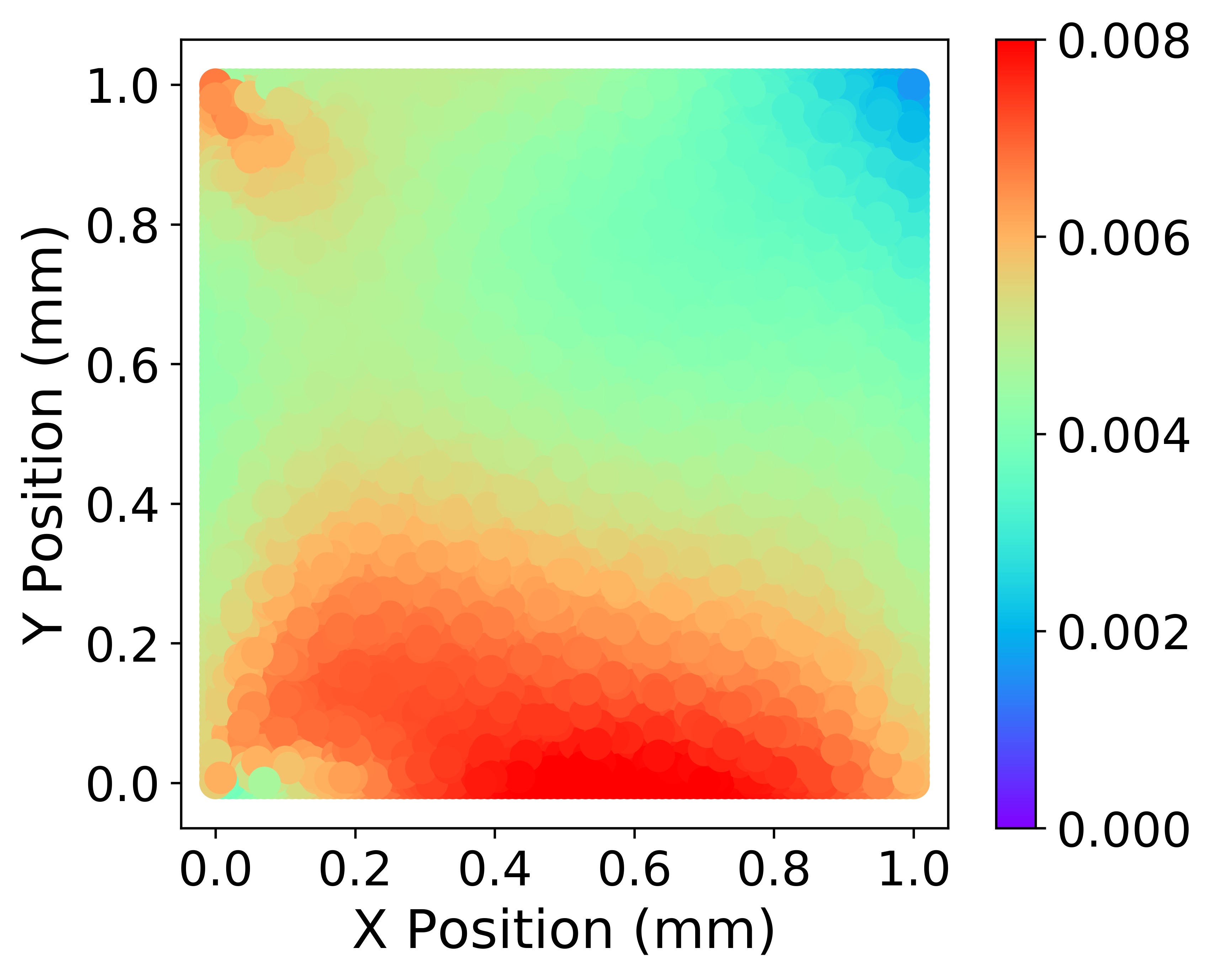}
    \end{minipage}
     & 
   \begin{minipage}{.19\textwidth}
      \includegraphics[width=0.9\linewidth]{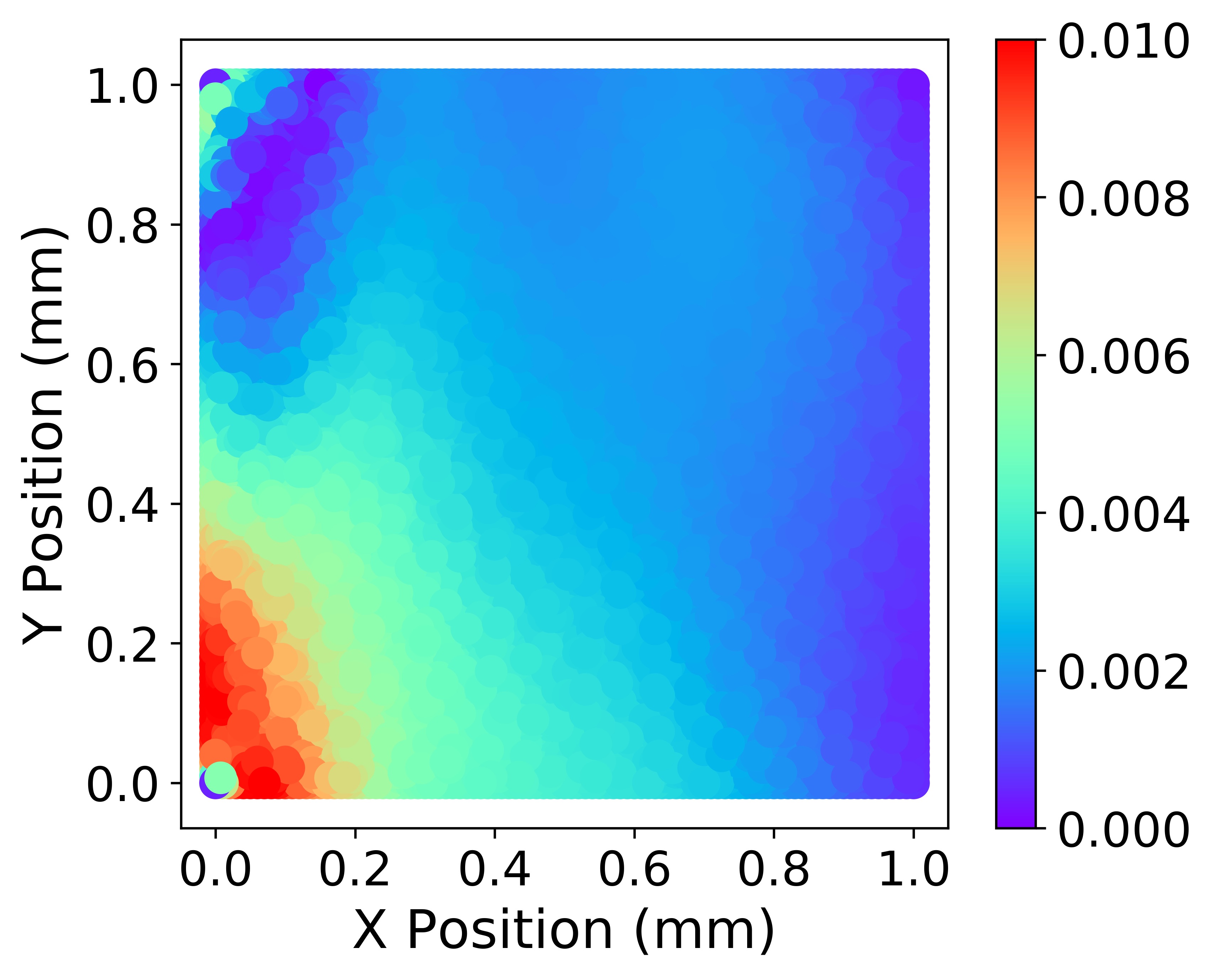}
    \end{minipage}
    \\ \hline
    PINN-FEM
    &
    \begin{minipage}{.19\textwidth}
      \includegraphics[width=0.9\linewidth]{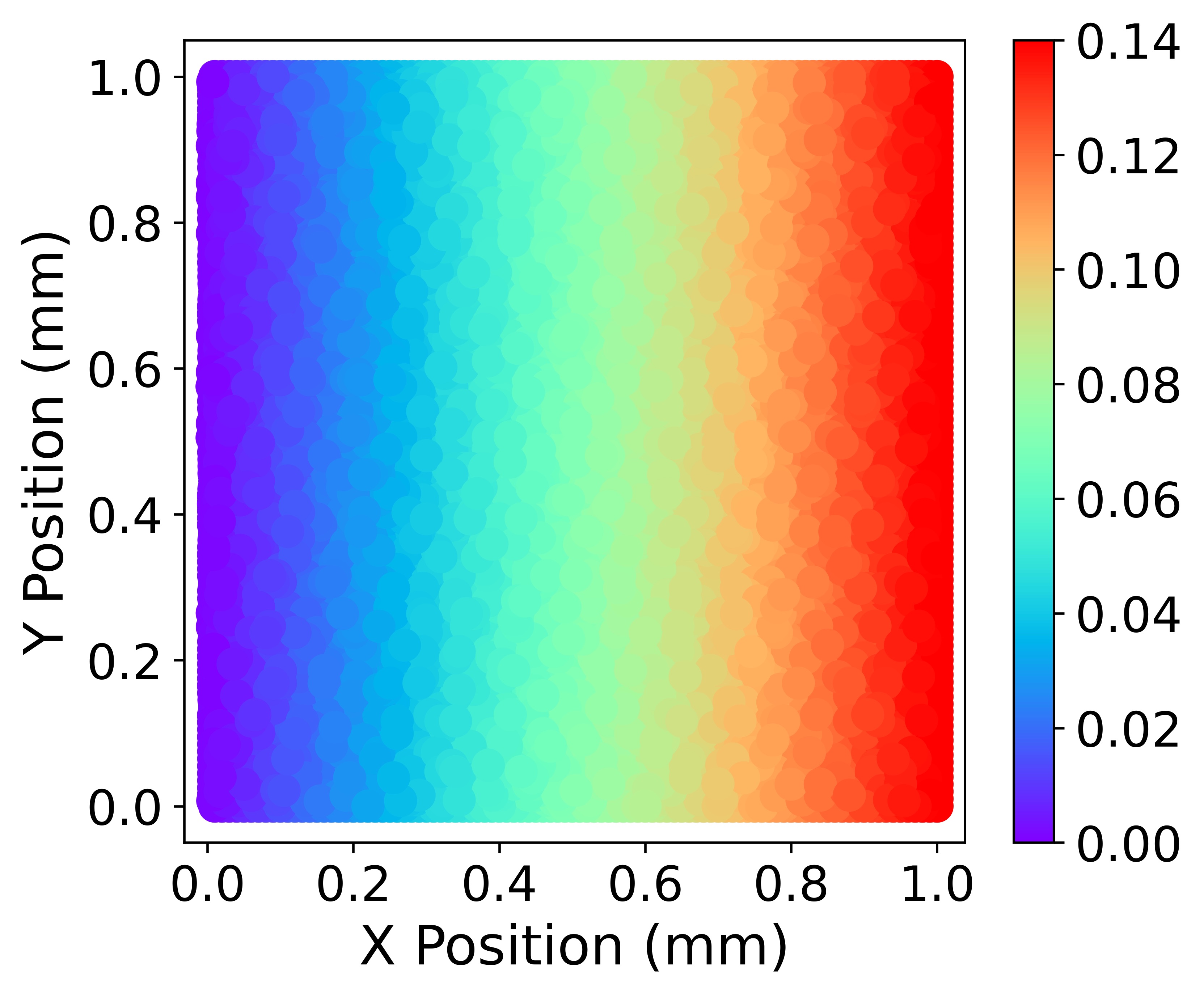}
    \end{minipage}
    &
   \begin{minipage}{.19\textwidth}
      \includegraphics[width=0.9\linewidth]{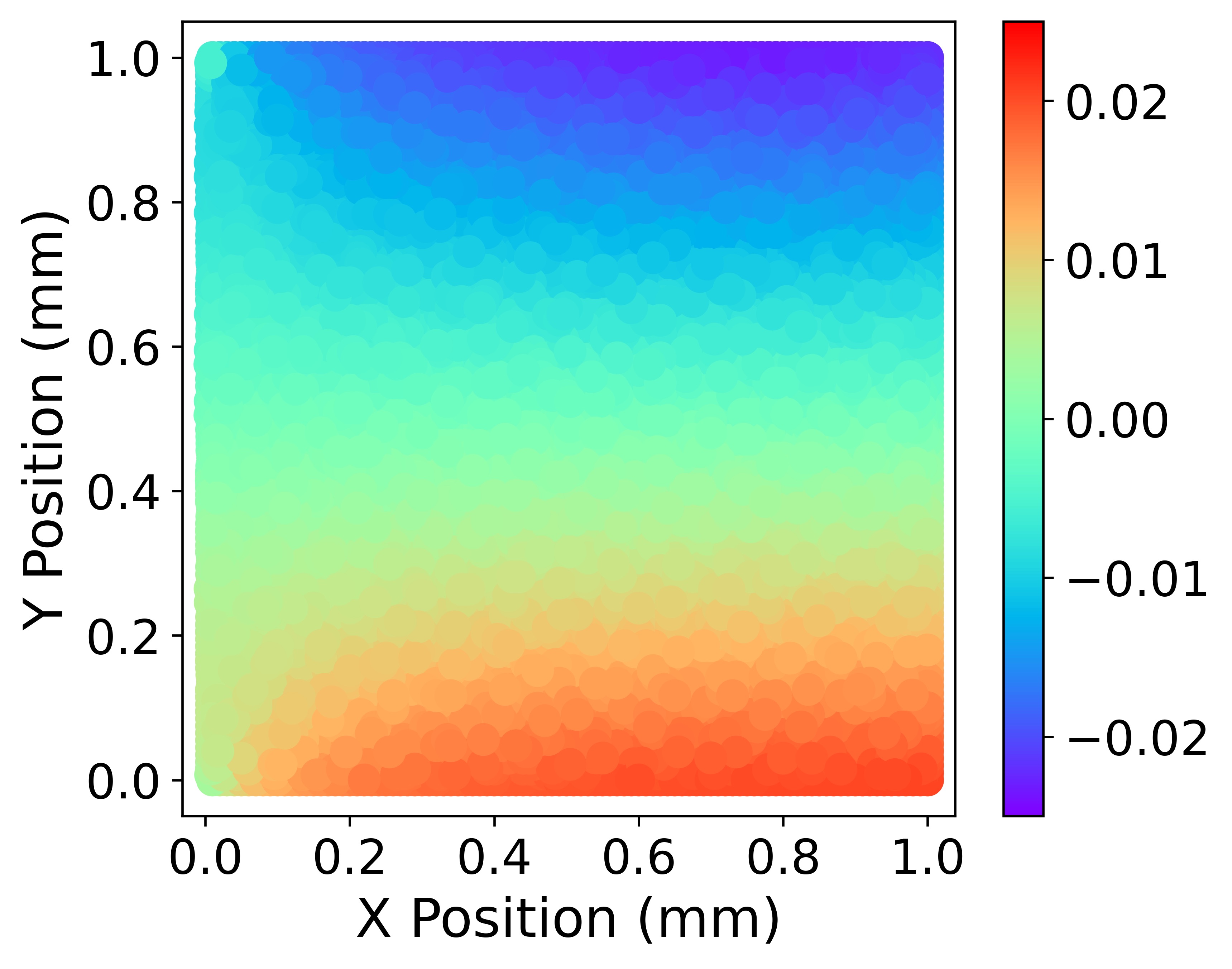}
    \end{minipage}
    & 
   \begin{minipage}{.19\textwidth}
      \includegraphics[width=0.9\linewidth]{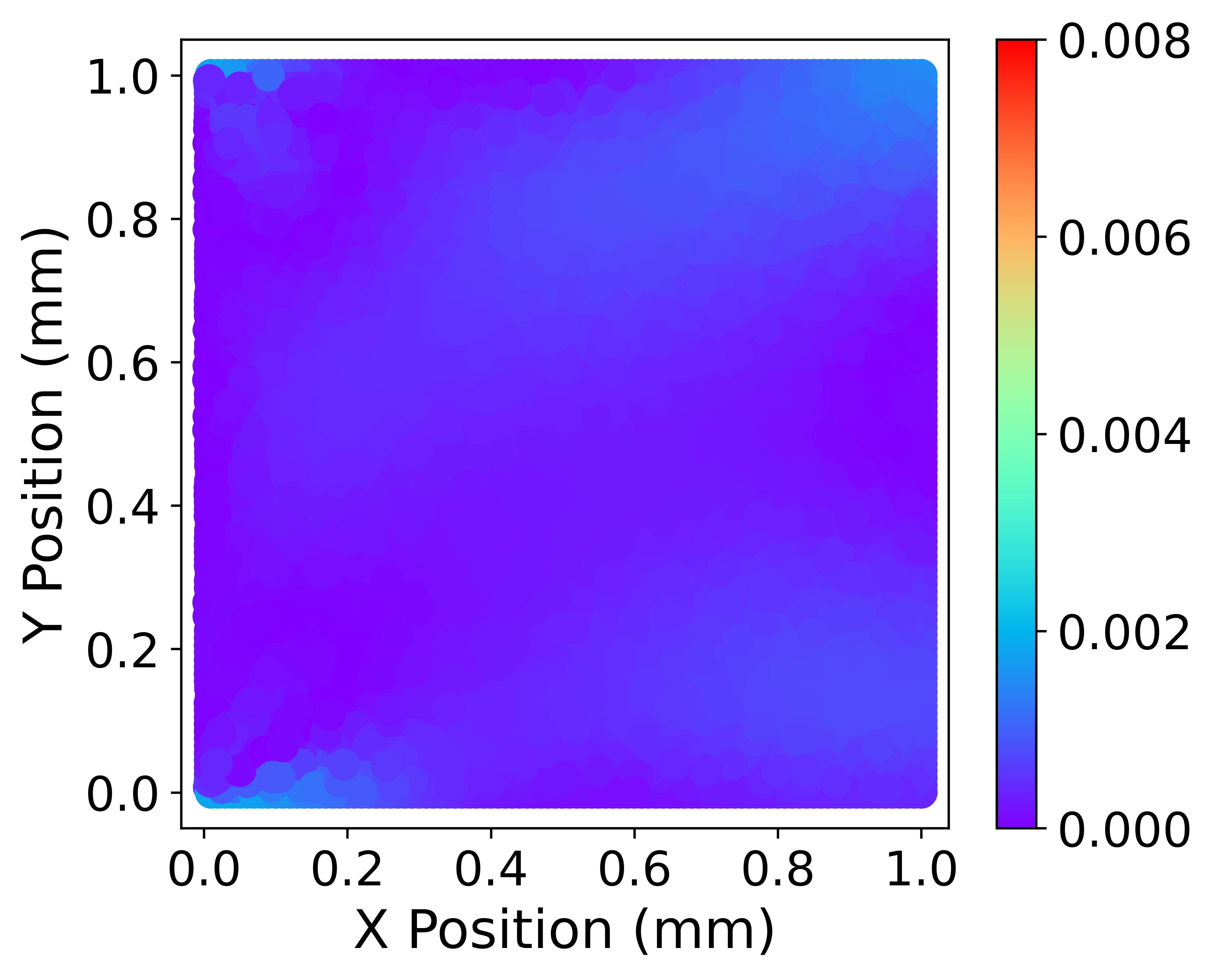}
    \end{minipage}
     & 
   \begin{minipage}{.19\textwidth}
      \includegraphics[width=0.9\linewidth]{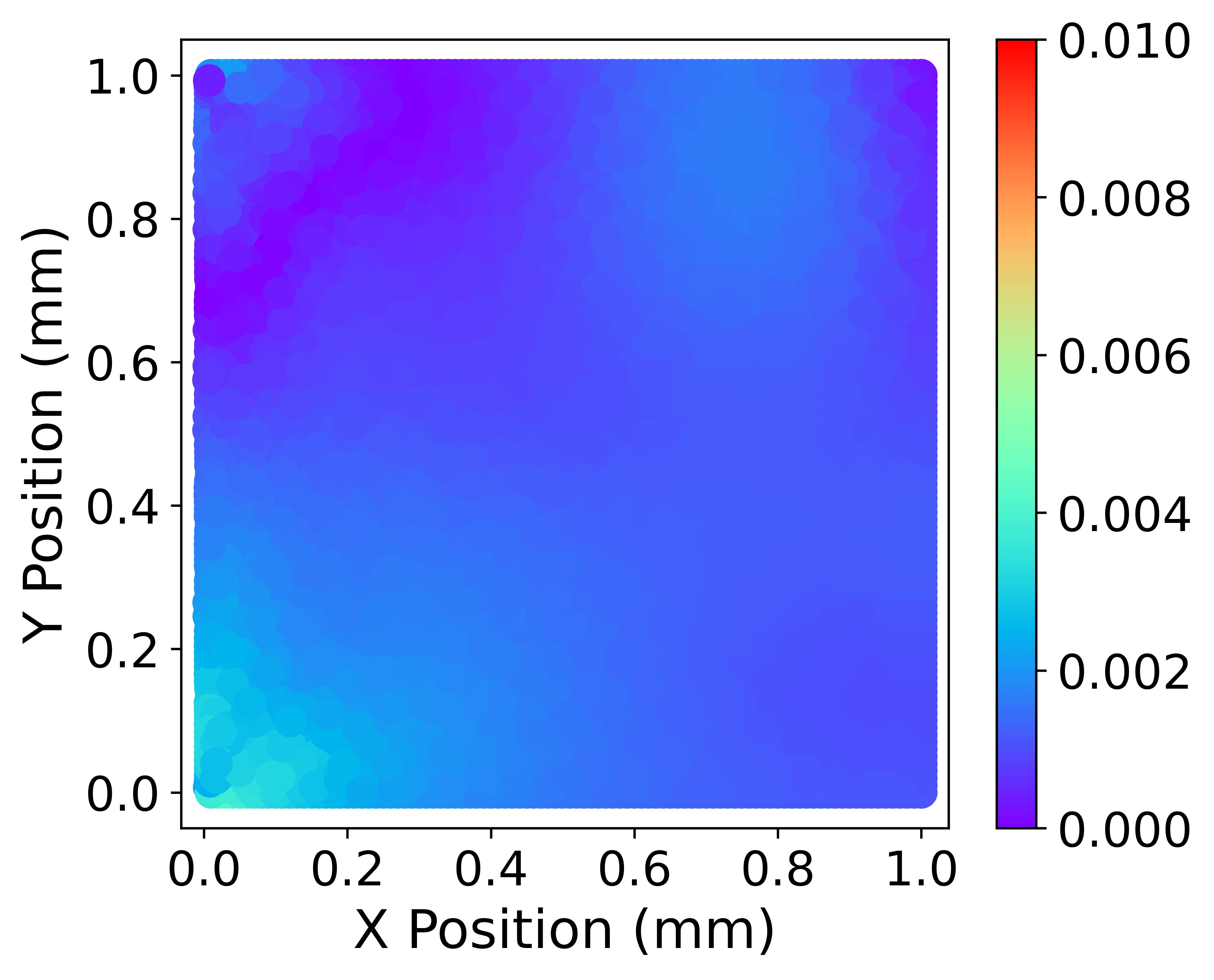}
    \end{minipage}
    \\ \hline
  \end{tabular}
  \label{fig:exp_pointbc_pred}
\end{table}

\subsubsection{Square plate with a crack}

For the fifth experiment, we consider a square elastic plate with a crack at the top edge as shown in Table \ref{table:exps_doms_bcs}. Fixed boundary conditions are applied on the left edge and a uniform stretching force of 10 units is applied on the right edge. The equations of the prescribed boundary conditions are given in Eq.~\eqref{bc_example5}. The ground truth and the collocation points are generated in the same way as the previous experiments, using a mesh with mesh size of 0.1 mm. Table \ref{fig:crack_pred} shows the distribution of displacement fields and the corresponding error compared to the ground truth. We observe that all the benchmark models exhibit higher errors in displacement prediction for the right half of the square, where the displacement values are higher due to the presence of the crack. While the vanilla PINN with soft imposition is the worst performs the worst. While PINN with ADF and DF demonstrate comparatively better performance, the errors are higher near the left and top boundaries. The best performing model is the proposed PINN-FEM approach, with low errors throughout the domain. This is also evident from the relative errors, $e$, as given in Table \ref{table:rel_error}.

\begin{table}[!ht]
  \centering
    \caption{The distribution of predicted displacement fields, $\hat{u}_x$ and $\hat{u}_y$, and the corresponding error for the square elastic plate with a crack, under plane stress conditions, with boundary conditions as given in Eq.~\eqref{bc_example5}.}
  \begin{tabular}{ |c | c | c |c | c| }
    \hline
    PINN Model & $\hat{u}_x$ & $\hat{u}_y$ & $|\hat{u}_x-u_x|$ & $|\hat{u}_y-u_y|$ \\ \hline
    Ground Truth
    &
    \begin{minipage}{.19\textwidth}
      \includegraphics[width=0.9\linewidth]{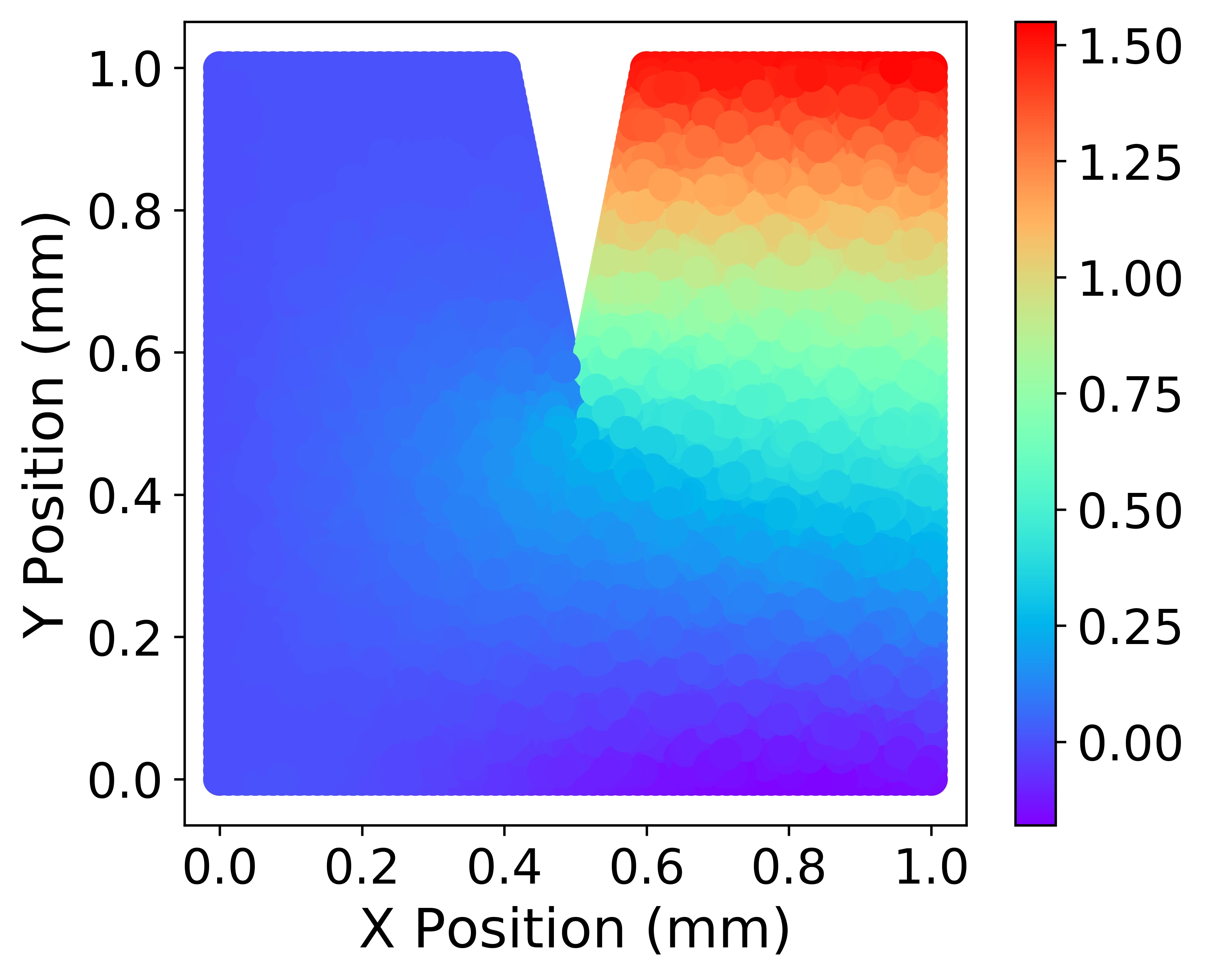}
    \end{minipage}
    &
   \begin{minipage}{.19\textwidth}
      \includegraphics[width=0.9\linewidth]{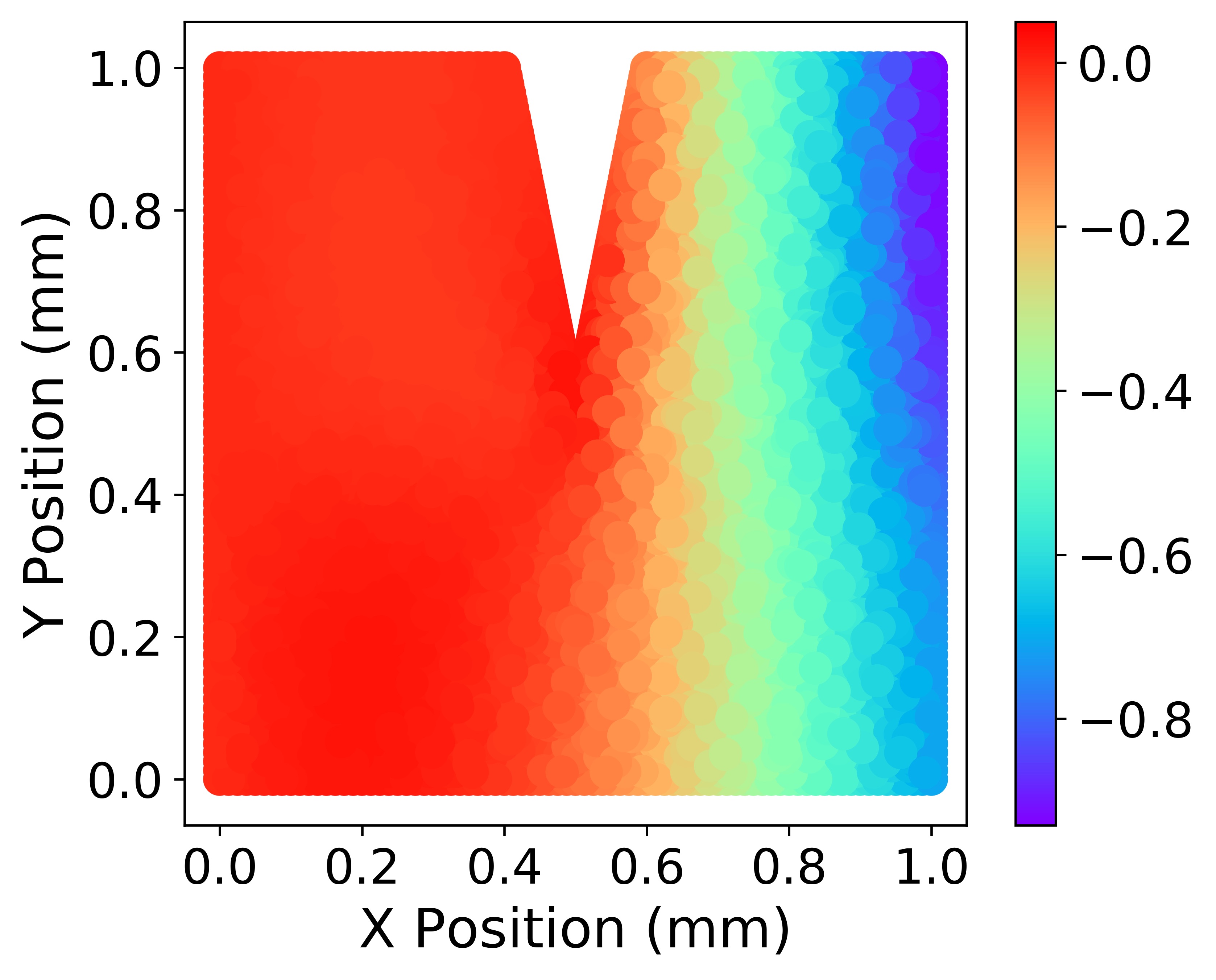}
    \end{minipage}
    & 
     & 
    \\ \hline
    Soft
    &
    \begin{minipage}{.19\textwidth}
      \includegraphics[width=0.9\linewidth]{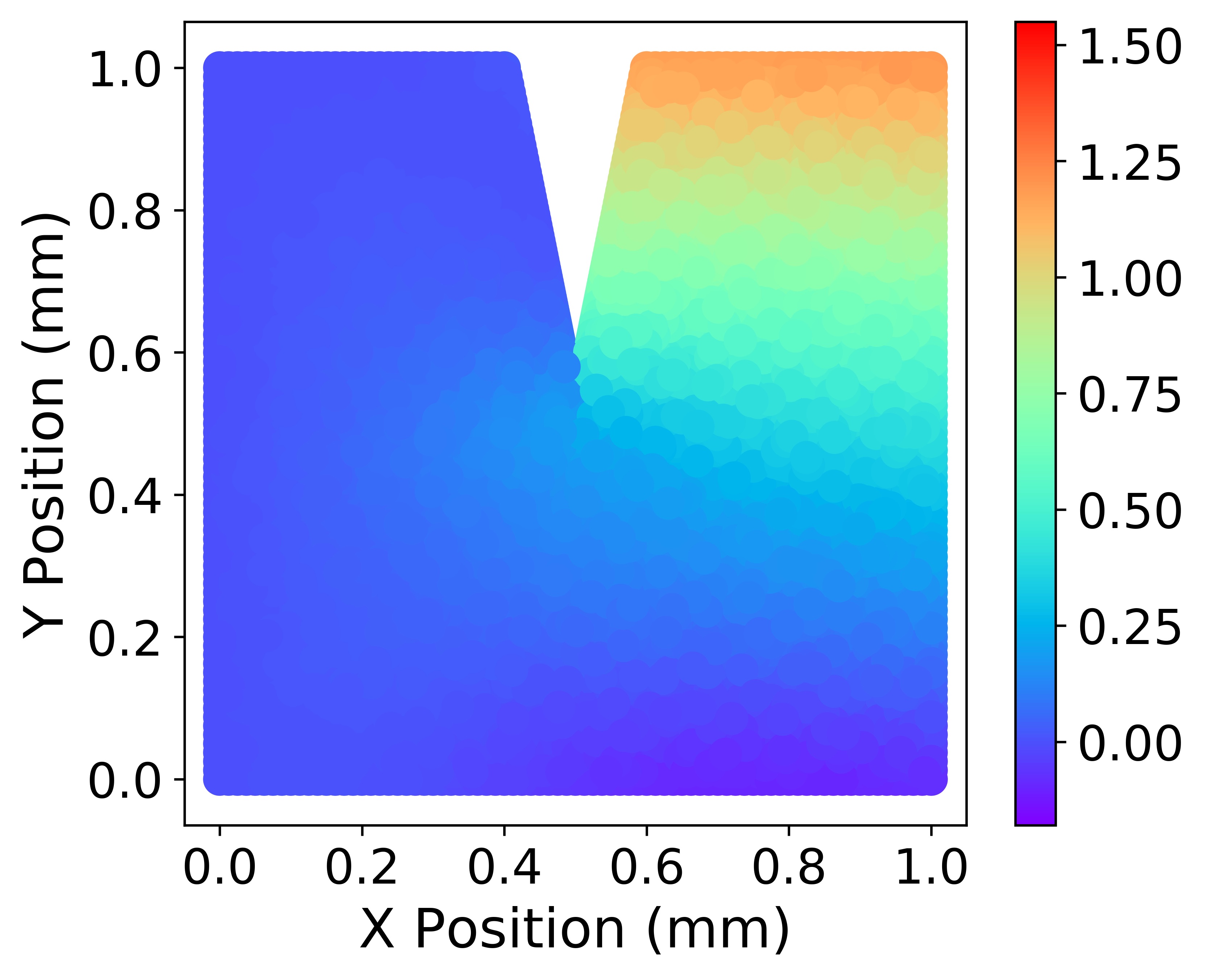}
    \end{minipage}
    &
   \begin{minipage}{.19\textwidth}
      \includegraphics[width=0.9\linewidth]{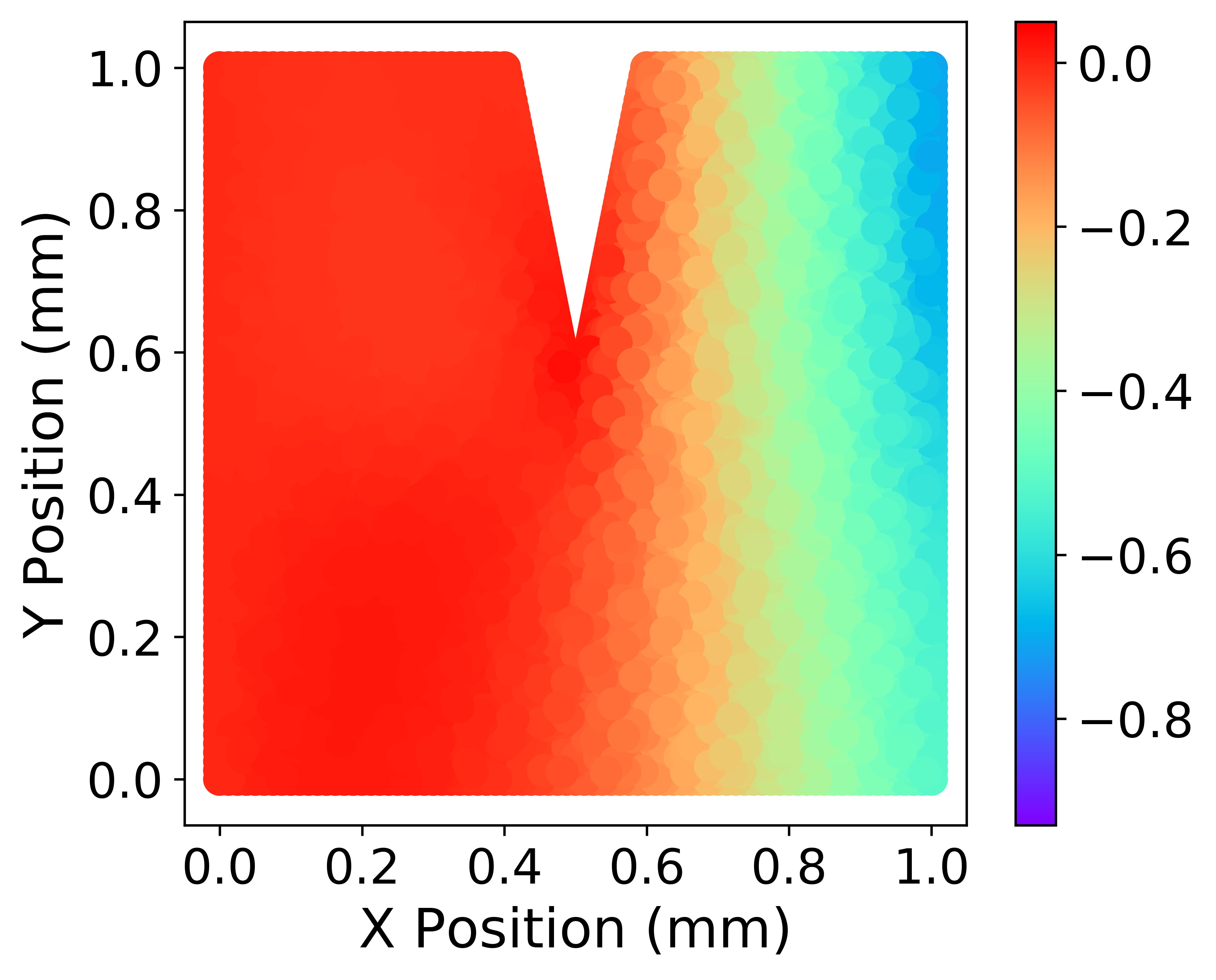}
    \end{minipage}
    & 
   \begin{minipage}{.19\textwidth}
      \includegraphics[width=0.9\linewidth]{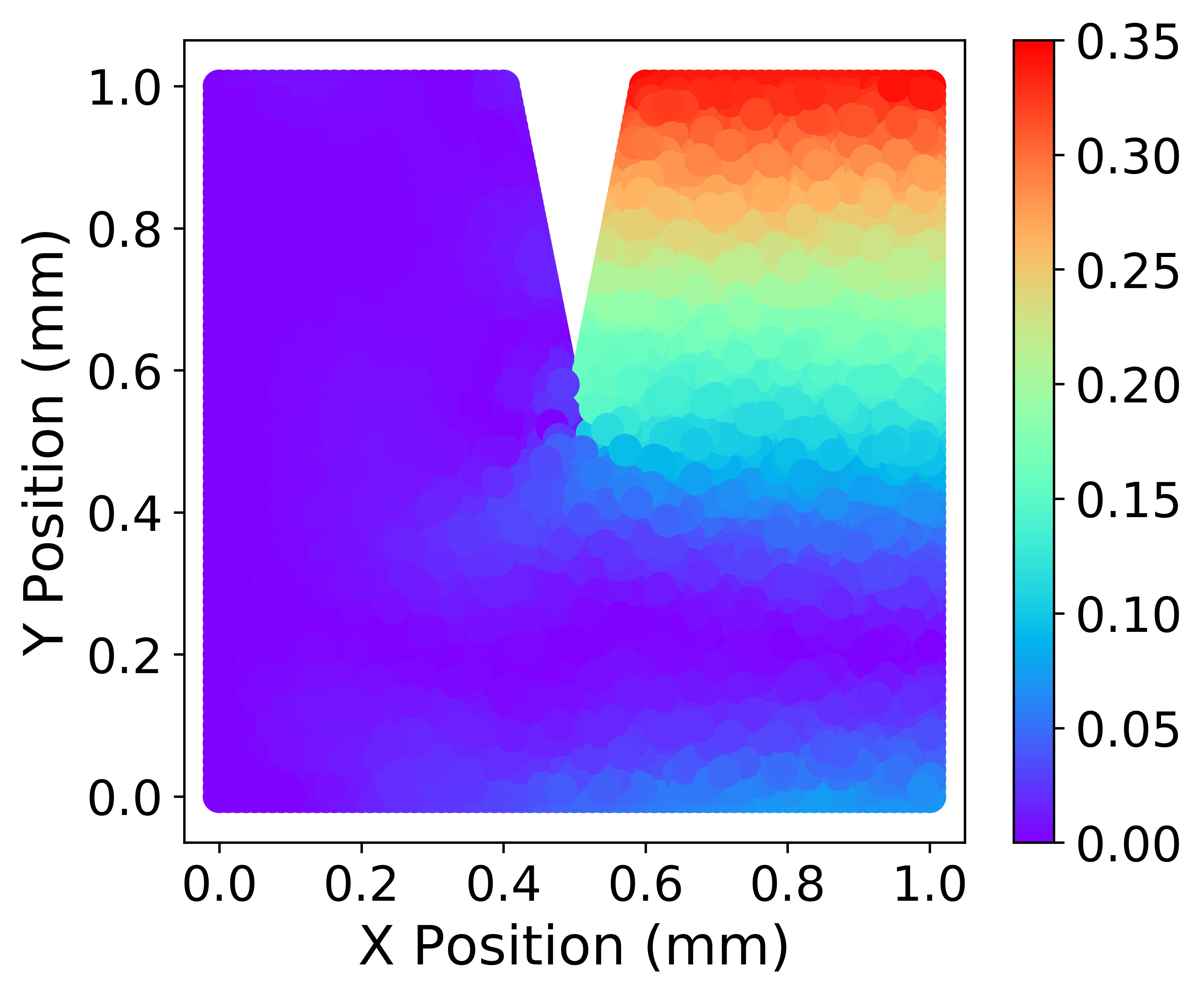}
    \end{minipage}
     & 
   \begin{minipage}{.19\textwidth}
      \includegraphics[width=0.9\linewidth]{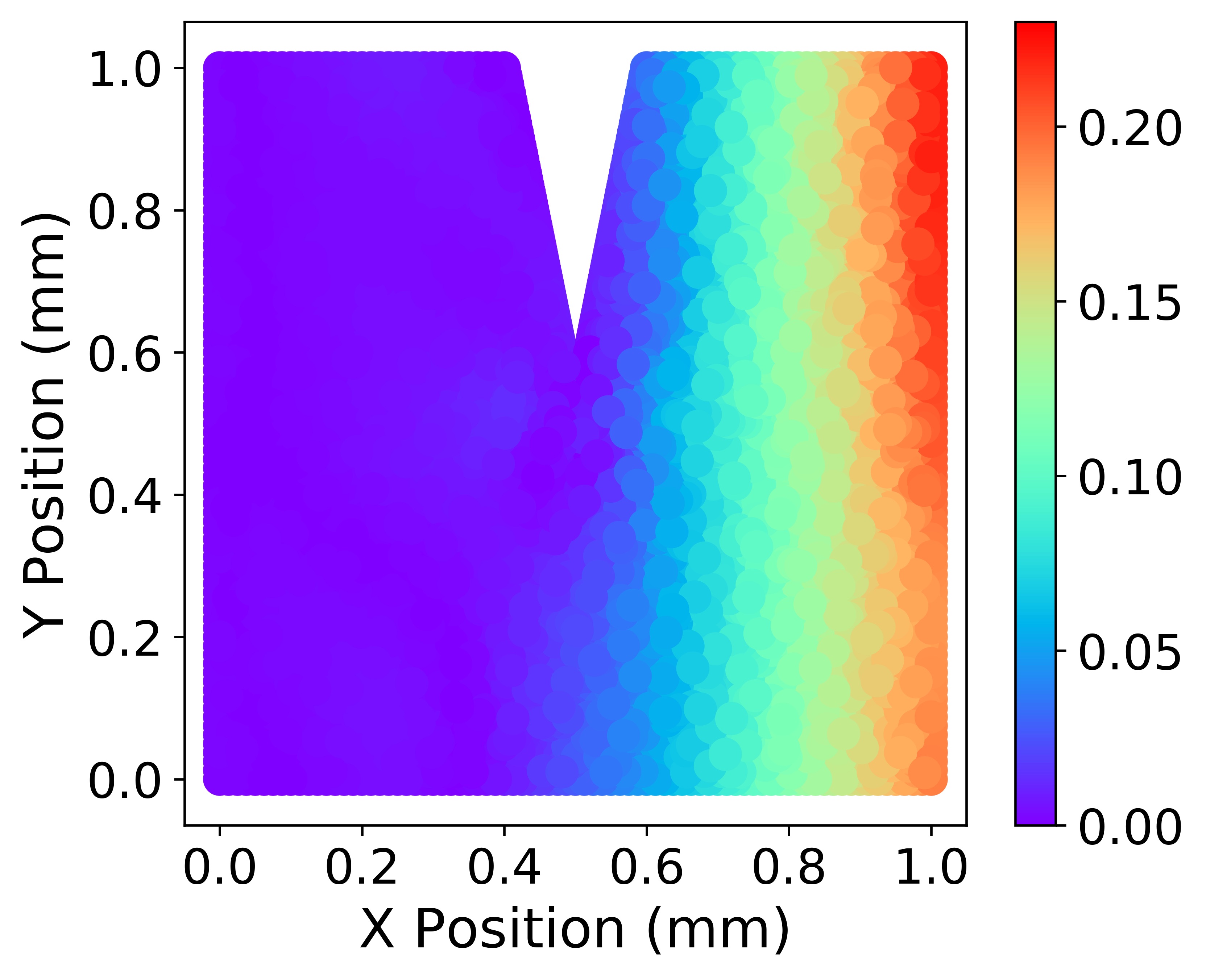}
    \end{minipage}
    \\ \hline
    ADF
    &
    \begin{minipage}{.19\textwidth}
      \includegraphics[width=0.9\linewidth]{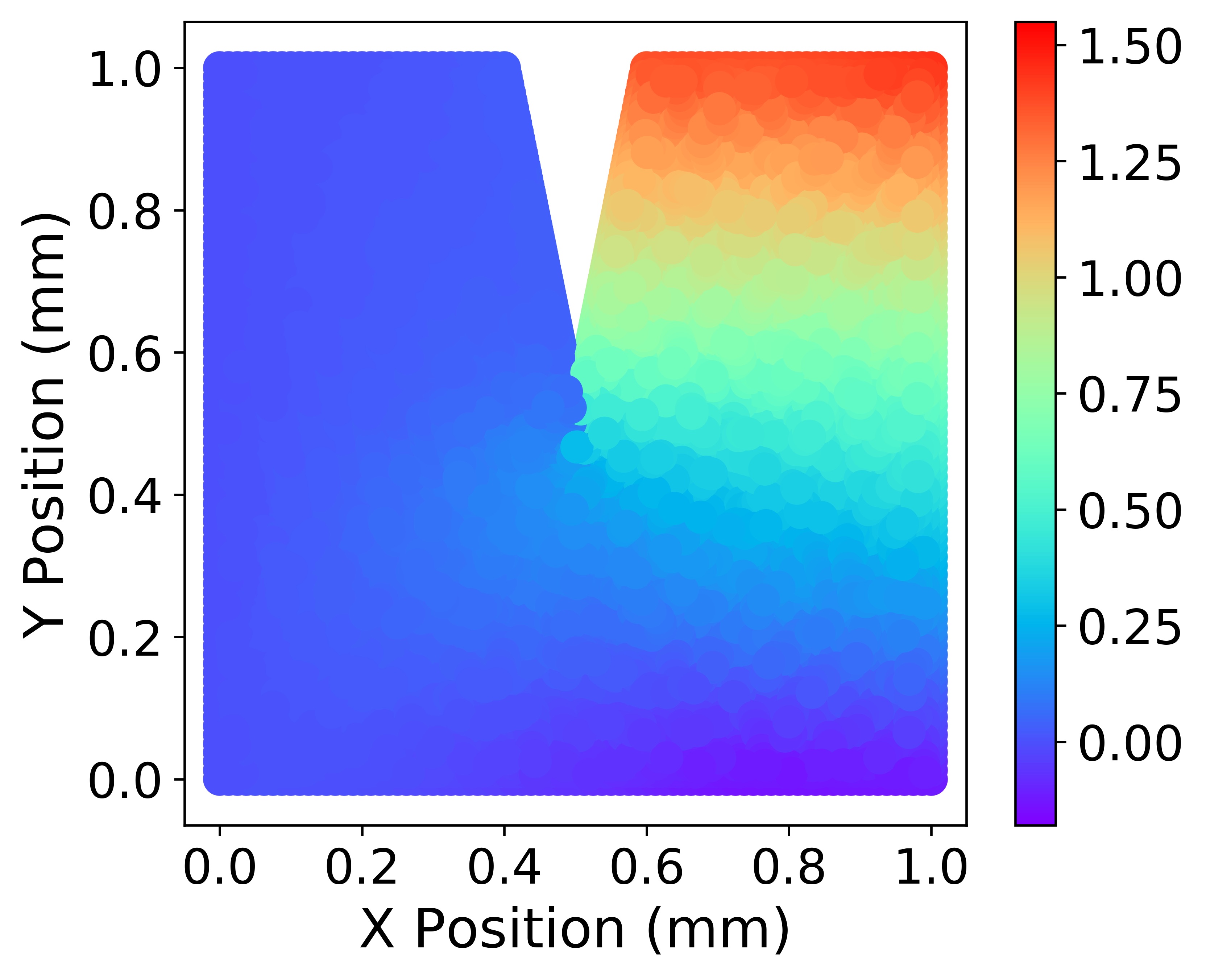}
    \end{minipage}
    &
   \begin{minipage}{.19\textwidth}
      \includegraphics[width=0.9\linewidth]{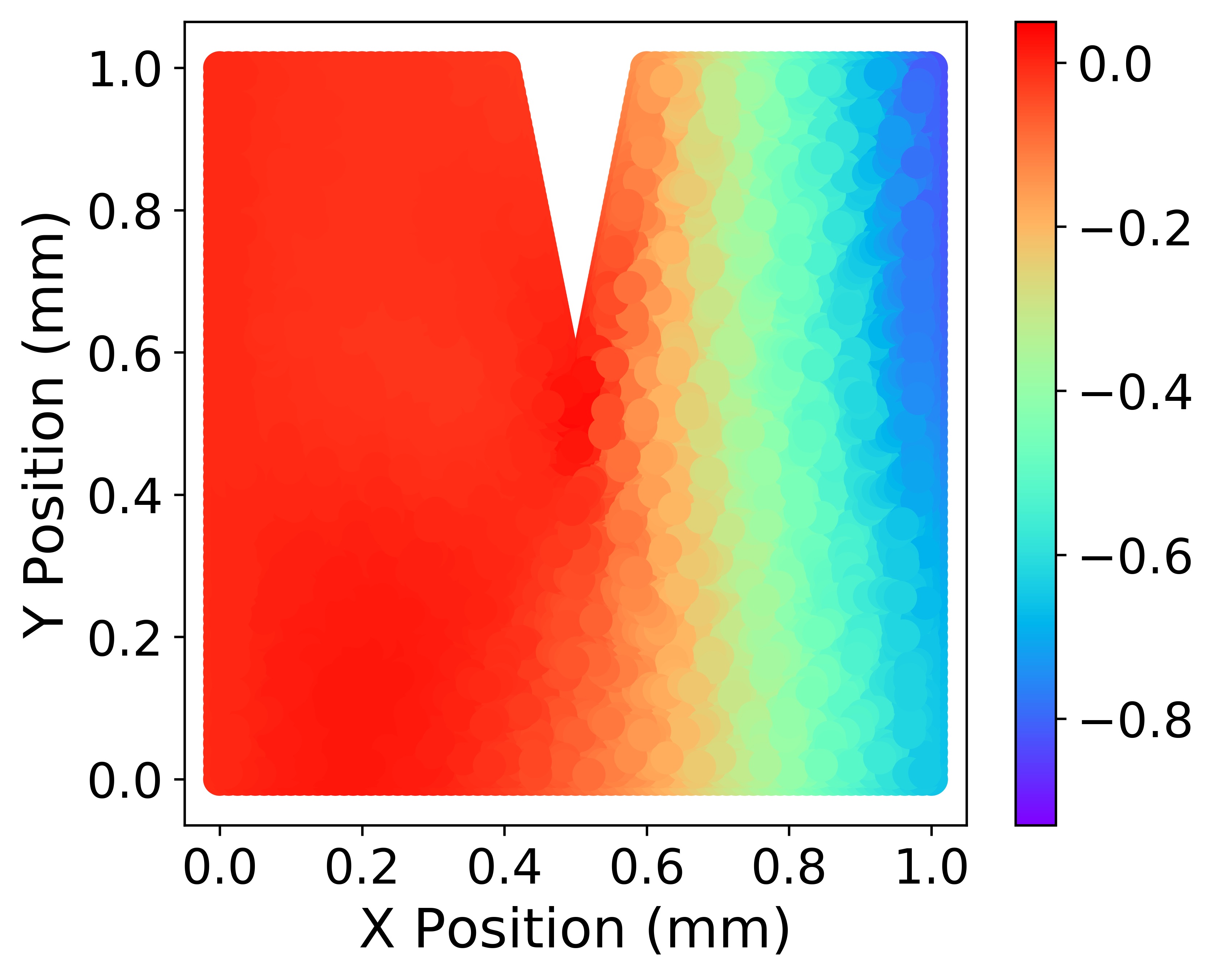}
    \end{minipage}
    & 
   \begin{minipage}{.19\textwidth}
      \includegraphics[width=0.9\linewidth]{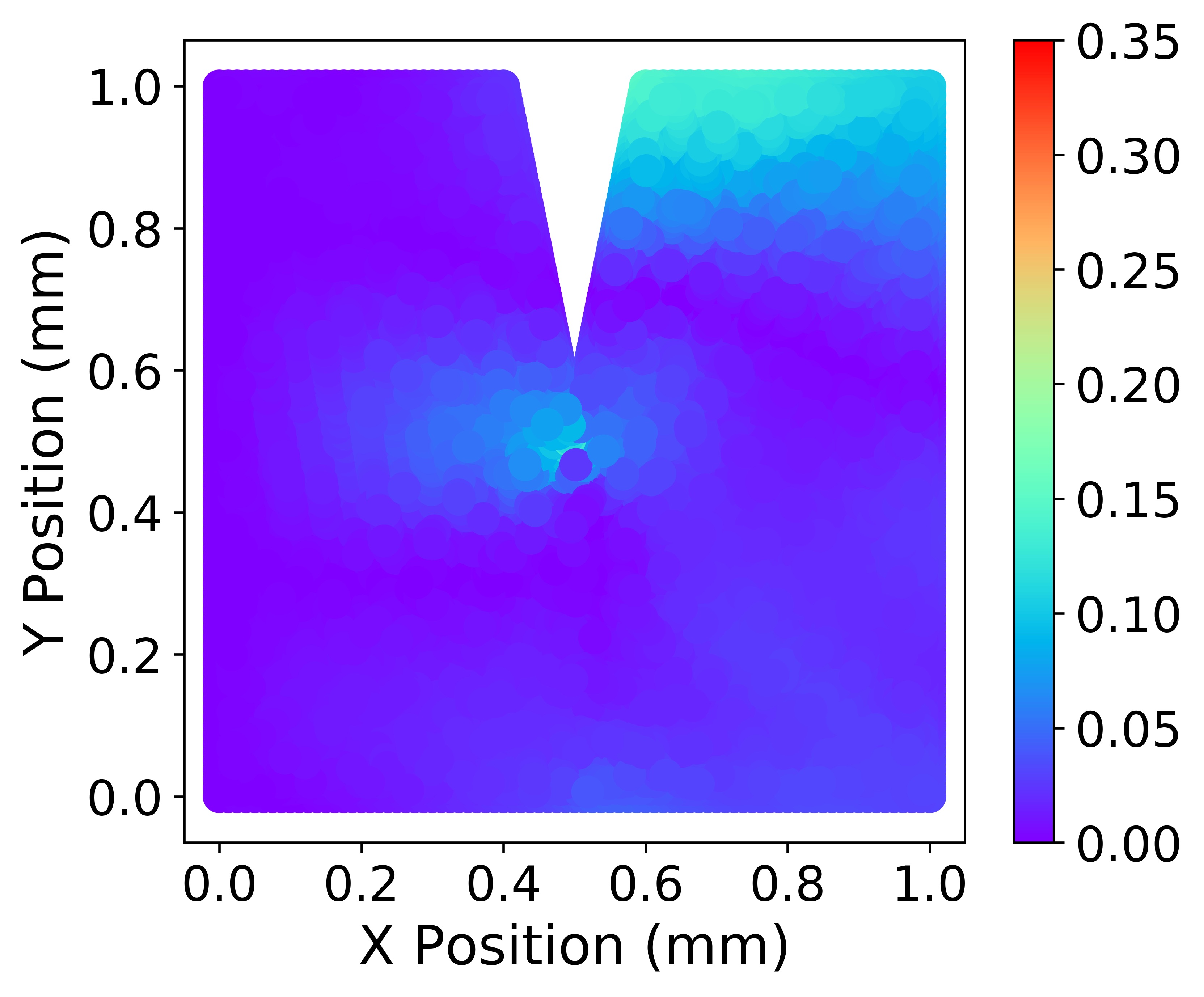}
    \end{minipage}
     & 
   \begin{minipage}{.19\textwidth}
      \includegraphics[width=0.9\linewidth]{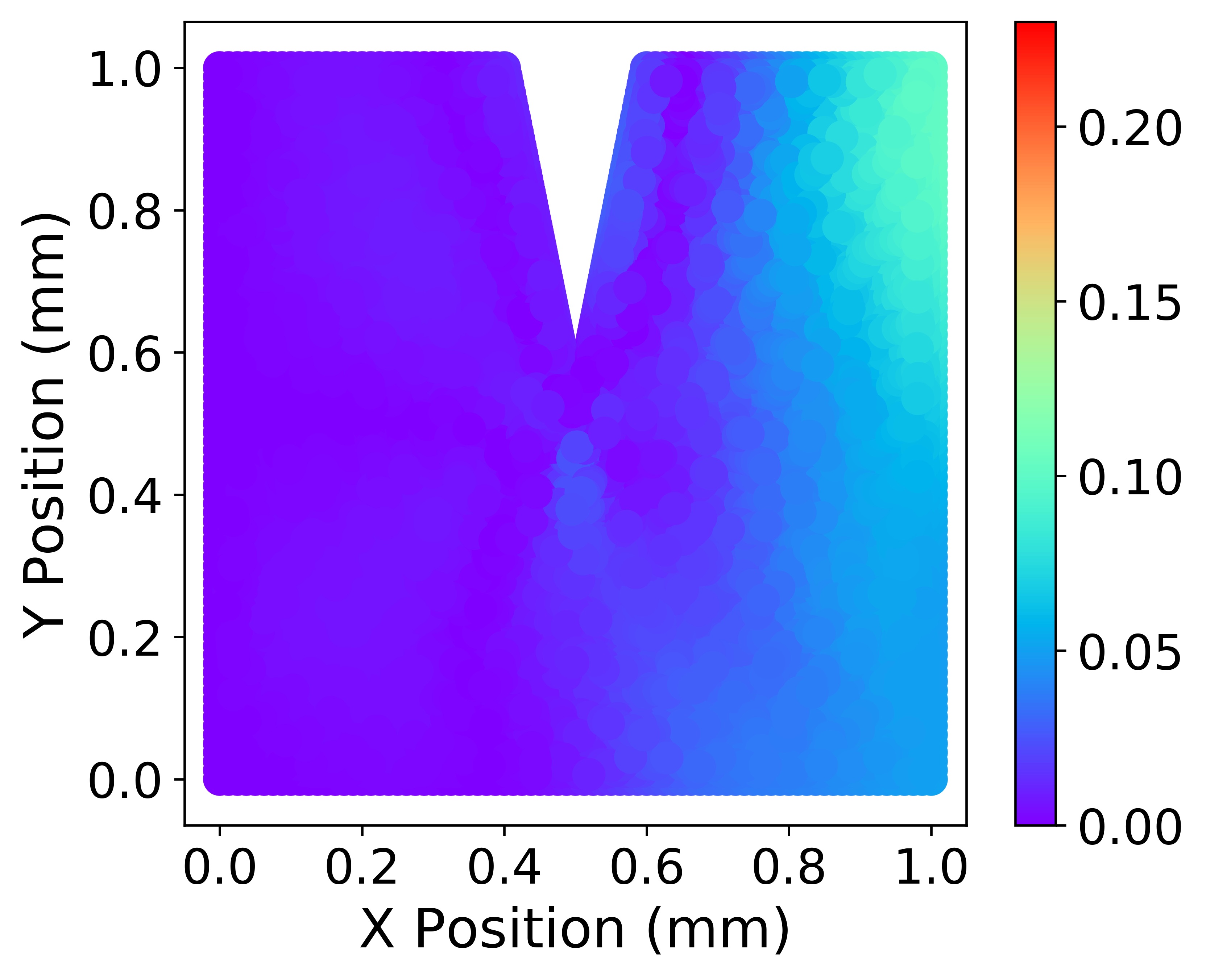}
    \end{minipage}
    \\ \hline
    DF
    &
    \begin{minipage}{.19\textwidth}
      \includegraphics[width=0.9\linewidth]{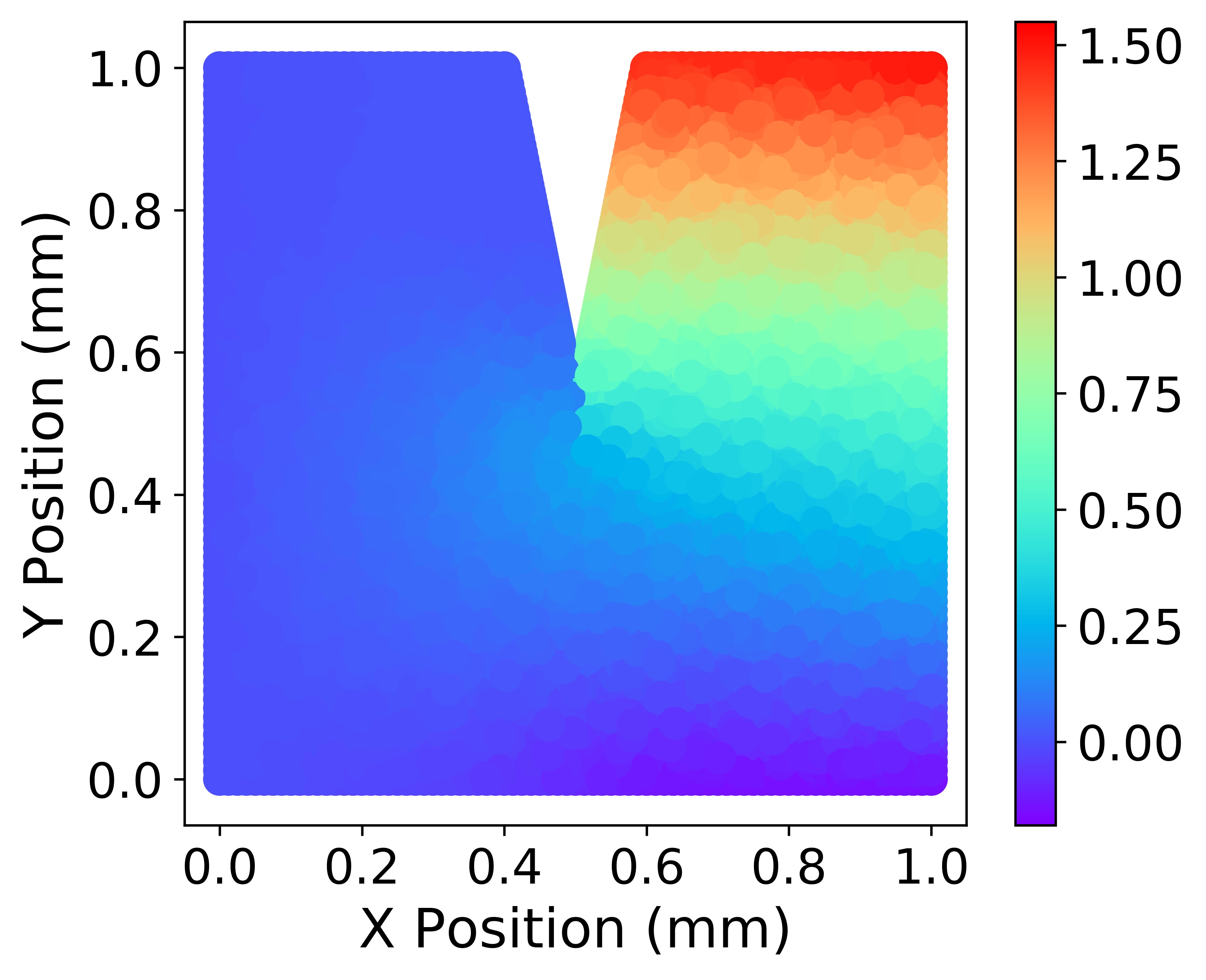}
    \end{minipage}
    &
   \begin{minipage}{.19\textwidth}
      \includegraphics[width=0.9\linewidth]{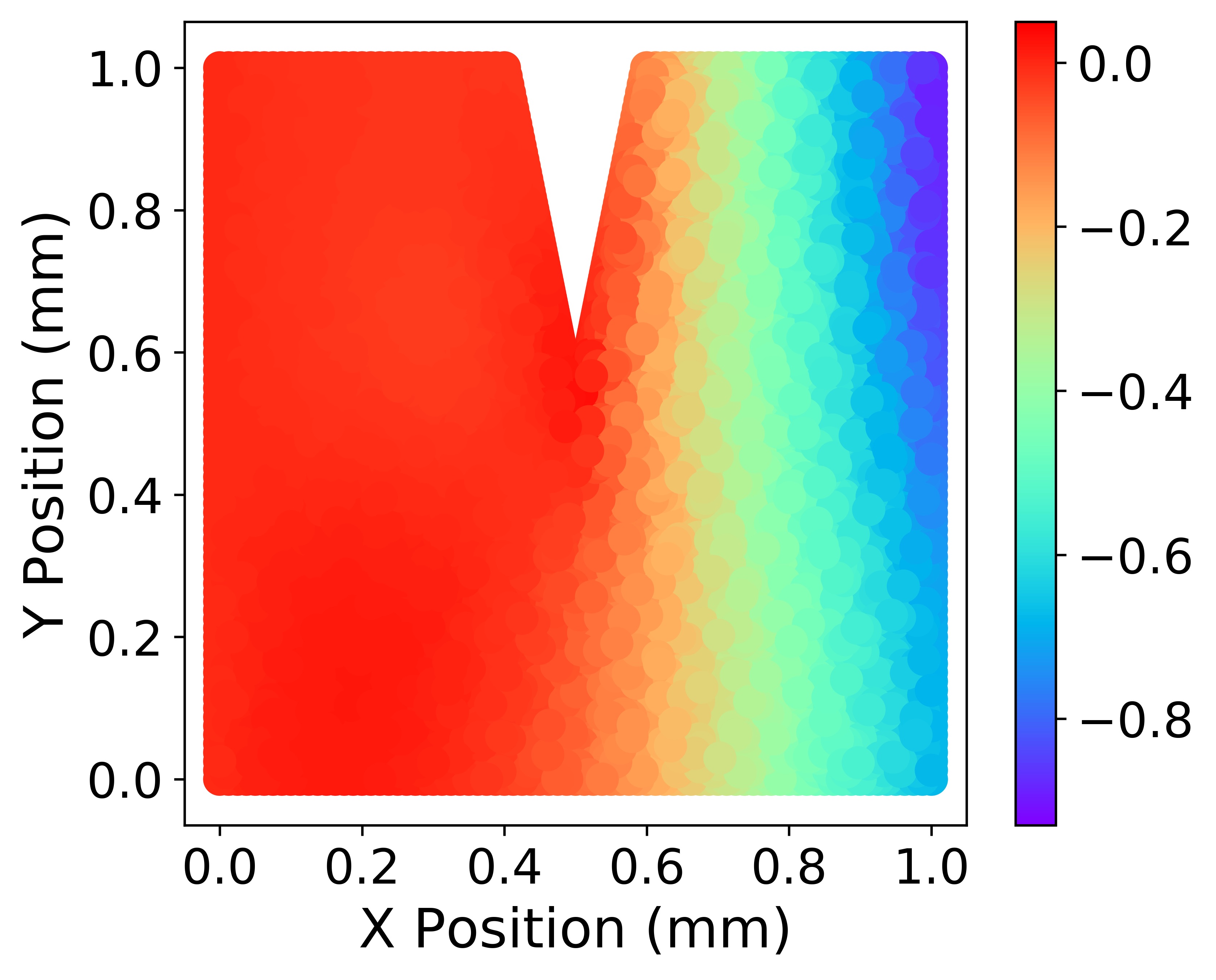}
    \end{minipage}
    & 
   \begin{minipage}{.19\textwidth}
      \includegraphics[width=0.9\linewidth]{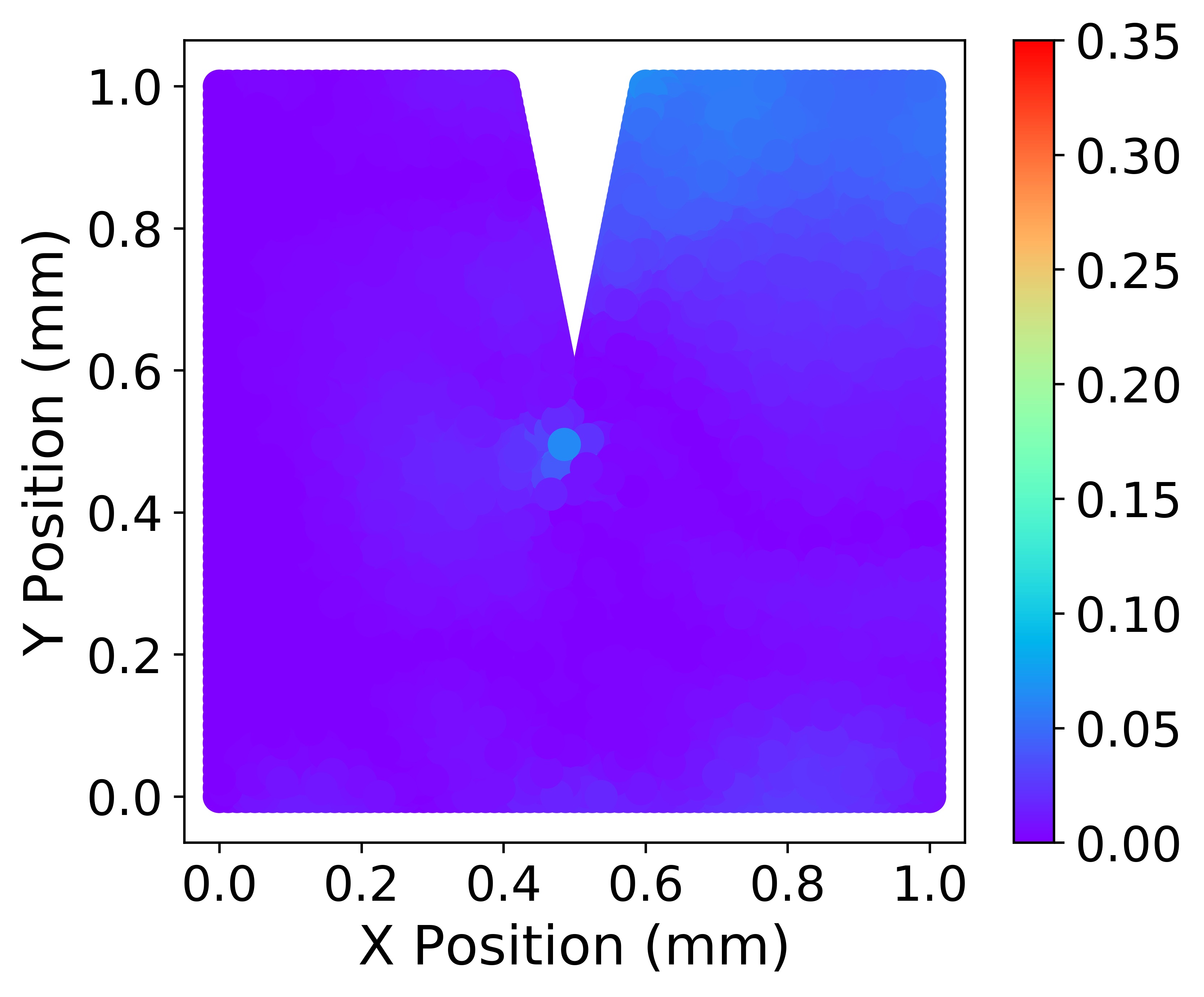}
    \end{minipage}
     & 
   \begin{minipage}{.19\textwidth}
      \includegraphics[width=0.9\linewidth]{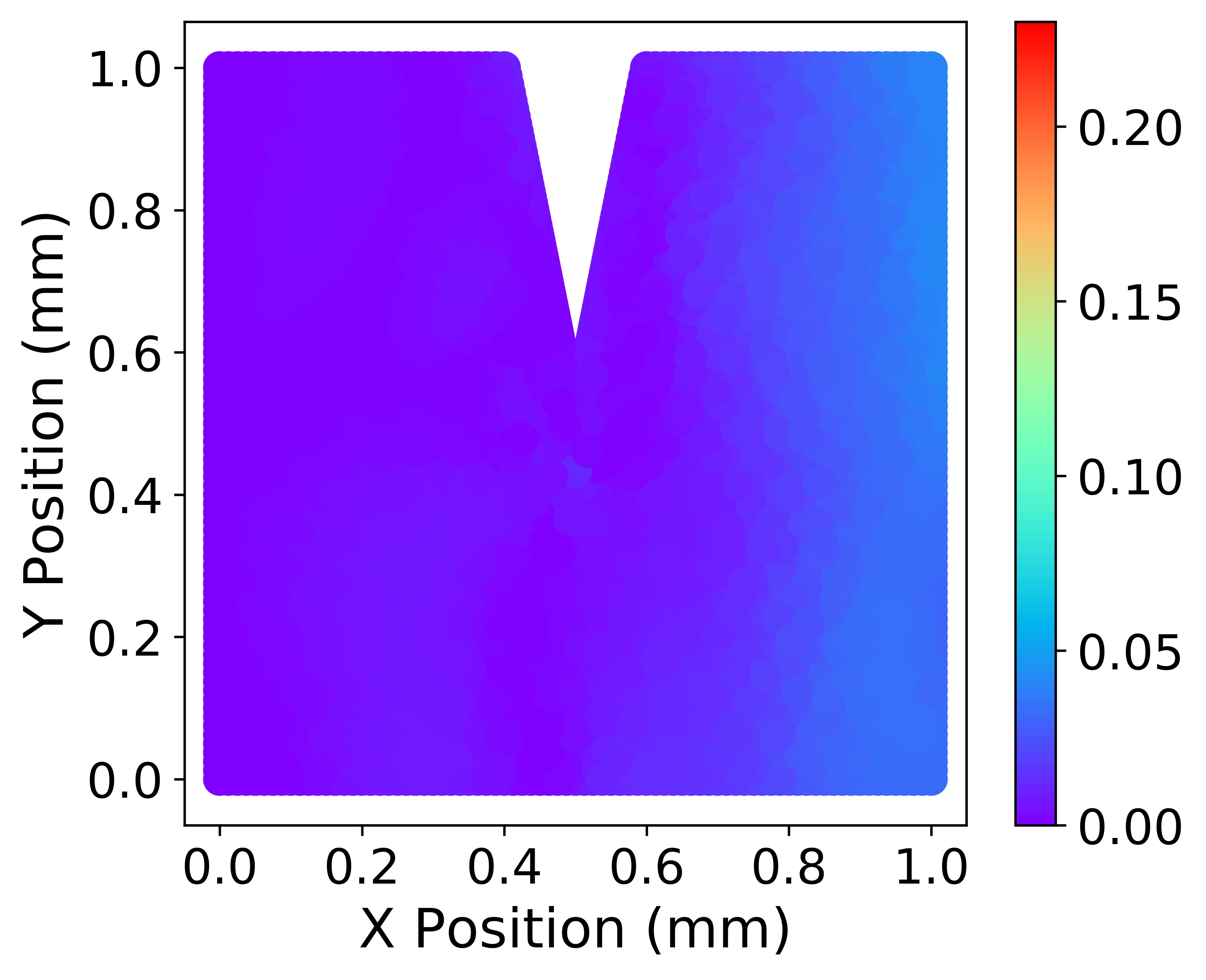}
    \end{minipage}
    \\ \hline
    PINN-FEM
    &
    \begin{minipage}{.19\textwidth}
      \includegraphics[width=0.9\linewidth]{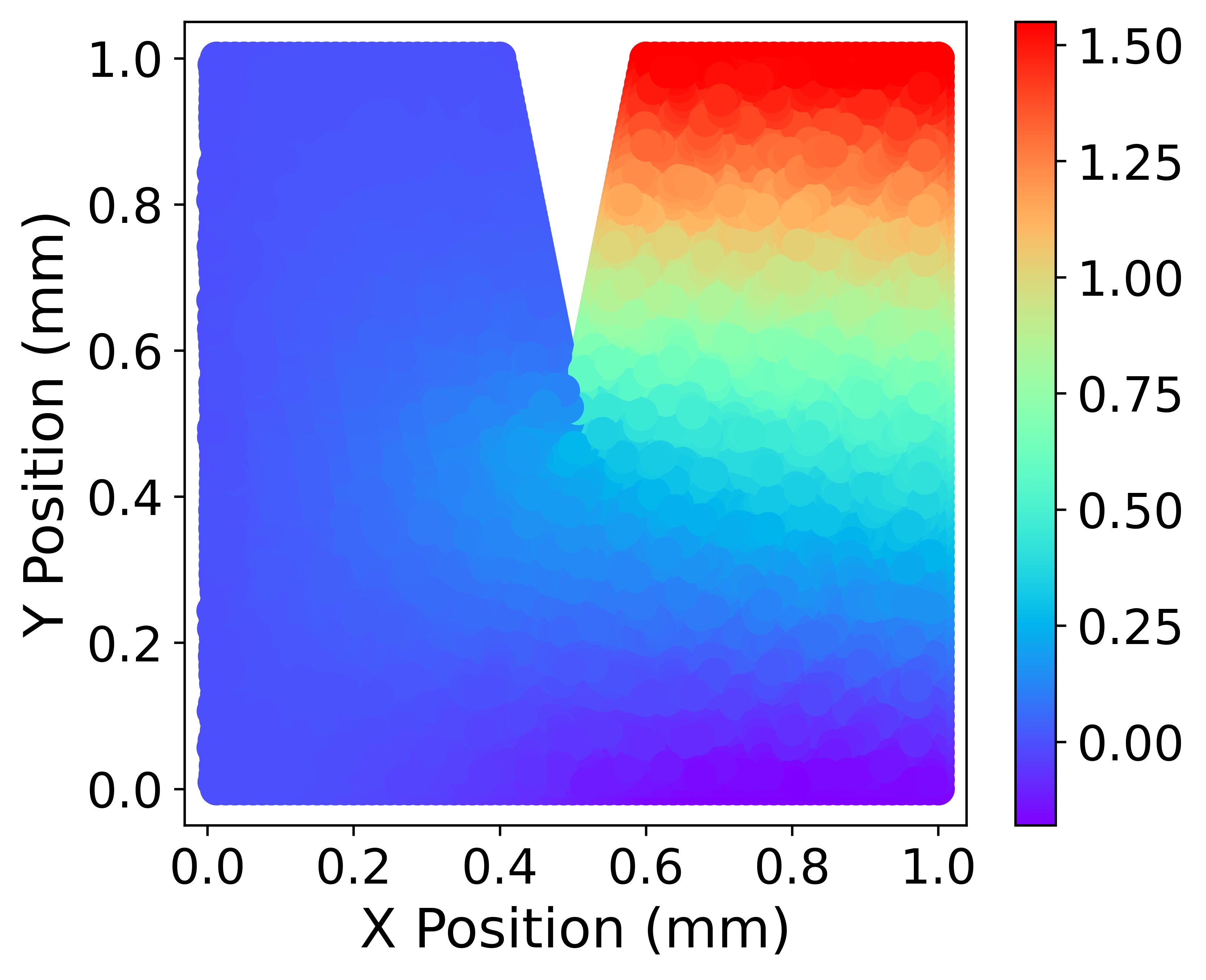}
    \end{minipage}
    &
   \begin{minipage}{.19\textwidth}
      \includegraphics[width=0.9\linewidth]{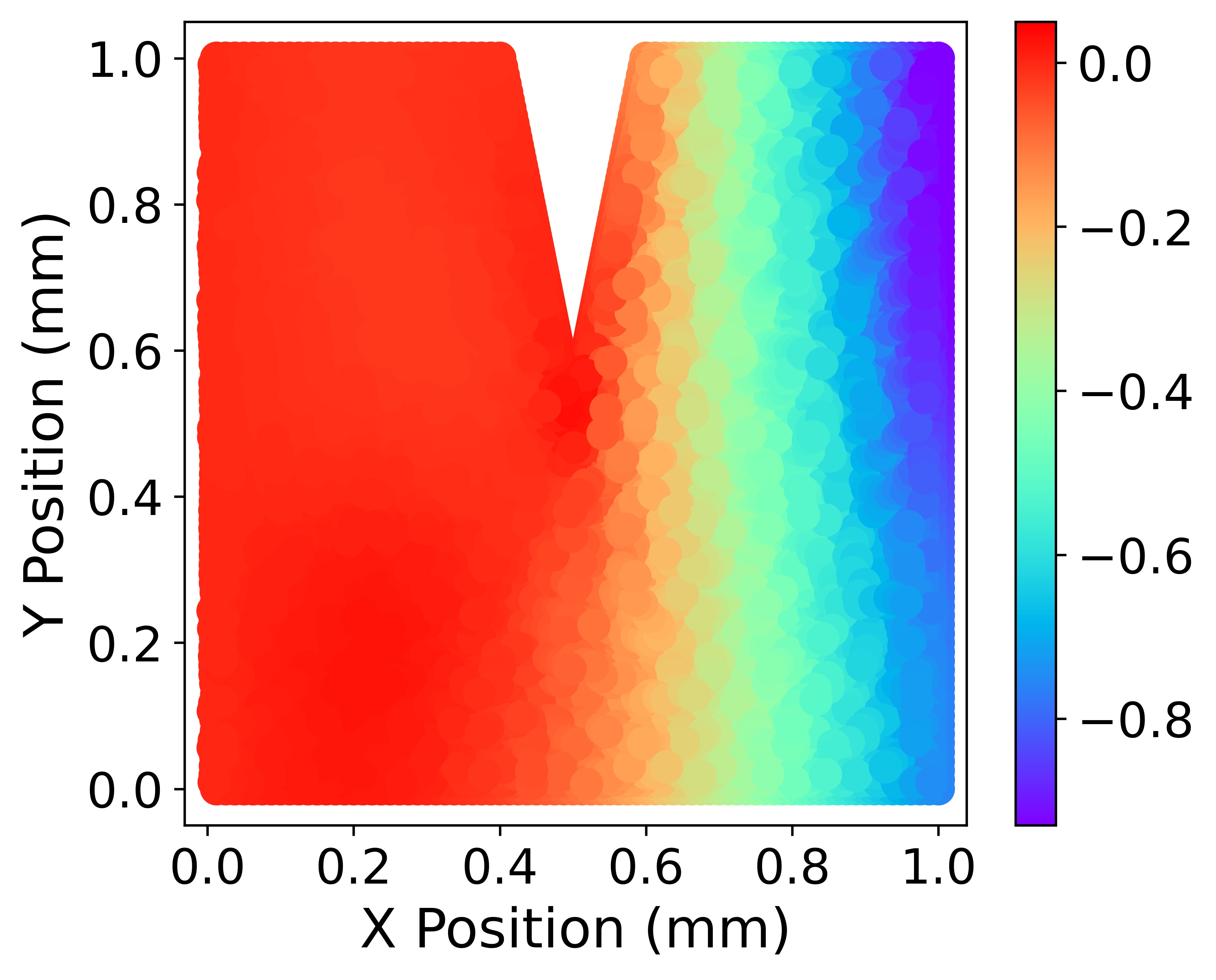}
    \end{minipage}
    & 
   \begin{minipage}{.19\textwidth}
      \includegraphics[width=0.9\linewidth]{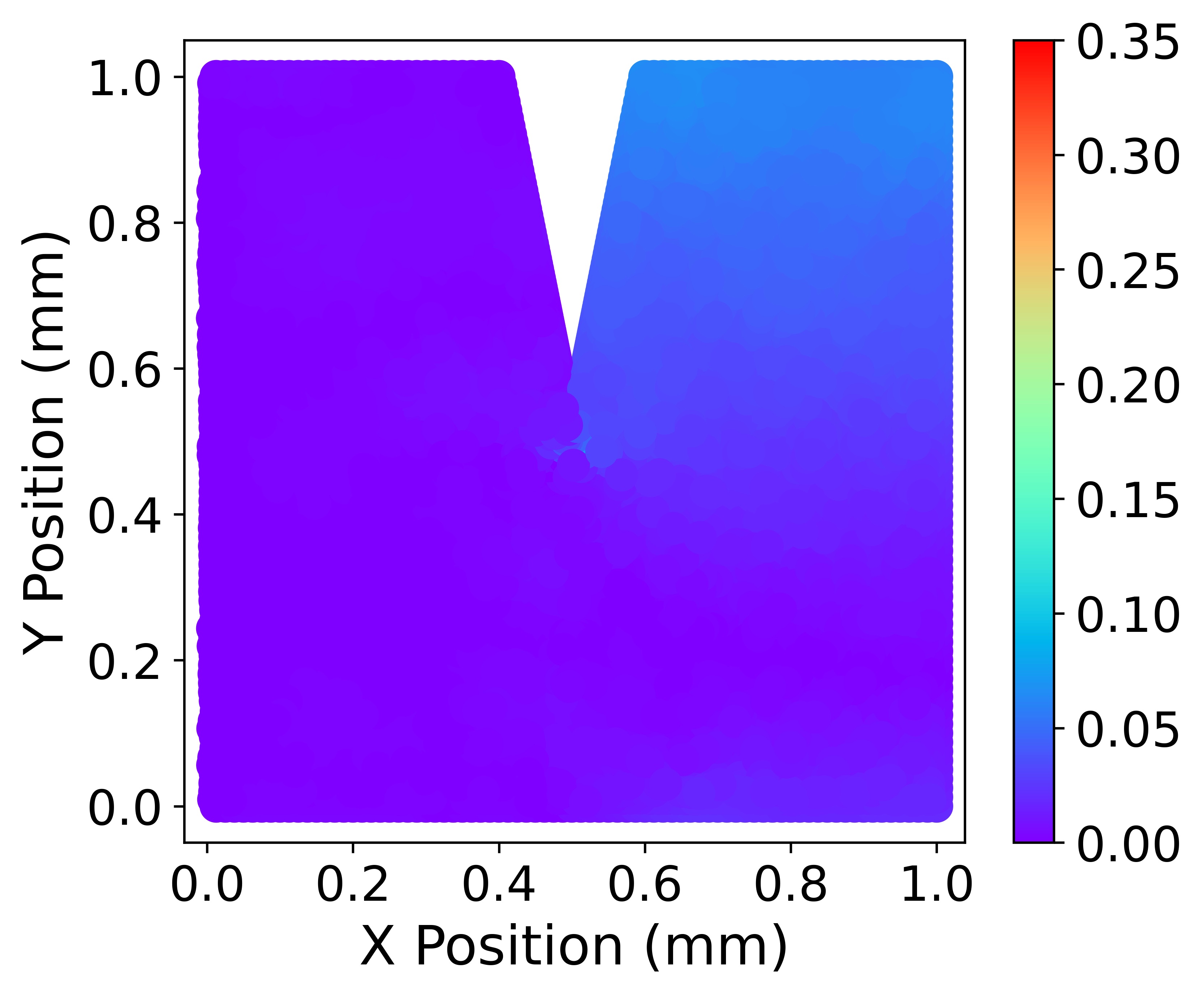}
    \end{minipage}
     & 
   \begin{minipage}{.19\textwidth}
      \includegraphics[width=0.9\linewidth]{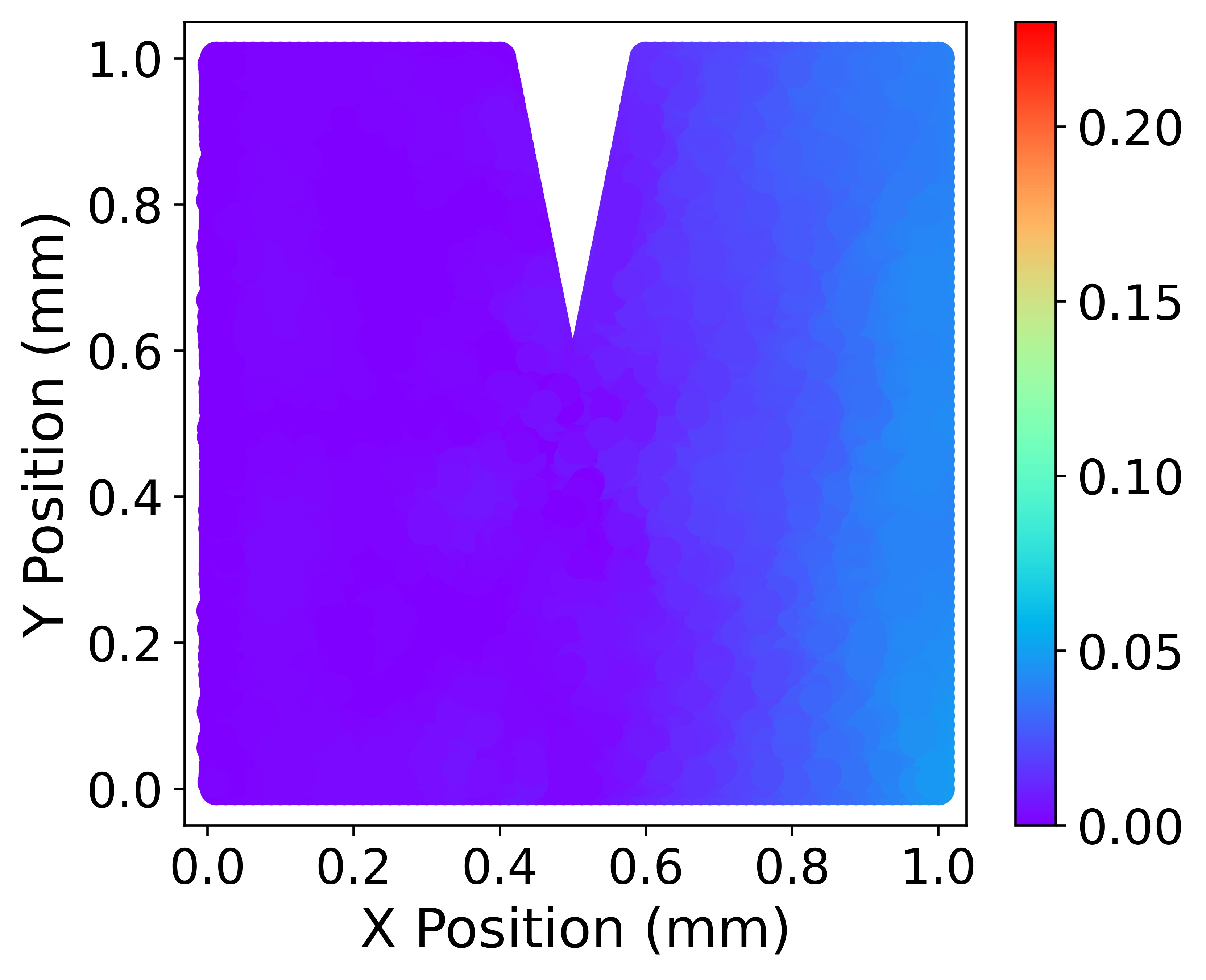}
    \end{minipage}
    \\ \hline
  \end{tabular}
  \label{fig:crack_pred}
\end{table}

\subsubsection{Cantilever beam with parabolic traction}

For the final experiment, we consider a cantilever beam under plane strain conditions. Let the  cantilever beam be of depth D, length L, and unit thickness, with a parabolic shear load of magnitude $P$ at the free end. The displacement boundary conditions at the left end of the beam are prescribed as described in Table \ref{table:exps_doms_bcs}. Timoshenko and Goodier, in \cite{timoshenko1975elasticity} showed that the stress field in the cantilever is given by

\begin{equation}
    \begin{aligned}
    \sigma_{xx} &= \frac{P(L-x)y}{I}\\
    \sigma_{yy} &= 0 \\
    \sigma_{xy} &= -\frac{P}{2I}(\frac{D^2}{4}-y^2),
\label{eq_timoshenko_stress}
    \end{aligned}
\end{equation}
and the exact solution for displacement field is given by,

\begin{equation}
    \begin{aligned}
    u_{x} &= \frac{Py}{6EI}[(6L-3x)x+(2+\nu)(y^2-\frac{D^2}{4})],\\
    u_{y} &= -\frac{P}{6EI}[3\nu y^2(L-x)+(4+5\nu)\frac{D^2x}{4}+(3L-x)x^2], \\
\label{eq_timoshenko_disp}
    \end{aligned}
\end{equation}
where $I=D^3/12$ is the moment of inertia by assuming unit thickness in 2D. This exact analytical solution is used as the ground truth (GT) for evaluating the performance of the models. The collocation points are generated using the mesh with Gmsh, using a mesh size of 0.1 mm. Table \ref{fig:exp_canti_pred} shows the distribution of the actual and predicted displacement fields in $x$ and $y$ directions and the corresponding absolute error for the vanilla PINN and the proposed PINN-FEM model. The proposed model outperforms the vanilla PINN by four orders of magnitude in terms of the relative error, as observed from the values given in Table \ref{table:rel_error}.

\begin{table}[!ht]
  \centering
    \caption{The distribution of predicted displacement fields, $\hat{u}_x$ and $\hat{u}_y$, and the corresponding error for the cantilever beam with point boundary conditions as given in Eq.~\eqref{bc_example6}. Ground truth shows the displacement fields calculated using the equations given in Eq. \ref{eq_timoshenko_disp}.}
  \begin{tabular}{ |c | c | c |c | c| }
    \hline
    PINN Model & $\hat{u}_x$ & $\hat{u}_y$ & $|\hat{u}_x-u_x|$ & $|\hat{u}_y-u_y|$ \\ \hline
    Ground Truth
    &
    \begin{minipage}{.19\textwidth}
      \includegraphics[width=0.9\linewidth]{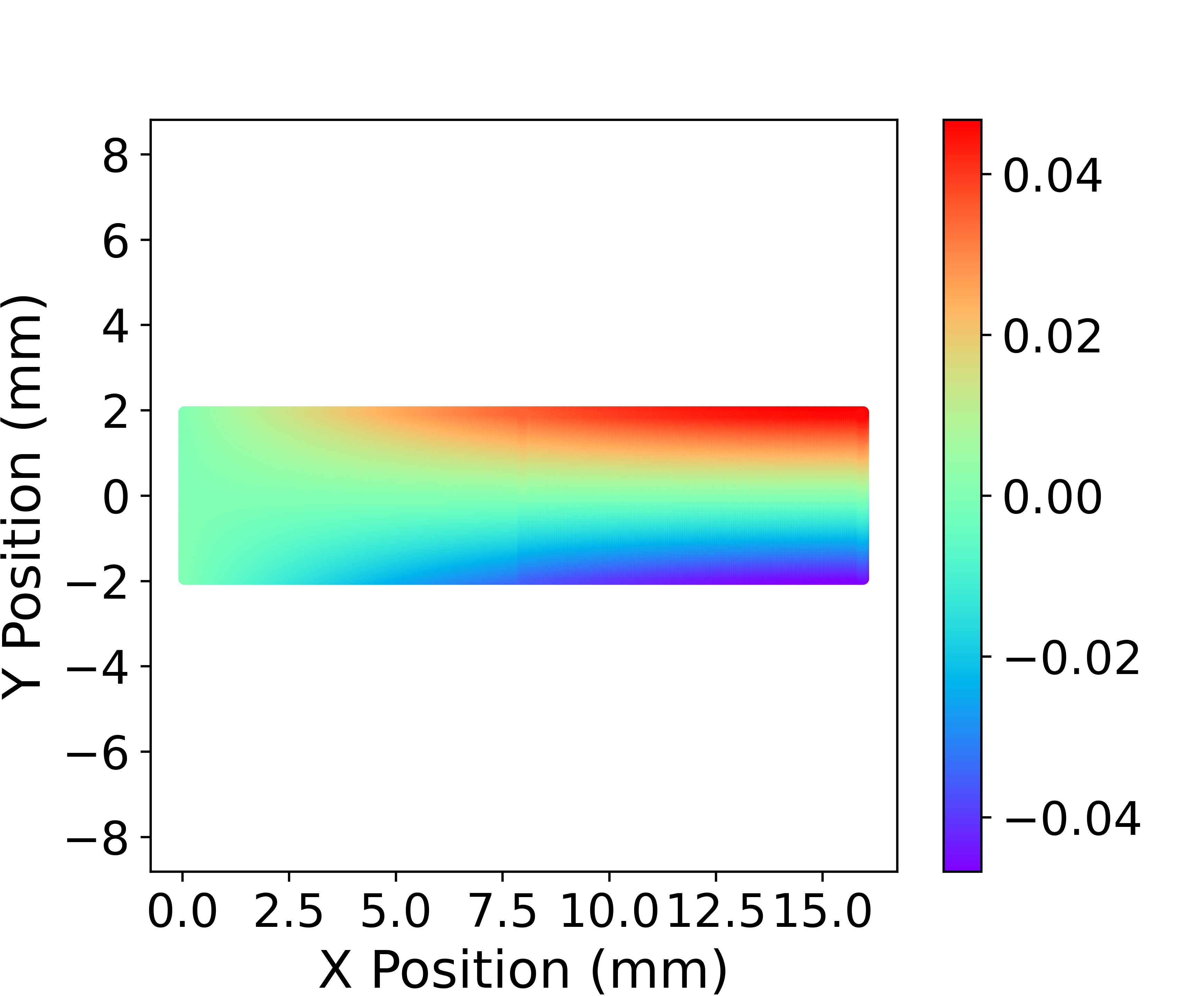}
    \end{minipage}
    &
   \begin{minipage}{.19\textwidth}
      \includegraphics[width=0.9\linewidth]{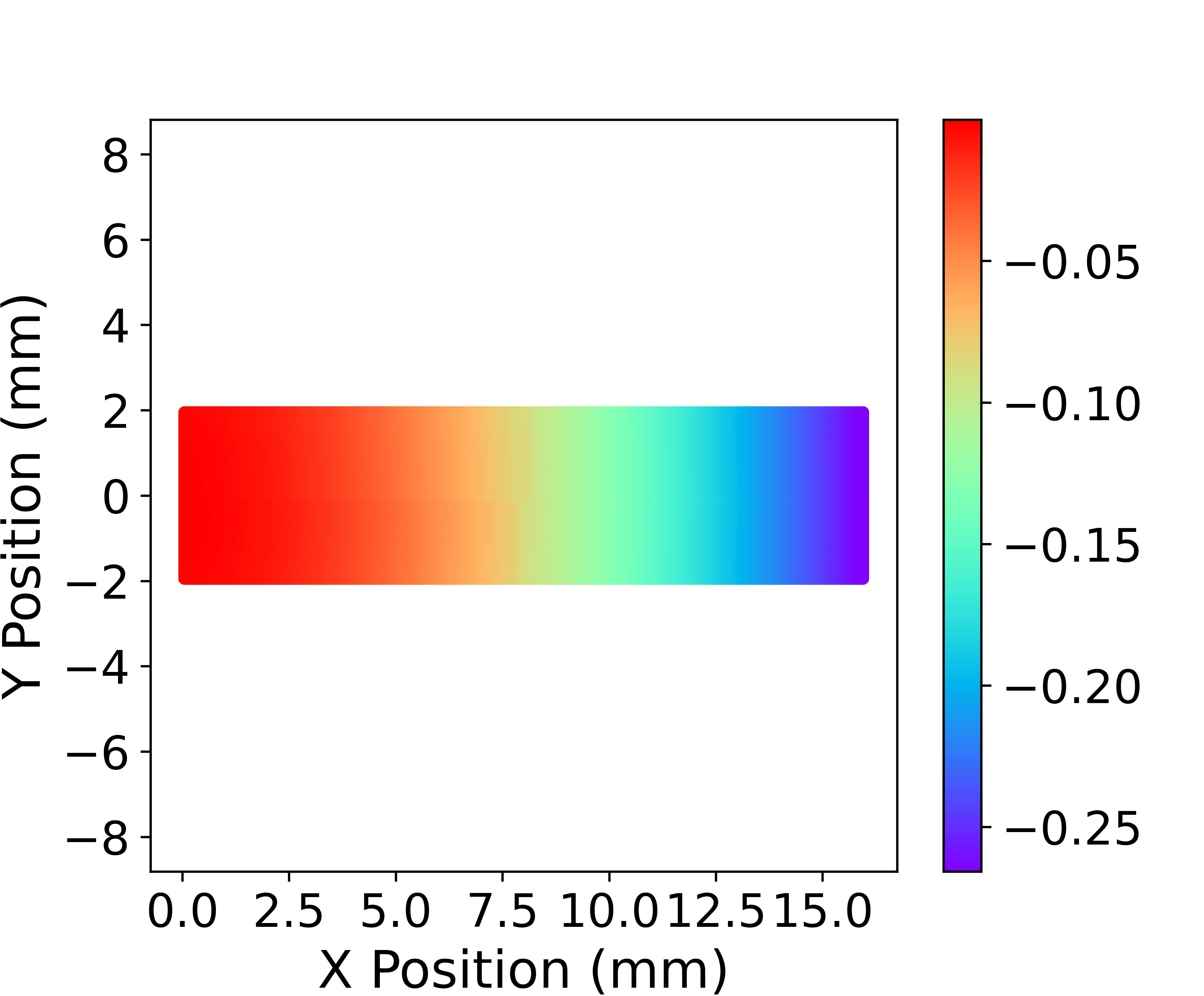}
    \end{minipage}
    & 
     & 
    \\ \hline
    Soft
    &
    \begin{minipage}{.19\textwidth}
      \includegraphics[width=0.9\linewidth]{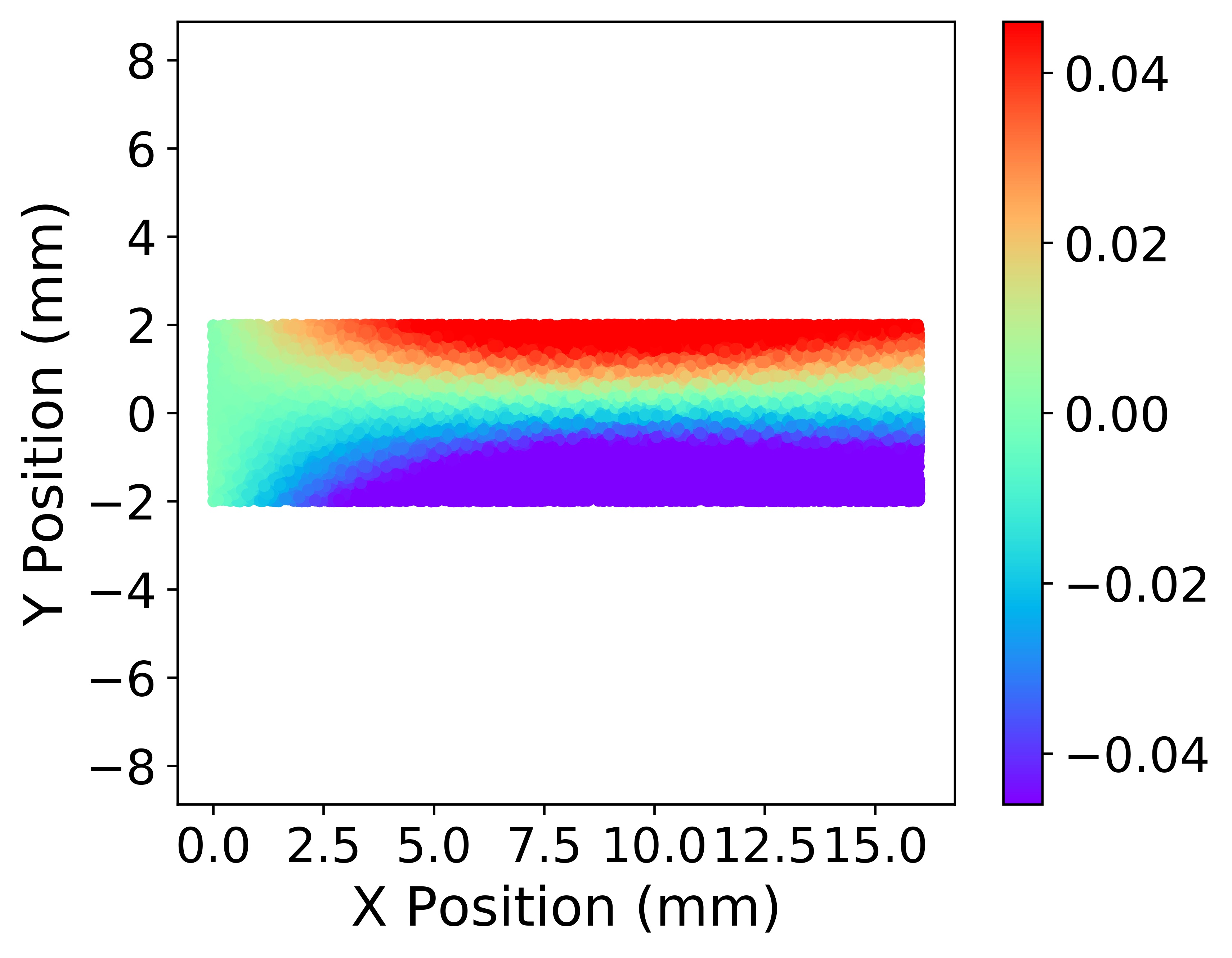}
    \end{minipage}
    &
   \begin{minipage}{.19\textwidth}
      \includegraphics[width=0.9\linewidth]{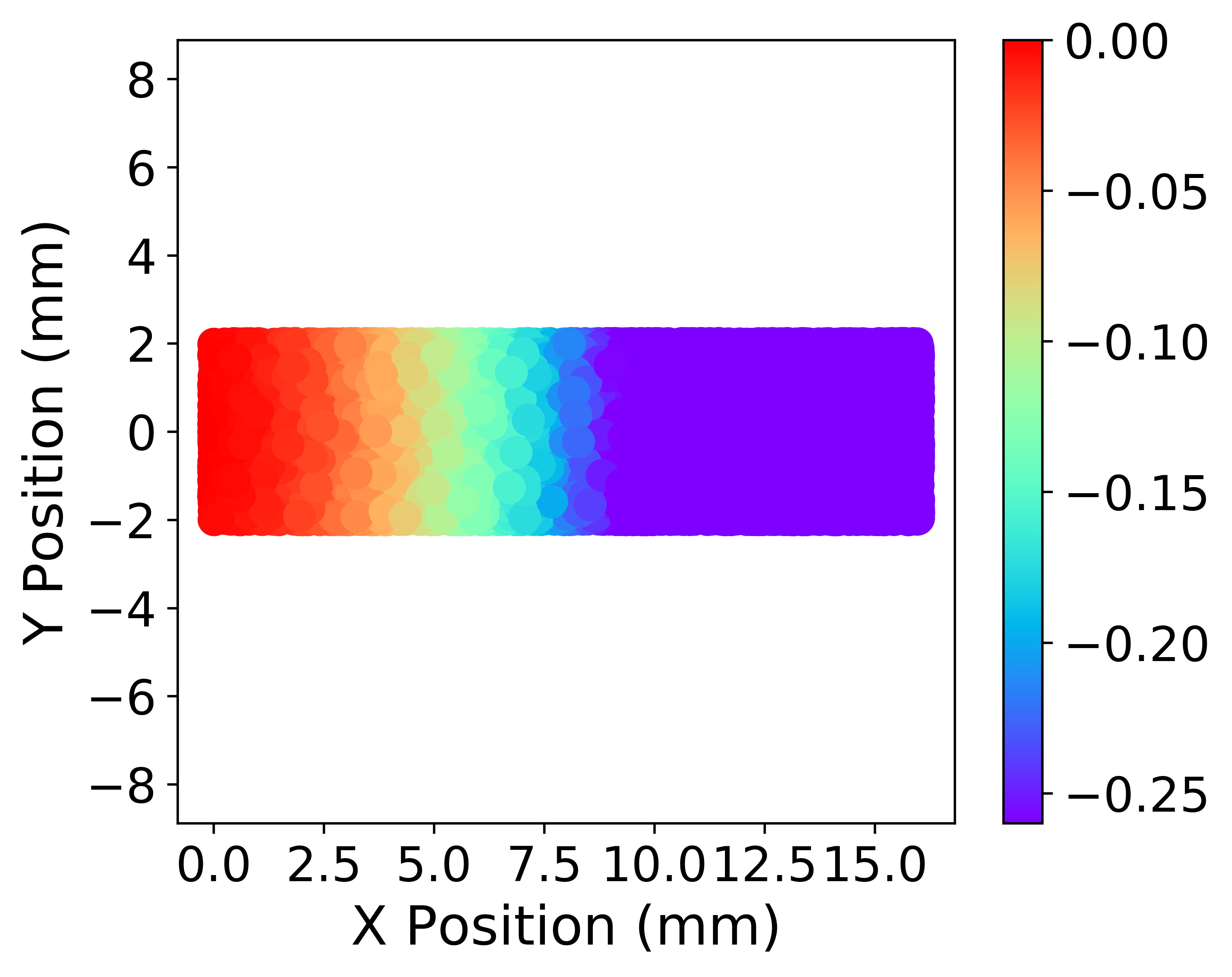}
    \end{minipage}
    & 
   \begin{minipage}{.19\textwidth}
      \includegraphics[width=0.9\linewidth]{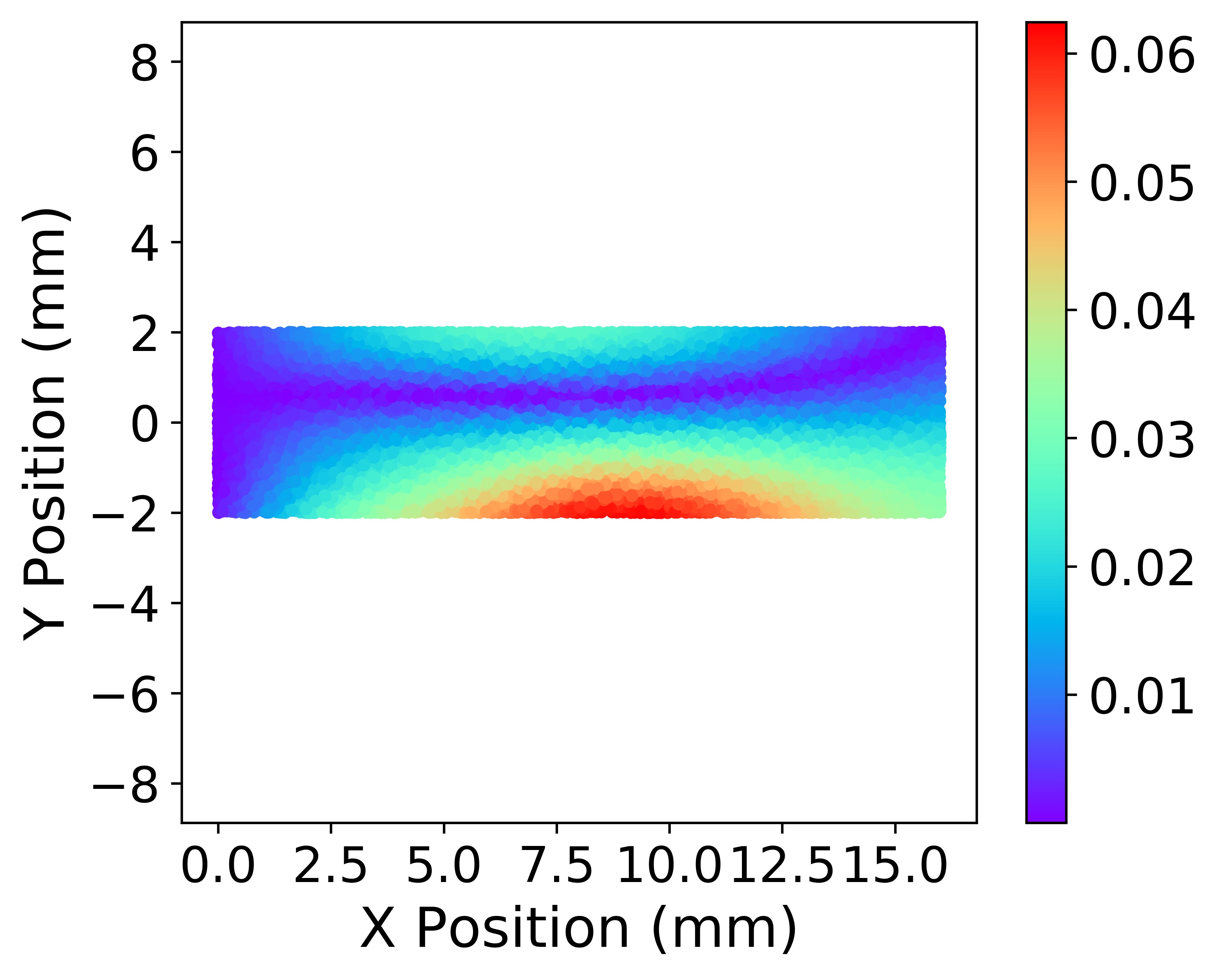}
    \end{minipage}
     & 
   \begin{minipage}{.19\textwidth}
      \includegraphics[width=0.9\linewidth]{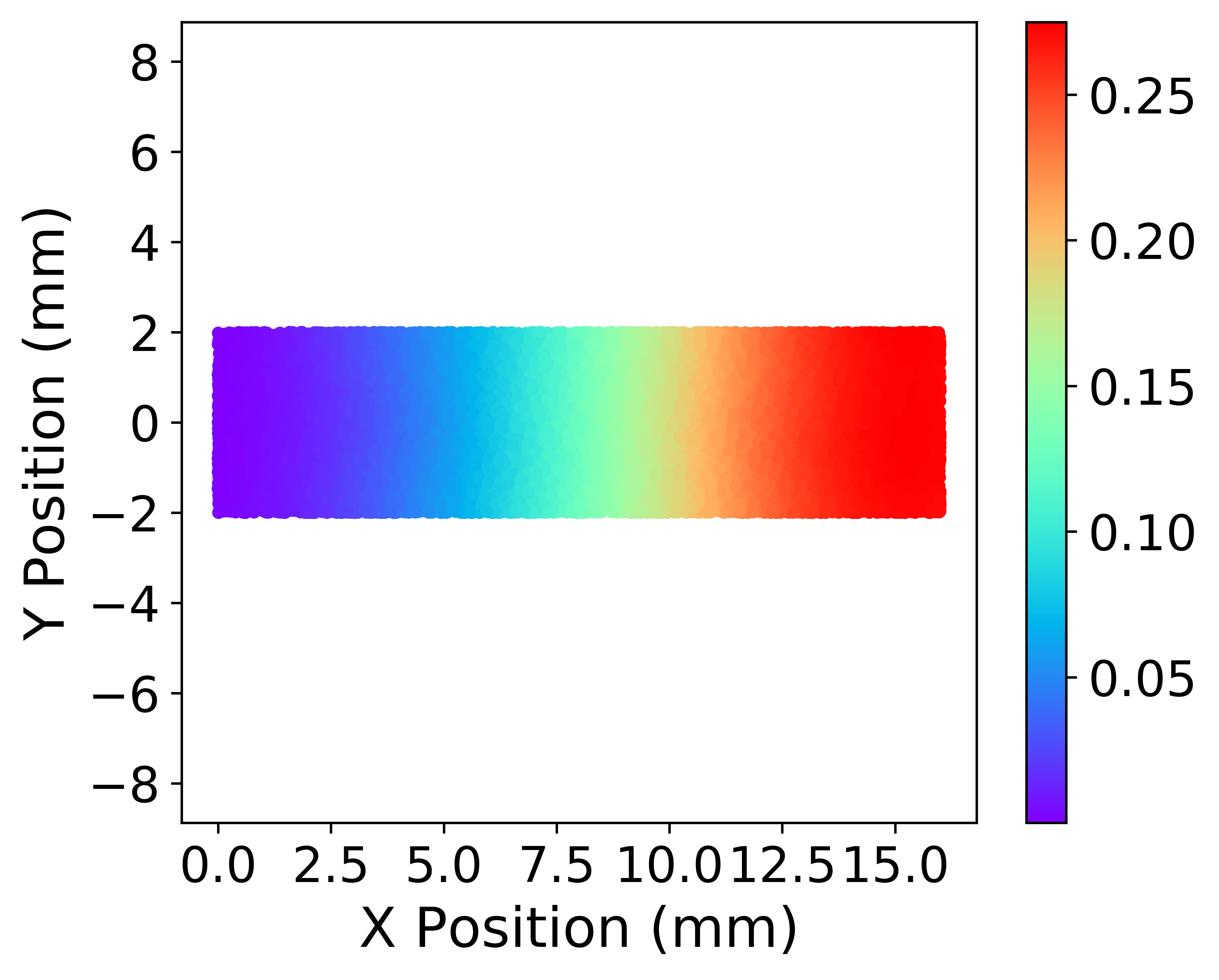}
    \end{minipage}
    \\ \hline
    PINN-FEM
    &
    \begin{minipage}{.19\textwidth}
      \includegraphics[width=0.9\linewidth]{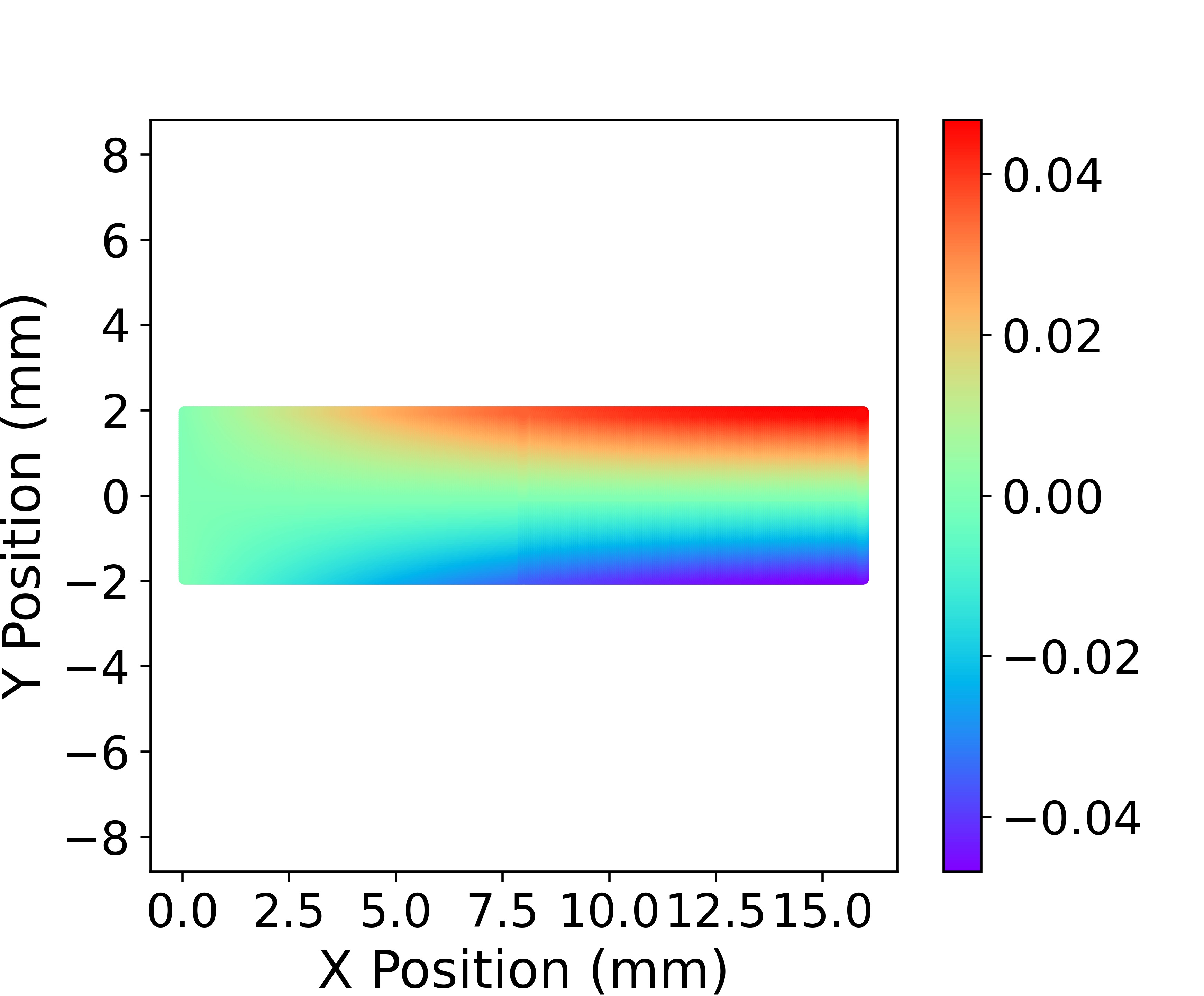}
    \end{minipage}
    &
   \begin{minipage}{.19\textwidth}
      \includegraphics[width=0.9\linewidth]{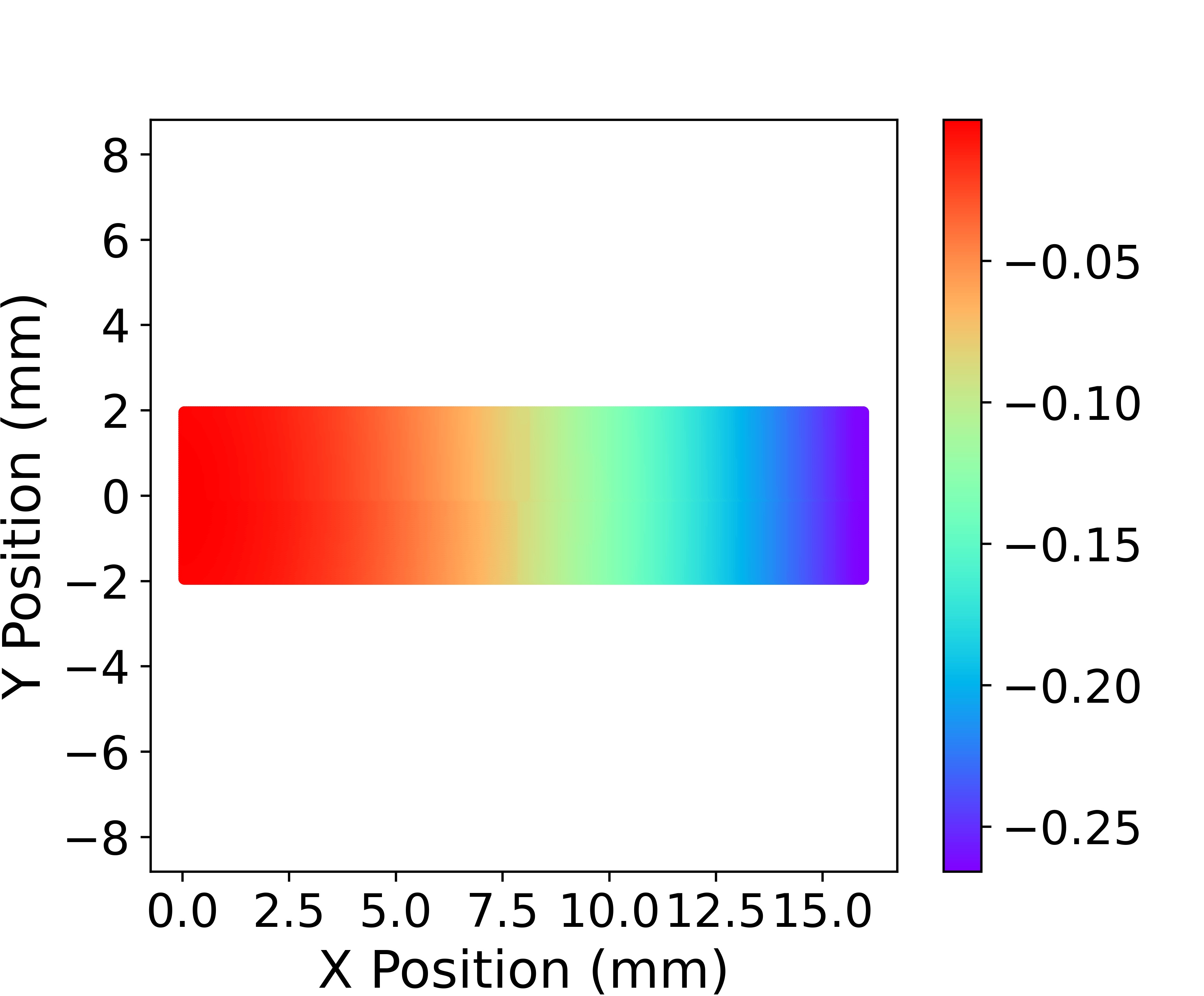}
    \end{minipage}
    & 
   \begin{minipage}{.19\textwidth}
      \includegraphics[width=0.9\linewidth]{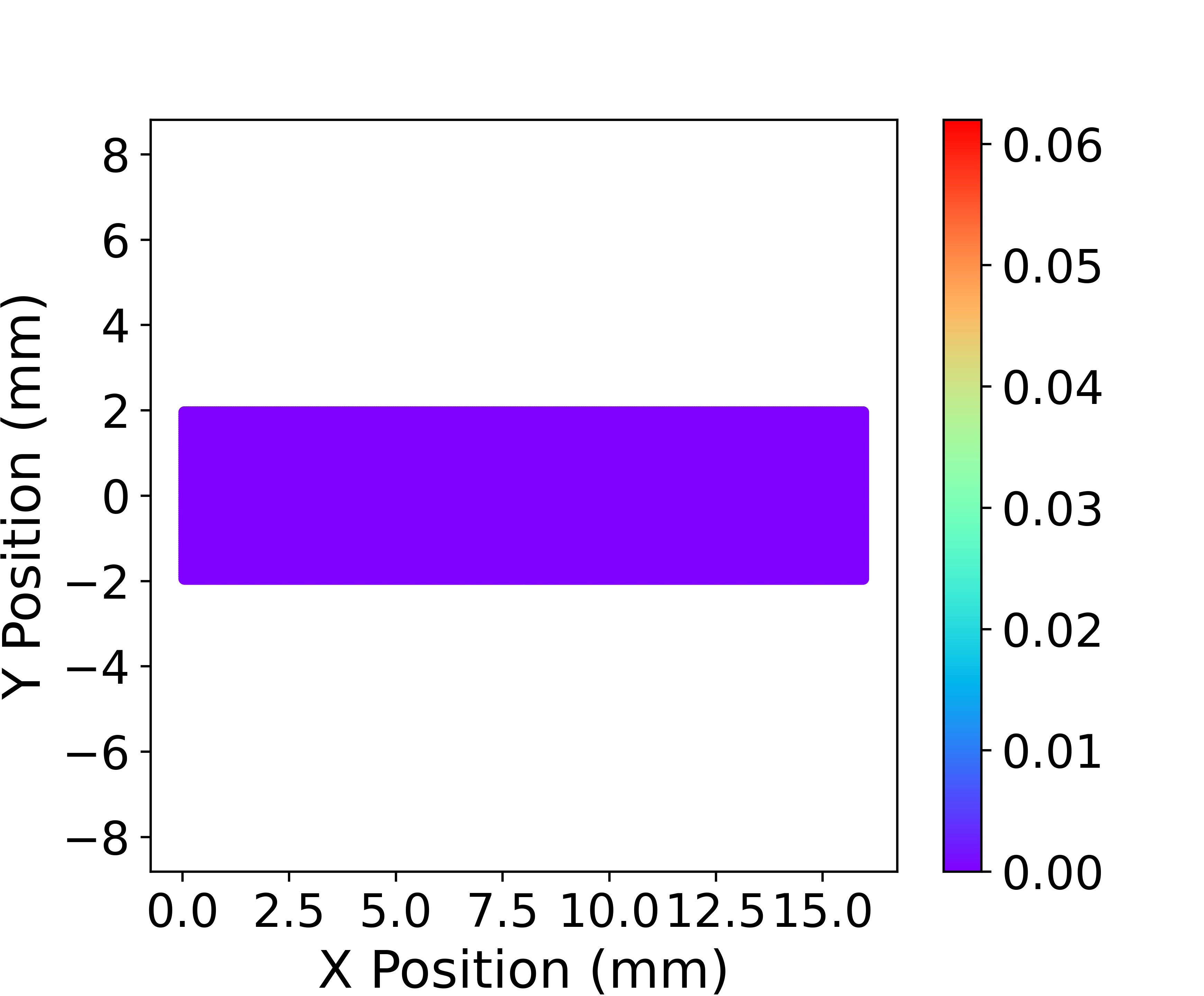}
    \end{minipage}
     & 
   \begin{minipage}{.19\textwidth}
      \includegraphics[width=0.9\linewidth]{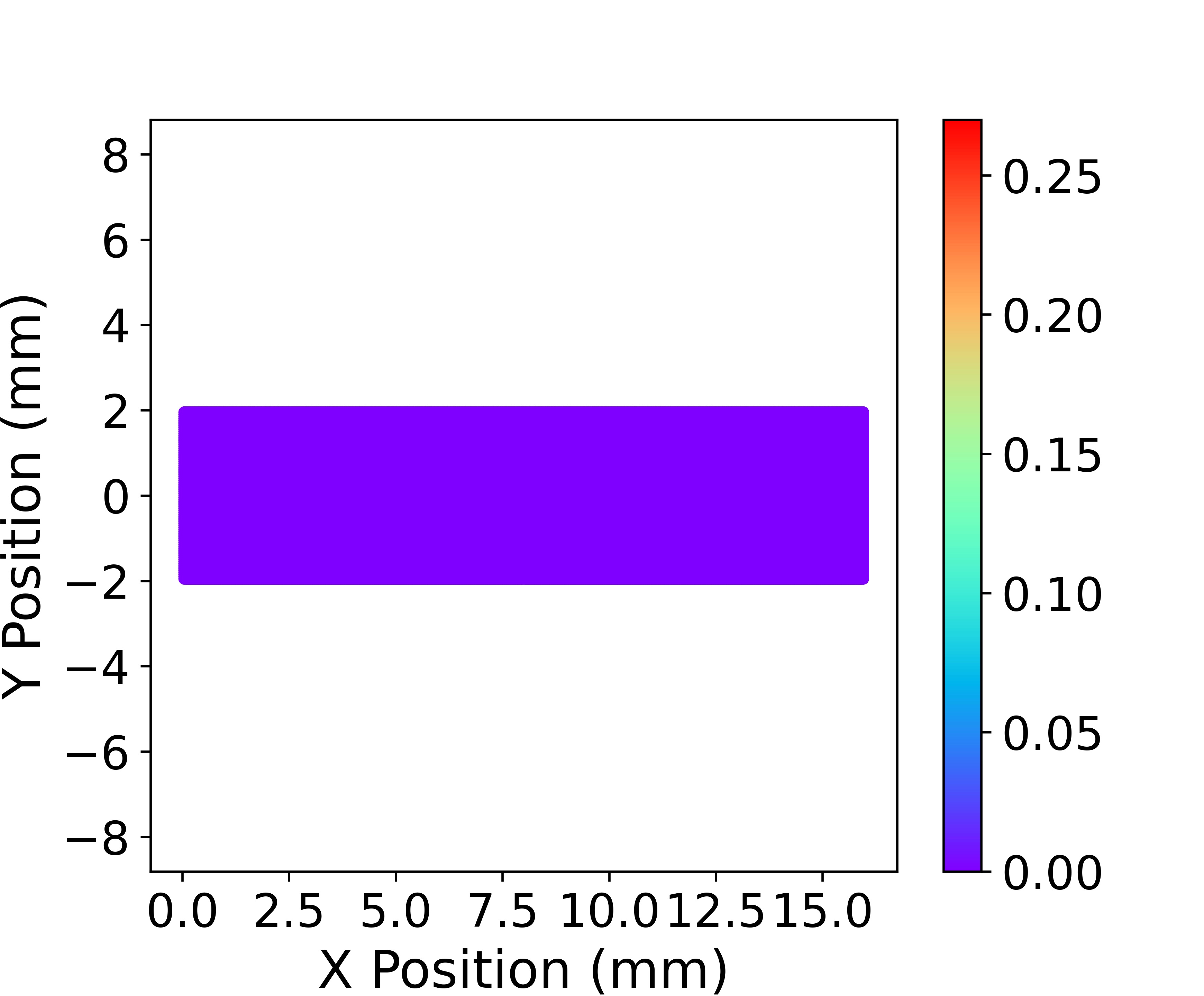}
    \end{minipage}
    \\ \hline
  \end{tabular}
  \label{fig:exp_canti_pred}
\end{table}

\subsection{Summary}
\label{results_summary}

For all the experiments, the proposed PINN-FEM model consistently demonstrates best performing. The accuracy of all the PINN models with exact Dirichlet BC imposition, i.e., PINNs with ADF and Distance Function as well as PINN-FEM method, are similar for the simple case - square plate with constant stress. However, global functions cannot serve as distance functions for BC imposition in cases with discontinuous BCs, as seen in square plate with circular hole, plate with discontinuous boundary for the left edge and plate with point boundaries. PINN with ADF does not work for point boundary conditions, as ADF cannot be defined for points. Moreover, these models have high errors when predicting displacement for plates with crack. For all the experiments, the vanilla PINN consistently performs the worst, as it employs soft imposition of Dirichlet BC. The proposed model successfully handles a wide range of boundary conditions and accurately predicts displacement for all cases, with exact imposition of Dirichlet BCs.

\section{Discussion}
\label{discussion}

In this study, we presented a novel blending method, PINN-FEM, designed for the exact imposition of Dirichlet boundary conditions in Physics-Informed Neural Networks (PINNs). By leveraging domain decomposition, PINN-FEM integrates the computational flexibility of PINNs with the robust enforcement capabilities of finite element methods (FEM). The method applied FEM near Dirichlet boundaries to ensure strong boundary condition enforcement, while PINNs approximate the solution in the interior domain. An energy-based loss function, derived from the variational principle, enhances stability and efficiency during training and inherently satisfies Neumann boundary conditions, simplifying the overall training process. The effectiveness of PINN-FEM was validated through comparisons with baseline PINN models, including those employing soft and exact Dirichlet boundary condition implementations. Our results demonstrated that PINN-FEM consistently outperformed the baseline models across a wide range of experiments, including cases with discontinuous and point Dirichlet boundary conditions, as well as domains with complex geometries such as cracks. Traditional PINN approaches relying on approximate distance functions or global functions for enforcing exact boundary struggled with discontinuous boundaries, often resulting in significant errors near abrupt boundary condition changes. In contrast, PINN-FEM effectively handles these challenges, showcasing its robustness and adaptability. The proposed PINN-FEM method addresses several key limitations of traditional PINNs, particularly for enforcing essential boundary conditions in complex domains. This makes it a promising tool for industrial and scientific applications where exact boundary condition enforcement is critical.

\bibliographystyle{unsrt}  
\bibliography{references}  

\end{document}